\documentclass[pdflatex,sn-mathphys-num]{sn-jnl}

\usepackage{graphicx}%
\usepackage{multirow}%
\usepackage{amsmath,amssymb,amsfonts}%
\usepackage{amsthm}%
\usepackage{mathrsfs}%
\usepackage[title]{appendix}%
\usepackage{xcolor}%
\usepackage{textcomp}%
\usepackage{manyfoot}%
\usepackage{booktabs}%
\usepackage{algorithm}%
\usepackage{algorithmicx}%
\usepackage{algpseudocode}%
\usepackage{listings}%

\usepackage[utf8]{inputenc} %
\usepackage[T1]{fontenc}    %
\usepackage{url}            %
\usepackage{booktabs}       %
\usepackage{natbib}

\usepackage{multicol}
\usepackage{xcolor}   
\usepackage{multirow}
\usepackage{microtype}      %
\usepackage{enumitem}
\usepackage{tabularx}
\usepackage{graphicx}       %
\usepackage{caption} 
\usepackage{float}
\usepackage{adjustbox}
\usepackage{amsmath, amsfonts, amssymb, bbm}
\usepackage{pifont}
\usepackage{lineno}

\newcommand{\cmark}{\textcolor{green}{\ding{51}}} 
\newcommand{\xmark}{\textcolor{red}{\ding{55}}}   

\newcommand{\pkd}[1]{\texttt{{\{\{\MakeUppercase{#1}\}\}}}}

\usepackage{array}
\usepackage{minted}
\usepackage[most]{tcolorbox}
\usepackage{listings}
\usepackage{xcolor}
\definecolor{softblue}{RGB}{100, 149, 237}
\definecolor{darkpurple}{RGB}{160, 0, 160}
\definecolor{darkred}{RGB}{160, 0, 0}
\definecolor{darkblue}{RGB}{0, 0, 160}
\definecolor{softgreen}{RGB}{120, 180, 100}

\definecolor{darkgreen}{rgb}{0.0, 0.5, 0.0}
\definecolor{darkred}{rgb}{0.5, 0.0, 0.0}
\definecolor{bg}{rgb}{0.97,0.97,0.97}   
\definecolor{codegray}{rgb}{0.3,0.3,0.3} 
\definecolor{codegreen}{rgb}{0,0.6,0}    
\definecolor{backcolour}{rgb}{0.95,0.95,0.92} 

\tcbuselibrary{listingsutf8}
\newtcolorbox{bg_yaml}{
  enhanced, 
  sharp corners=southwest,
  colback=lightgray!20, 
  colframe=black,       
  boxrule=0.5mm,       
  before=\vspace{1mm}, 
  after=\vspace{1mm}   
}

\lstdefinestyle{pythonstyle}{
    language=Python, 
    basicstyle=\ttfamily,  
    keywordstyle=\color{blue}, 
    commentstyle=\color{gray},   
    stringstyle=\color{red}, 
    showstringspaces=false,        
    numbers=left,                 
    numberstyle=\tiny\color{gray},
    stepnumber=1,                
    numbersep=5pt,               
    frame=single,                
    tabsize=4                     
}

\lstdefinestyle{yamlstyle}{
    basicstyle=\ttfamily\footnotesize,
    keywordstyle=\color{blue},
    commentstyle=\color{green!60!black},
    stringstyle=\color{red!60!black},
    numbers=left,
    numberstyle=\tiny\color{gray},
    breaklines=true,
}

\lstdefinelanguage{YAML}{
  keywords={true,false,null,y,n},
  keywordstyle=\color{blue}\bfseries,
  basicstyle=\ttfamily\small,
  commentstyle=\color{green!50!black},
  stringstyle=\color{red!60!black},
  identifierstyle=\color{black},
  numbers=left,
  numberstyle=\tiny\color{gray},
  breaklines=true,
  literate=
   *{:}{{{\color{blue}:}}}{1}
    {-}{{{\color{red}-}}}{1}
}

\theoremstyle{thmstyleone}%
\theoremstyle{thmstyletwo}%

\theoremstyle{thmstylethree}%

\begin{document}

\title[Article Title]{SelfAI: A self-directed framework for long-horizon scientific discovery}


\author[1,2]{\fnm{Xiao} \sur{Wu}}

\author*[1]{\fnm{Ting-Zhu} \sur{Huang}}\email{tingzhuhuang@126.com}

\author[1]{\fnm{Liang-Jian} \sur{Deng}}

\author[5]{\fnm{Xiaobing} \sur{Yu}}

\author[1]{\fnm{Yu} \sur{Zhong}}

\author[3]{\fnm{Shangqi} \sur{Deng}}

\author[2]{\fnm{Ufaq} \sur{Khan}}

\author[6]{\fnm{Jianghao} \sur{Wu}}

\author[4]{\fnm{Xiaofeng} \sur{Liu}}

\author[2]{\fnm{Imran} \sur{Razzak}}

\author[2]{\fnm{Xiaojun} \sur{Chang}}

\author*[2]{\fnm{Yutong} \sur{Xie}}\email{yutong.xie@mbzuai.ac.ae}

\affil[1]{\orgname{University of Electronic Science and Technology of China}}
\affil[2]{\orgname{Mohamed bin Zayed University of Artificial Intelligence}}
\affil[3]{\orgname{Xian Jiaotong University}}
\affil[4]{\orgname{Yale University}}
\affil[5]{\orgname{Washington University in St. Louis}}
\affil[6]{\orgname{Monash University}}














\abstract{
Scientific discovery increasingly entails long-horizon exploration of complex hypothesis spaces, yet most existing approaches emphasize final performance while offering limited insight into how scientific exploration unfolds over time, particularly balancing efficiency-diversity trade-offs and supporting reproducible, human-in-the-loop discovery workflows. We introduce SelfAI, a self-directed, multi-agent-enabled discovery system that automates scientific exploration as a strategic, trajectory-driven decision-making process. SelfAI translates high-level research intent into executable experiments, reasons over accumulated experimental trajectories to guide subsequent exploration, and applies adaptive stopping decisions to terminate unproductive search paths within a closed-loop workflow governed by explicit efficiency-diversity trade-offs. Evaluated using real-world experiments spanning domains from machine learning to drug discovery, SelfAI consistently discovers high-quality solutions with substantially fewer redundant trials than classical optimization and recent LLM-based baselines. The proposed methods establish a general framework for organizing long-horizon scientific discovery and adaptive decision-making in complex scientific and engineering systems.
}

\maketitle

\clearpage



\section{Introduction}

Scientific discovery across engineering~\cite{bradshaw1983studying,angelopoulos2024transforming}, physics~\cite{lupoiu2025multi,kaiser2025large}, biology~\cite{lyu2024alphafold2,nguyen2024sequence}, and medicine~\cite{yakavets2025machine,grisoni2021combining,jiang2022artificial} increasingly relies on artificial intelligence (AI) systems to support or partially automate key stages of the scientific workflow, ranging from data analysis and hypothesis generation to experimental design and execution. As scientific problems grow in scale, dimensionality, and uncertainty, computational methods have become indispensable for exploring complex hypothesis spaces. AI-assisted discovery systems hold the promise of fundamentally changing the way scientific discovery is practiced~\cite{xin2025towards}.

Most existing AI-assisted discovery systems~\cite{aira,yakavets2025machine,erps2021accelerated} adopt paradigms rooted in machine learning, emphasizing predictive modeling on large datasets and evaluating success primarily through final performance metrics. While effective for many applications, these paradigms provide limited support for reasoning about the discovery process itself, including how hypotheses are explored, how experiments are prioritized, and how decisions are made under uncertainty. In particular, they rarely treat scientific discovery as a sequential, long-horizon decision process in which exploration strategies, resource allocation, and stopping decisions must be jointly optimized over time. Unlocking the full potential of AI in discovery, therefore, requires moving beyond performance-centric automation toward frameworks that explicitly support exploration, decision-making, and adaptation throughout the discovery trajectory.

Recent advances in large language models (LLMs)~\cite{openai2023gpt,bai2023qwen,guo2025deepseek} have significantly expanded the potential of AI systems in scientific research. Improvements in reasoning, multimodal understanding, and autonomous tool use now enable LLM-based systems to integrate planning, execution, and feedback within unified workflows. As a result, these systems can support key components of scientific discovery, including hypothesis generation, experiment design, and iterative refinement across domains~\cite{huang2022towards,ferrag2025llm}.
Building on these capabilities, prior work has progressively enhanced individual aspects of the scientific workflow. Early efforts demonstrated that LLMs can extract actionable scientific knowledge to guide experiments~\cite{lin2023evolutionary,angelopoulos2024transforming} and answer domain-specific professional questions~\cite{alampara2025probing,polak2024extracting,dagdelen2024structured}. Subsequent research extended these capabilities to cross-stage planning~\cite{biomni}, automated hypothesis generation~\cite{wang2025discovery,aira,baek2025researchagent}, multimodal knowledge integration~\cite{steyaert2023multimodal,gao2025chemical}, and automated experimentation across scientific domains. These advances have led to a new generation of scientific discovery systems, ranging from AI research assistants to fully autonomous workflows capable of executing end-to-end benchmarks~\cite{king2004functional,mandal2025evaluating,ai_co_scientist}.

While these systems demonstrate the feasibility of system-level automation in scientific workflows, most existing approaches focus primarily on improving individual stages of the discovery process or optimizing final outcomes. Even recent LLM-driven optimization methods~\cite{liao2025llm4eo,jiang2025agenticsciml,kochnev2025optuna}, which infer evolutionary patterns and synthesize update rules, operate largely at the level of execution or local improvement.
However, scientific discovery is inherently a sequential and strategic process that unfolds over time. It requires reasoning not only about isolated steps, but about the structure and quality of the entire exploration trajectory. Current systems rarely treat discovery as a trajectory-level decision-making problem. Consequently, they do not explicitly address fundamental challenges such as determining optimal stopping points, designing efficient exploration paths, or evaluating the structural effectiveness of discovery strategies over time.

\begin{figure*}[hbtp]
    \centering
    \includegraphics[width=1.0\textwidth]{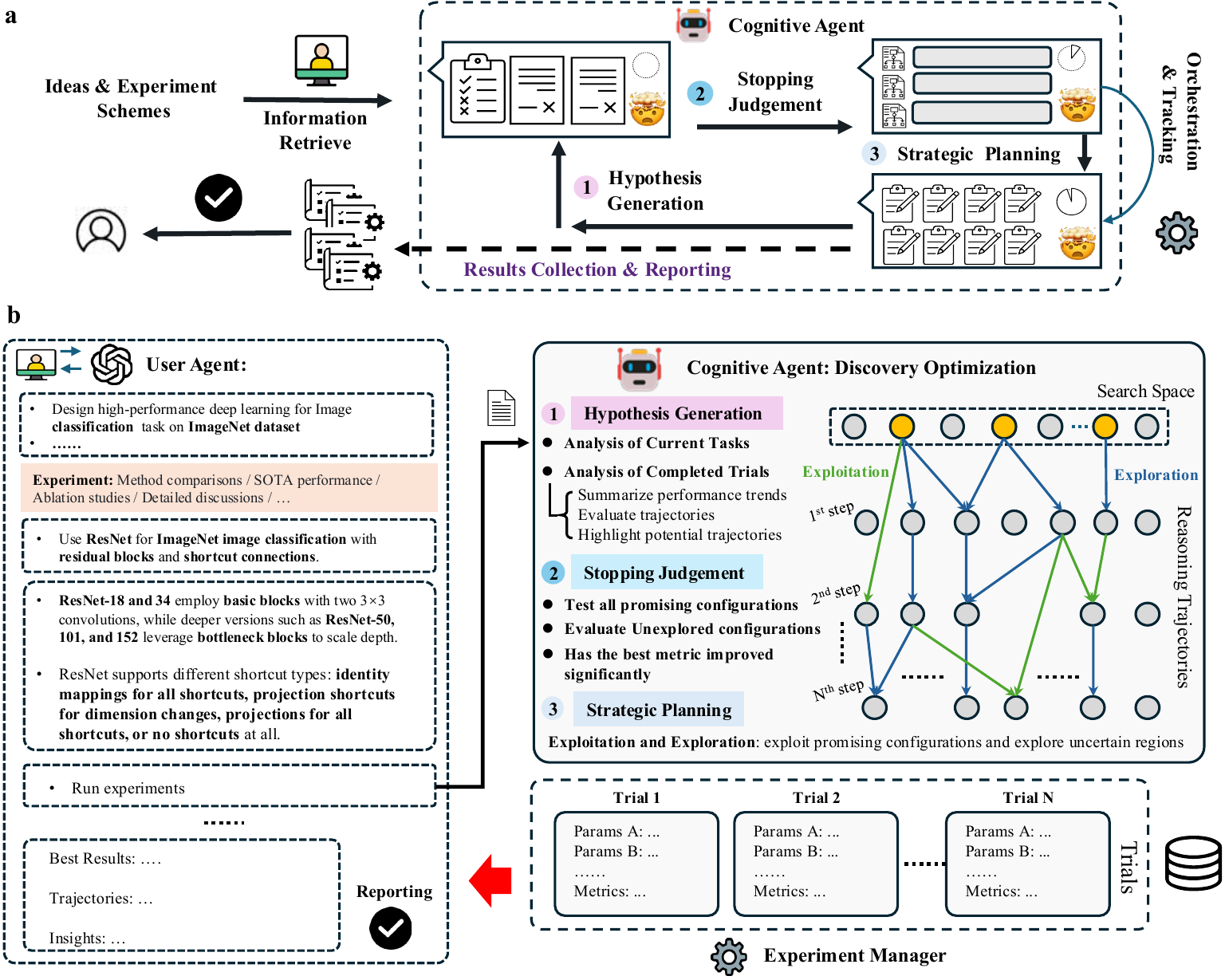}
    \caption{\textbf{SelfAI Framework for Automated Scientific Experimentation.} \textbf{a}, Holistic architecture of the multi-agent system, which transforms various experiments in the research process into a structured workflow. \textbf{b}, User intents, comprising ideas and experiment schemes, are transformed into structured configurations via a predefined prompt. These inputs are processed through successive stages: hypothesis generation, strategic planning, trial execution, and result collection.}
    \label{fig:flowchart}
\end{figure*}

Addressing these limitations requires a shift from execution-oriented autonomy toward cognitively oriented autonomy, in which discovery systems reason about exploration strategies, evaluate efficiency-diversity trade-offs, and adaptively decide when to stop. Motivated by this perspective, we introduce SelfAI, a self-directed, multi-agent-enabled discovery system that treats scientific discovery as a trajectory-level decision-making process. We integrate human intent, exploration reasoning, and experimental execution into a unified closed-loop framework. These are built with three modules and external tools (Fig.~\ref{fig:flowchart}a), enabling efficiency-diversity exploration in long-horizon scientific workflows. To evaluate discovery quality, we introduce two novel complementary metrics, Score (Discovery Efficiency) and $\text{AUP}_D$ (Area Under the Performance Diversity). Score aggregates, across tasks, the normalized improvement over the search space together with penalties for discovering good configurations late and for stopping far from the best-found point. $\text{AUP}_D$ explicitly encodes how broadly a solver explores the search space by summarizing the performance-diversity tradeoffs across the entire trajectory, enabling detailed analysis of exploration behavior and stopping decisions in long-term searches. Together, these metrics measure quantitative assessments of exploration behavior, reasoning structure, and stopping decisions in long-term autonomous experiments.

As illustrated in Fig.~\ref{fig:flowchart}b, SelfAI operates as a closed-loop discovery system that unifies research intent, trajectory-aware reasoning, and a diverse set of executive experiment tools to support various scientific problems and long-horizon scientific discovery. The workflow begins with a \textbf{User Agent}, which translates high-level research objectives and exploratory questions into standardized experiment configurations. These structured configurations provide a consistent interface across heterogeneous environments, enabling reliable comparison and reuse of results. \textbf{Cognitive Agent} evaluates trial outcomes, reasons over accumulated discovery trajectories, and adaptively revises exploration strategies over time. Experimental execution is coordinated by \textbf{Experiment Manager}, which autonomously manages resource allocation, environment provisioning, checkpoint-based execution, and comprehensive experiment logging. Through continual feedback between reasoning and execution, the Cognitive Agent and Experiment Manager jointly assess the expected utility of continued exploration and apply adaptive stopping decisions to terminate unproductive search paths. In this way, SelfAI enables sustained improvement through long-horizon, trajectory-driven discovery.

In summary, SelfAI provides a self-directed framework that treats scientific discovery as a strategic, trajectory-driven process. By iteratively integrating intent specification, trajectory-aware reasoning, and adaptive stopping, SelfAI enables principled allocation of experimental resources while preserving exploration diversity under realistic computational and experimental constraints. Across a broad range of scientific discovery tasks, SelfAI consistently achieves efficient, diverse, and high-quality exploration in heterogeneous and large-scale search spaces. These results demonstrate that trajectory-aware reasoning and adaptive stopping allow discovery systems to reduce redundant experimentation, identify diminishing returns early, and redirect effort toward more informative regions of the search space. More broadly, this work highlights the role of LLM-guided decision-making in organizing scientific exploration under explicit efficiency-diversity trade-offs. 

\begin{figure*}
    \centering
    \includegraphics[width=1.0\linewidth]{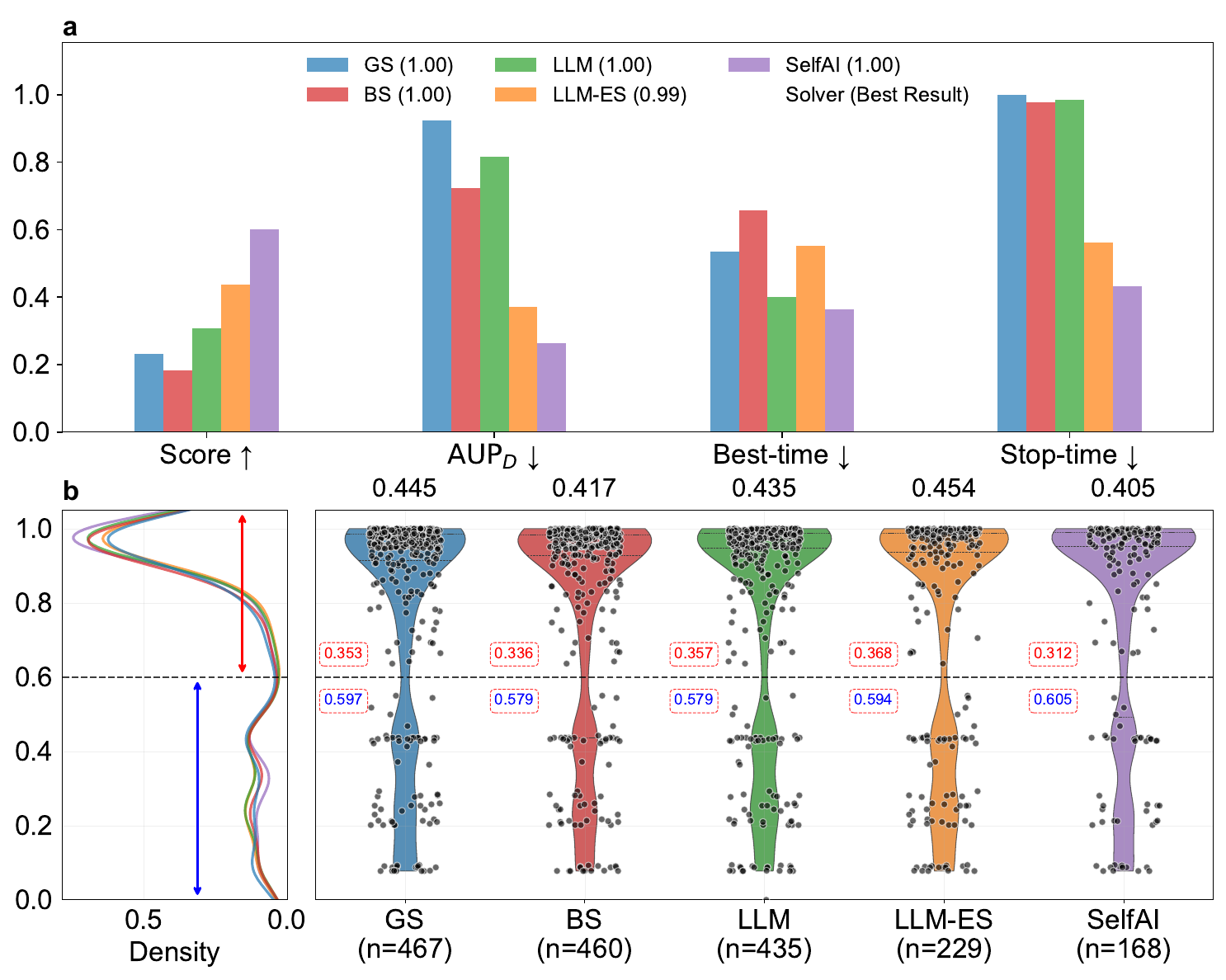}
    \caption{\textbf{Performance comparison and efficiency-diversity trade-offs across solvers.} \textbf{a}, Normalized performance of five solvers (GS, BS, LLM, LLM-ES, and SelfAI) evaluated by Score (higher is better), $\text{AUP}_D$, Best-time, and Stop-time (lower is better). Bars indicate solver-level averages, with values in parentheses denoting the best-achieved or optimal reference performance. \textbf{b}, Outcome distributions across solvers. Left: kernel density estimates illustrating score concentration and convergence tendencies. Right: violin plots with overlaid individual trials (sample size n shown), capturing performance variability. The dashed horizontal line denotes the evaluation threshold separating low- and high-performance regimes; annotated statistics (red/blue) highlight solver behavior above and below this threshold, where broader exploration facilitates rapid escape from low-performance regions and tighter distributions in high-performance regions enable localized refinement.}
    \label{fig:comparison}
\end{figure*}

\section{Results}
\label{sec:results}

\subsection{SelfAI: Long-horizon scientific discovery}

SelfAI is a trajectory-aware paradigm for long-horizon scientific discovery (Fig.~\ref{fig:flowchart}). It treats accumulated experimental outcomes as structured evidence within an evolving discovery trajectory, from which exploration decisions, hypothesis refinement, and stopping judgments are iteratively and autonomously derived. SelfAI builds on recent LLM-based optimization and agentic frameworks that interleave reasoning with trial proposals but typically emphasize short-horizon improvement or final performance. Across experiments, SelfAI consistently synthesizes accumulated information, analyzes experimental trajectories, evaluates performance trends, and prioritizes promising regions of the search space.

\subsection{Evaluation of long-horizon scientific discovery}

To characterize efficiency-diversity trade-offs in long-horizon exploration, we propose four metrics: Score, $\text{AUP}_D$, $t_{\text{best}}$, and $t_{\text{stop}}$ (Fig.~\ref{fig:comparison}a), the details of which are described in Sect.~\ref{sec:evaluation_reasoning}. These metrics not only reflect final performance but also explicitly capture when high-quality solutions are discovered, how broadly the search space is explored, and how efficiently exploration is terminated. Score aggregates normalized improvement while penalizing both delayed discovery of optimal configurations and prolonged exploration after performance plateaus. $\text{AUP}_D$ (Area Under the Performance-Diversity Curve) quantifies the diversity of explored high-quality solutions and summarizes how performance gains are distributed along the discovery trajectory. The metrics $t_{\text{best}}$ and $t_{\text{stop}}$ denote the normalized time first to identify the best result and the time at which exploration terminates, respectively. An effective solver minimizes both quantities, achieving early discovery while avoiding unnecessary trials. In this framework, effective exploration strategies attain a high Score with a relatively low $\text{AUP}_D$, indicating efficient discovery with minimal wasted exploration. Detailed per-task comparisons of Score, $\text{AUP}_D$, $t_{\text{best}}$, and $t_{\text{stop}}$ are provided in the Supplementary Materials~\ref{sec:supp_details_benchmark}.

\subsection{SelfAI reorganizes exploration dynamics across scientific domains}
We evaluate long-horizon scientific discovery using a benchmark constructed from real-world scientific discovery tasks (Materials and Methods) and assess solver performance under these efficiency-diversity trade-offs (Fig.~\ref{fig:comparison}). Using the same underlying language model (GPT-4o) across all solvers, we observe that while most methods are eventually able to identify high-performing configurations, their exploration dynamics differ substantially. As shown in Fig.~\ref{fig:comparison}a, SelfAI consistently outperforms LLM and LLM-ES baselines on $\text{AUP}_D$, Best-Time, and Stop-Time, indicating reduced redundant experimentation and more effective allocation of trials toward informative regions of the search space. Distributional analyses (Fig.~\ref{fig:comparison}b) further reveal that SelfAI reallocates experimental effort away from diminishing-return regions earlier while maintaining sufficient coverage of uncertain regions. Across the benchmark, for low-performance regimes, SelfAI obtains higher variances, conducts fewer trials to escape low-performance regimes, and concentrates effort in promising regions with lower variances. Finally, SelfAI achieves highly efficient exploration with 168 runs by terminating exploration near the best-achieved configuration under fixed trial budgets. Benchmark construction details and additional analyses are provided in Fig.~\ref{fig:LCBench_analysis}.

\subsubsection{Case studies}
\label{sec:case_studies}
We present representative case studies to illustrate how trajectory-aware reasoning and adaptive stopping manifest in concrete scientific discovery scenarios, illustrating how different solvers organize exploration trajectories over time, allocate experimental effort, and determine when further experimentation becomes uninformative. In addition, we illustrate that trajectory-level metrics capture discovery efficiency, diversity, and how reasoning limitations can emerge in structured optimization settings. Additional task-specific analyses, including 12 tasks, are provided in Supplementary Sect.~\ref{sec:supp_details_benchmark}.

{\noindent\bf Machine learning.} We first evaluate SelfAI on the Boston housing price prediction task~\cite{boston} using a random forest regression model. As shown in Fig.~\ref{fig:bar_score_aup} and Supplementary Table~\ref{tab:Boston}, nearly all solvers achieve a similar and high best result ($\approx 0.841$), indicating that differences in final predictive accuracy are minimal. Small- and medium-sized models outperform other solvers, but while large models DeepSeek-R1-70B and Llama3.3-70B demonstrated consistent and sustained exploration capabilities, their stopping behavior was less efficient, leading to longer search trajectories and delayed convergence. These results illustrate the importance of trajectory-aware reasoning and adaptive stopping in practical scientific optimization, as larger-scale models may exhibit either overly diffuse exploration or premature stopping and are not always aligned with task-specific objectives. Further examples are presented below in this section.

{\noindent\bf Computer vision.}
Fig.~\ref{fig:siren_surface} illustrates these trajectory-level dynamics in the hyperparameter search landscape for image segmentation using SIREN (see also Supplementary Figs.~\ref{fig:siren_surface1}-\ref{fig:siren_surface3}). The smoothed response surface reveals several obvious peaks, indicating multiple competing high-performance regions. Starting from identical initial points, the tree-based Bayesian optimizer follows a spiral-like trajectory that underutilizes promising regions and exhibits limited global coverage, frequently converging to suboptimal local minima. In contrast, LLM-based solvers infer trends from accumulated evaluations and adapt exploration accordingly, resulting in broader coverage of the search space and more reliable identification of globally competitive configurations. Importantly, these solvers also monitor trajectory-level progress and terminate exploration once improvements plateau. This behavior directly translates into higher scores and lower $\text{AUP}_D$, demonstrating how trajectory-aware metrics capture both discovery efficiency and controlled diversity beyond final performance alone. 

{\noindent\bf Medical image analysis.} We introduce medical image segmentation using nnU-Net~\cite{nnunet}, exploring performance optimization and best practices in medical image segmentation across 19 representative CNN-, Transformer-, and Mamba-based models. These results provide controlled settings for evaluating long-horizon optimization behavior. As shown in Fig.~\ref{fig:bar_score_aup} and Supplementary Tables~\ref{tab:MIABench}-\ref{tab:nnUnet}, trajectory-aware metrics reveal substantial differences in solver behavior that are not apparent from final segmentation accuracy alone. In particular, on the BraTS~\cite {brats} and BTCV~\cite {btcv} datasets, small- and medium-sized models achieve higher scores and lower $\text{AUP}_D$, reflecting their ability to rapidly identify high-performing configurations and terminate exploration. Trajectory-aware metrics concentrate exploration earlier in high-potential regions and avoid prolonged exploration once performance gains saturate. Conversely, solvers with comparable final performance but lower scores exhibit broader and more redundant exploration trajectories, leading to delayed stopping and reduced discovery efficiency.

{\noindent\bf Scientific computing.} Tensor decomposition~\cite{hackbusch2014numerical,verstraete2008matrix} is a fundamental tool in scientific computing for high-dimensional data analysis, with broad applications in data compression~\cite{li2025lossless}, computational acceleration~\cite{berezutskii2025tensor}, and multimodal data fusion~\cite{steyaert2023multimodal}. The associated optimization problem~\cite{tw} presents a structured landscape characterized by competing objectives between compression quality and computational efficiency. Effective solvers must therefore balance refinement of promising configurations with controlled exploration of alternative regimes. As shown in Supplementary Table~\ref{tab:TW}, GPT4-o3-mini performed well and ranked highly. However, Qwen2.5-72b may have over-relied on general reasoning ability and failed to balance mathematical knowledge and reasoning ability. Among the DeepSeek series models, although the DeepSeek-R1-7b model failed to find the optimal solution, its search strategy was acceptable, and its score was moderate. Although the DeepSeek-R1-70b and GPT-4o models have excellent search diversity and early stopping capabilities, their scores dropped. These results reveal that some large-scale models exhibit either overly diffuse exploration or premature stopping for task-specific objectives. This case illustrates that expressive reasoning capacity alone is insufficient for effective scientific discovery. Instead, successful optimization requires trajectory-level alignment between exploration strategy, stopping decisions, and problem structure. Supplementary analyses further document representative success and failure patterns across tasks (Supplementary Sect.~\ref{sec:supp_limitations}). In particular, hit rates for identifying globally optimal configurations vary substantially across models, with premature stopping limiting convergence in some cases. Moreover, increased model scale does not systematically improve long-horizon inference: larger models often sustain extended hypothesis exploration, leading to delayed commitment and greater performance variability under fixed experimental budgets. In contrast, several medium-sized models exhibit more stable trajectory organization, reinforcing the role of disciplined inference and adaptive stopping in efficient scientific discovery.

{\noindent\bf Result summary} Across 12 heterogeneous tasks and language-model backbones (Supplementary Sect. ~\ref{sec:supp_details_benchmark}), these results demonstrate that SelfAI yields consistent trajectory-level benefits. SelfAI does not merely improve convergence accuracy but fundamentally reorganizes exploration dynamics in long-horizon scientific discovery.  By reallocating experimental effort away from diminishing-return regions toward structurally informative areas of the search space, SelfAI implements a policy-adaptive exploration strategy that aligns with identifiable principles of long-horizon scientific discovery.

\subsection{Emergent principles of long-horizon scientific discovery}
\begin{figure*}[t]
    \centering
    \begin{tabular}{c}
        \includegraphics[width=1.0\textwidth]{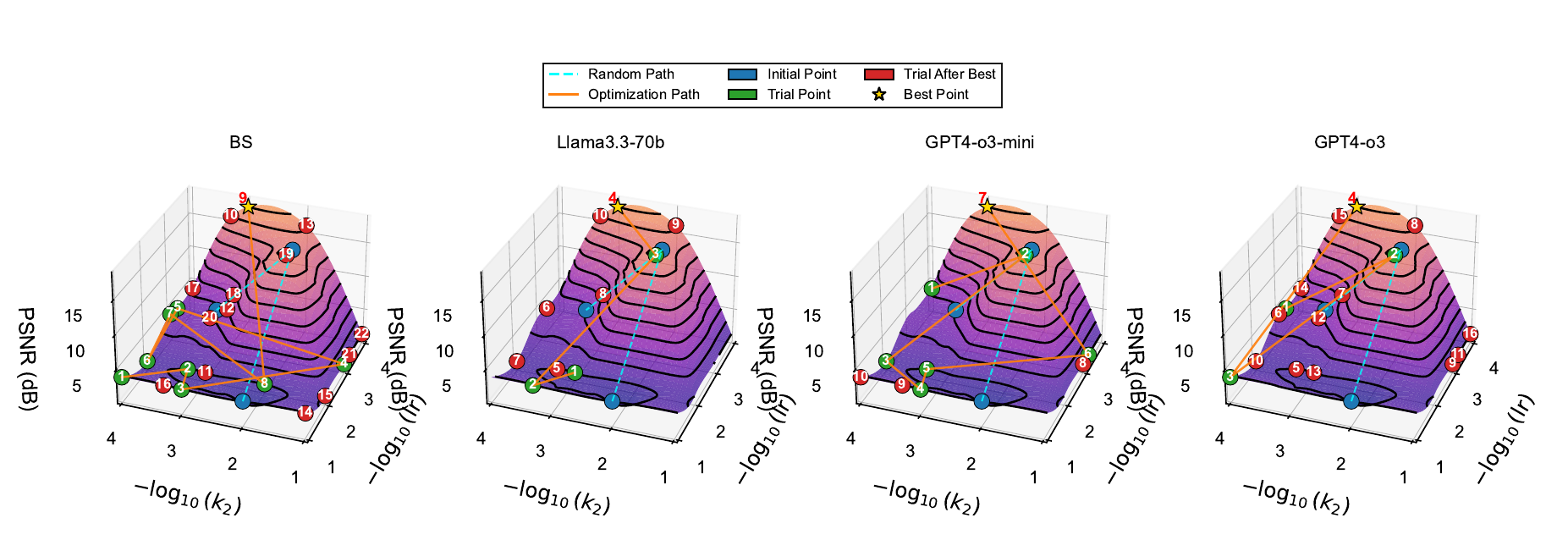} \\
        \includegraphics[width=1.0\textwidth]{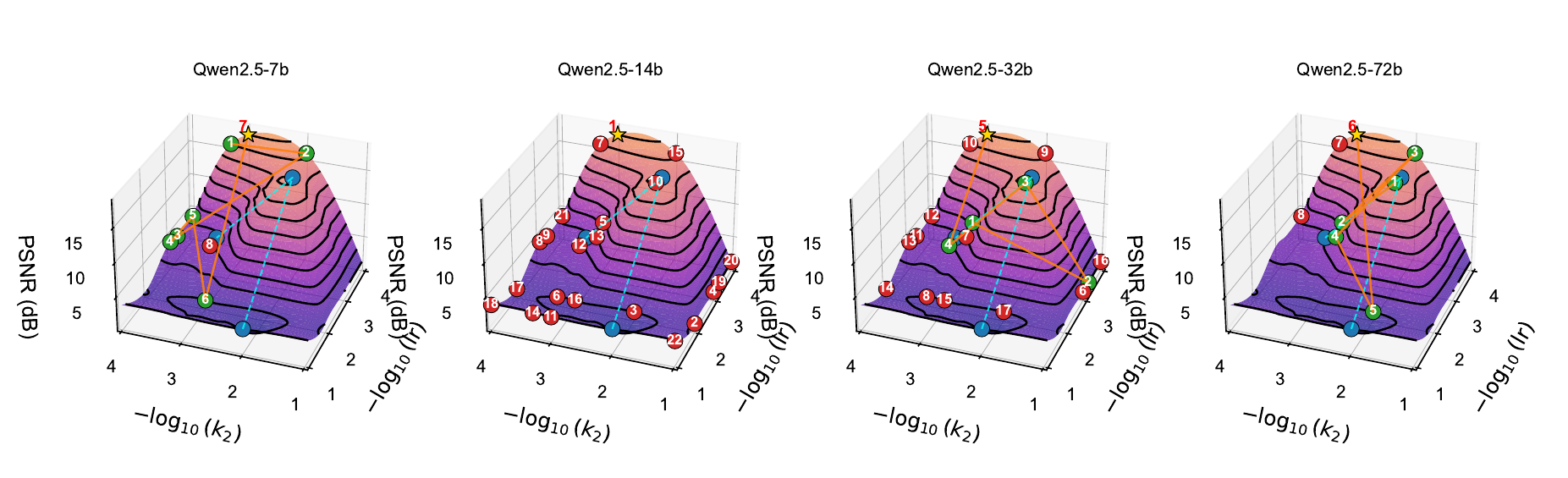} \\
        \includegraphics[width=1.0\textwidth]{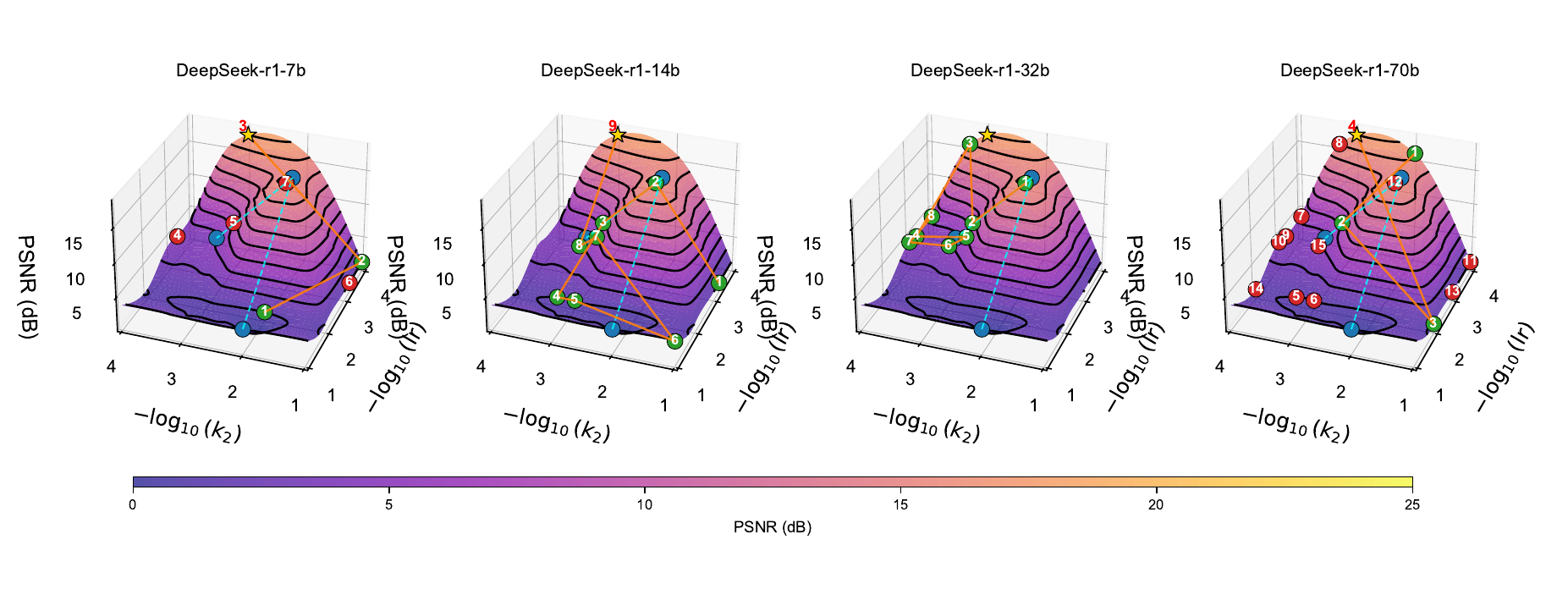} \\
    \end{tabular}
    \caption{Illustration of the optimized trajectory for the SIREN method for image segmentation. Green points are suggested points before reaching the optimal points. Red points are redundant suggestions when reaching out to the optimal points and failing to stop trials. The $\star$ is the optimal point. We show the serialization recommendations provided by LLM through the labeled numbers.}
    \label{fig:siren_surface}
\end{figure*}

We further investigate whether long-horizon scientific discovery exhibits consistent, task-agnostic principles when guided by trajectory-aware reasoning. Fig.~\ref{fig:siren_surface} illustrates these trajectory-level dynamics in the hyperparameter search landscape for image segmentation using SIREN (Sect.~\ref{sec:case_studies}). From principles of long-horizon scientific discovery, trajectory-level dynamics follow a characteristic three-stage trajectory: rapid escape from low-efficiency regions, early concentration of effort in high-potential areas, and termination near the optimal configuration. This structured progression highlights the central role of strategic trial allocation and adaptive stopping in the efficiency of long-horizon exploration. 

\begin{figure}
    \centering
    \includegraphics[width=0.9\linewidth]{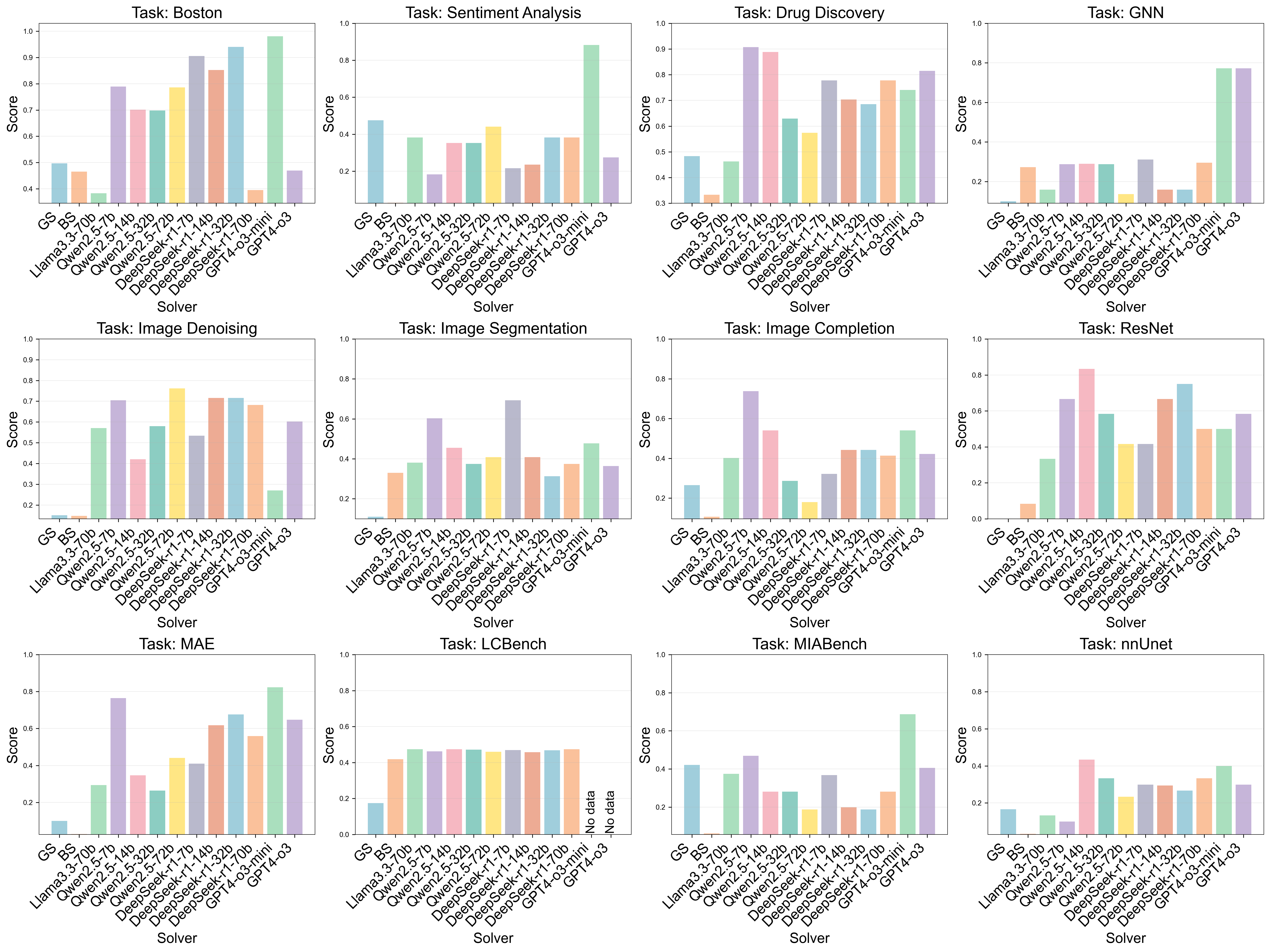} \\
    \includegraphics[width=0.9\linewidth]{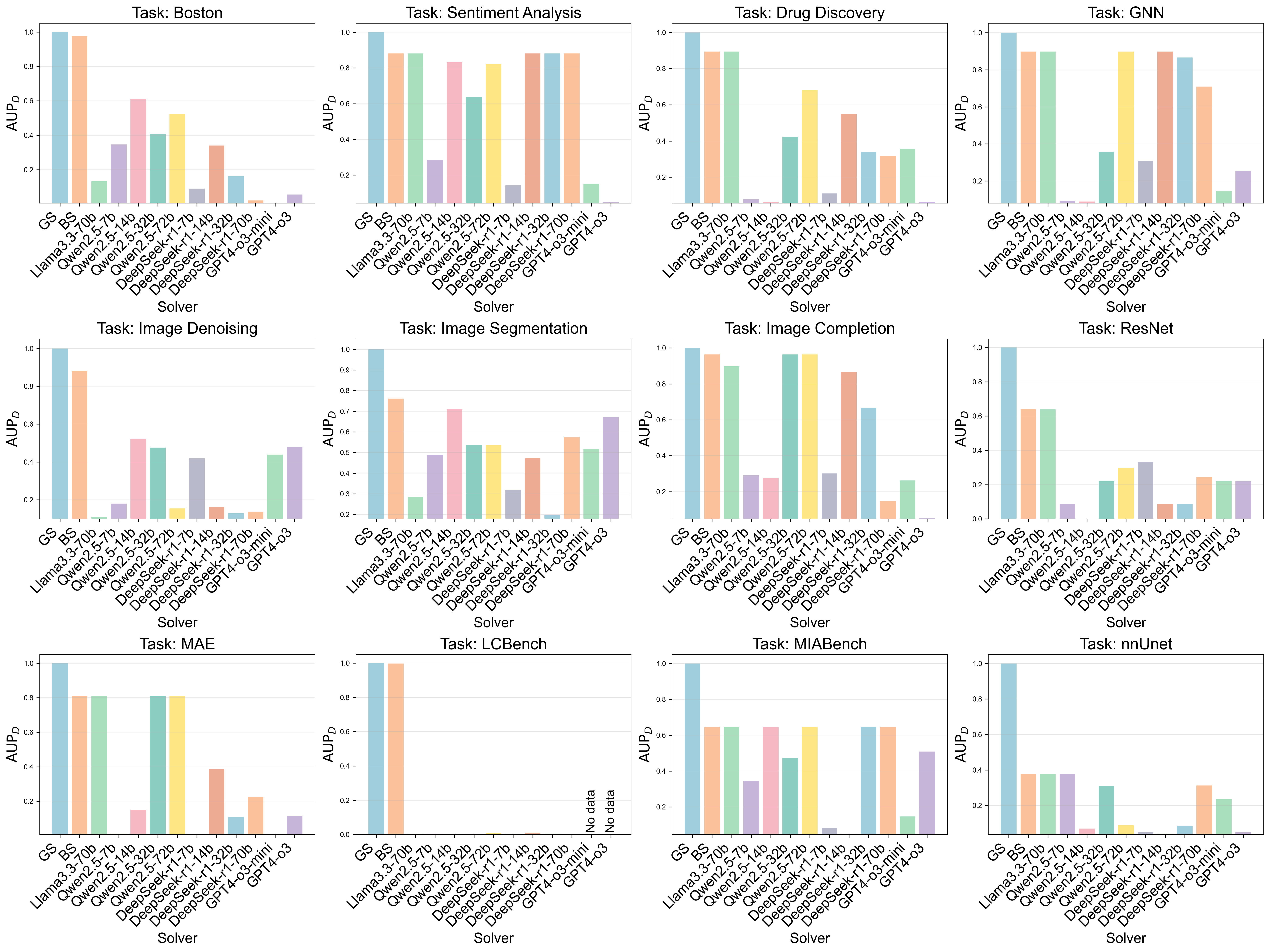}
    \caption{The top three rows present the Score of all solvers across different tasks, which assess the best stopping performance of SelfAI and baselines using different LLMs. The bottom three rows present the Diverse Metrics ($\text{AUP}_D$) of all solvers across different tasks, evaluating the diversity of solution trajectories. }
    \label{fig:bar_score_aup}
\end{figure}

Aggregated performance across tasks and domains reveals additional trajectory-level regularities (Fig.~\ref{fig:bar_score_aup}). Despite task-dependent variations in relative rankings (Supplementary Fig.~\ref{fig:rank_heatmap}), SelfAI consistently achieves high Scores while maintaining relatively low $\text{AUP}_D$ values across heterogeneous tasks and language-model backbones. These results indicate that efficient discovery trajectories can be organized according to shared principles that are not tied to any single task or domain. 

Model scale alone does not systematically improve long-horizon discovery behavior. Larger models often sustain prolonged exploration of alternative hypotheses, leading to delayed commitment and reduced adaptability under fixed experimental budgets. In contrast, small and mid-sized models often exhibit more stable cumulative progress, earlier concentration on informative regions, and more reliable stopping behavior. These observations show that baseline LLM solvers relying on exhaustive exploration without principled stopping often exhaust their search budgets (Fig.~\ref{fig:flowchart}). In contrast, SelfAI-enabled solvers achieve substantially higher hit rates while terminating exploration significantly earlier. Notably, final performance is often near saturation across methods, with the best results differing only marginally. Nevertheless, significant differences exist in performance and diversity metrics, revealing substantial differences in identifying high-quality configurations and effectiveness in avoiding redundant experiments.

\section{Discussion}
Autonomous scientific discovery systems (ASDS) have been widely proposed as a means to accelerate research by delegating large-scale experimentation to artificial intelligence. However, most existing agent frameworks and ASDS benchmarks primarily emphasize candidate generation and execution, offering limited support for managing long-horizon discovery processes. In particular, they provide little explicit reasoning about how search strategies should evolve over time, when exploration should be terminated, or how experimental effort should be redistributed as evidence accumulates.

This work addresses these limitations by reframing scientific discovery as a trajectory-level decision-making problem. By integrating structured intent interpretation, trajectory-aware reasoning, and adaptive stopping within a unified framework, SelfAI enables explicit reasoning over accumulated experimental evidence. Across diverse benchmarks, SelfAI consistently reallocates computational resources toward informative regions of the search space and terminates exploration under diminishing returns, reshaping discovery dynamics beyond what is achievable through endpoint-focused optimization alone.

The newly proposed trajectory-aware evaluation metrics are crucial for quantifying this advantage. Metrics such as Score and $\text{AUP}_D$ reveal substantial differences in discovery efficiency, redundancy, and stopping behavior that are obscured by conventional endpoint-based evaluation. While many solvers ultimately converge to similar final performance, trajectory-level metrics expose how quickly high-quality configurations are identified and how effectively redundant experimentation is avoided. These measurements highlight structural limitations of classical optimization methods, including grid search and Bayesian optimization, which lack adaptive reasoning mechanisms for long-horizon exploration and principled termination.

SelfAI's trajectory-aware reasoning enables long-term scientific exploration by integrating experimental history. While our benchmark results show that LLM-based agents generally exhibit broad exploration behavior and consistent final performance, they can also fall into the trap of excessively long hypothesis exploration times, model execution delays, and greater performance fluctuations. Although some small- to medium-sized models, guided by structured reasoning and adaptive stopping, demonstrate more stable trajectory organization and earlier convergence, optimal stopping decisions can still fail in highly complex environments. Premature stopping can still lead to the inability to identify the global optimum, while over-exploration can occur under extreme uncertainty (see Supplementary Sect.~\ref{sec:supp_limitations}). To enhance the long-term scientific exploration capabilities of LLM agents, dynamic knowledge integration and contextual learning capabilities can be integrated to overcome the inherent knowledge bottlenecks of LLMs.

Beyond benchmark performance, the proposed framework has broader implications for human-AI collaboration in scientific research. By separating high-level scientific intent from low-level experimental execution, SelfAI provides a structured interface through which human expertise can guide large-scale automated exploration. This design promotes transparency and reproducibility by ensuring that exploration policies remain anchored to explicit research objectives. More broadly, it enables a shift from using AI solely as an optimization tool toward studying scientific discovery itself as an object of inquiry—allowing researchers to analyze, compare, and refine discovery strategies under controlled conditions.

Finally, LLM-based agents enable explicit modeling of long-horizon scientific discovery, shifting discovery systems from task-specific performance optimization toward continuous discovery processes for broader scientific domains. Considering a wide range of scientific fields, including physical and materials sciences, there is broad interest but limited expertise in general-purpose LLM agents. SelfAI can interleave LLM agents from different application domains to support the design of entirely new systems. Moreover, this framework can be integrated with automated toolchains, such as code generation and more powerful inference models, to enable smarter and more adaptive discovery workflows. Overall, this work opens new opportunities for scientific discovery by establishing a pathway toward continuous, long-horizon exploration.
\section{Materials and Methods}

\subsection{Scientific discovery benchmark}

Existing machine learning benchmarks, such as MLE-bench~\cite{chan2024mle}, are primarily designed to accelerate the engineering of learning systems by optimizing performance under constrained computational budgets. In contrast, we introduce a scientific discovery benchmark specifically tailored to long-horizon discovery, formulated as a trajectory-driven decision-making process. Rather than emphasizing final performance, the benchmark requires solvers to engage in scientific reasoning by analyzing experimental outcomes, interpreting accumulated observations, and adaptively revising search strategies to uncover functional relationships, structural patterns, or optimal experimental configurations.

The benchmark comprises 12 tasks spanning six scientific categories, including scientific computing, machine learning, computer vision, medical image analysis, and drug discovery (Table~\ref{tab:task_list}). All benchmark data are collected from real experimental runs, with several tasks corresponding to configurations reported in published literature. The tasks cover both discrete and continuous search spaces and range from low- to high-dimensional settings. To simulate realistic user-system interaction, each task provides background information to the LLM to support intent interpretation. Most datasets are generated through systematic grid-based exploration, while LCBench and the Chagas EP20 drug discovery dataset are constructed from Bayesian optimization trajectories. LCBench is a widely used AutoML benchmark~\cite{zimmer2021auto}. All tasks are organized as YAML specifications containing metadata, task definitions, concise problem descriptions, search spaces, and trial records. The benchmark will be publicly released at \url{https://github.com/XiaoXiao-Woo/SelfAI}.
\subsection{Compared methods and evaluation setup}
We evaluate SelfAI across all 12 benchmark tasks and compare it with classical optimization methods, including grid search and the Tree Parzen Estimator (TPE) optimizer~\cite{optuna}, referred to as BS. To assess robustness across model families, we instantiate SelfAI with different language-model backbones, including OpenAI-o3~\cite{openai2023gpt}, Llama3.3~\cite{llama}, Qwen2.5~\cite{bai2023qwen}, and DeepSeek-R1~\cite{guo2025deepseek}. All methods are evaluated under identical computational settings. Reasoning traces are retained for analysis and reproducibility. For all trials, the random seed is fixed, and the temperature is set to zero. The detailed settings of the benchmark are available in the Supplementary Materials.

\begin{table}[h!]
\centering
\setlength{\tabcolsep}{4pt}
\renewcommand{\arraystretch}{1.2}
\caption{List of tasks in SelfAI with different hyperparameters for multiple tasks and datasets.}
\begin{tabular}{@{}lllcccc@{}}
\toprule
\textbf{Category} & \textbf{Method} & \textbf{Task} & \textbf{Dim} & \textbf{Count} & \textbf{Ref.} \\ 
\midrule
\multirow{1}{*}{Scientific Computing} & Tensor Network & Image Completion & 3 & 64 & \cite{tw} \\  
\hline
\multirow{3}{*}{Machine Learning} 
 & Random Forest & Pricing Prediction & 5 & 162 & \cite{boston} \\
 & LSTM & Sentiment Analysis & 2 & 20 & \cite{sentiment_analysis} \\ 
 & GraphSAGE & Node Classification & 22 & 25 & \cite{yan2023unreal} \\ 
 \hline
\multirow{5}{*}{Computer Vision} & SIREN & Image Denoising & 2 & 25 & \cite{sitzmann2020implicit} \\
 & SIREN & Image Segmentation & 2 & 25 & \cite{sitzmann2020implicit} \\
 & ResNet & Image Classification & 4 & 9 & \cite{resnet} \\ 
 & MAE & Image Classification & 2 & 20 & \cite{mae} \\ 
 & FashionMnist-NN & Image Classification & 5 & 2000 & \cite{zimmer2021auto} \\ 
\hline
\multirow{2}{*}{Medical Image Analysis} & nnUnet & BraTS~\cite{brats} & 3 & 18 & \cite{nnunet} \\
 & nnUnet-revisited & BTCV~\cite{btcv} & 5 & 19 & \cite{nnunet_revisit} \\
Drug Discovery & DNN & Bioactivity Prediction & 4 & 30 & \cite{korotcov2017comparison} \\
\bottomrule
\end{tabular}
\label{tab:task_list}
\end{table}
\subsection{SelfAI system}
\label{sec:method}
The SelfAI system operates in an integrated loop, comprising three core agents: User Agent, Cognitive Agent, and Experiment manager, which together constitute the overall experimental pipeline in Fig.~\ref{fig:flowchart}. It consists of three components: a User Agent for intent formalization, a Cognitive Agent for iterative decision-making, and an Experiment Manager for experiment orchestration and management. Before illustrating the details of SelfAI, we first situate SelfAI within the landscape of existing AI-assisted research frameworks by examining their functional coverage (Table~\ref{tab:functions_selfai}).

\begin{table*}[bhtp]
\centering
\setlength{\tabcolsep}{4.5pt}
\renewcommand\arraystretch{1.2}
\caption{Comparison of SelfAI with related AI research frameworks and benchmarks across system-level, agent-specific, and task-specific capabilities.}
\resizebox{1.0\textwidth}{!}{
\begin{tabular}{l|cccccccc}
\toprule
Functions & Ours & Code LLaMA~\cite{kochnev2025optuna} & MLGym~\cite{MLgym} & AI Co-Scientist~\cite{ai_co_scientist} & AIRA~\cite{aira} & MLAgentBench~\cite{mlagentbench} & Optuna~\cite{optuna} \\
\midrule
Interactive Research & \cmark & \xmark & \cmark & \xmark & \xmark & \cmark & \cmark \\ 
Flexible Artifacts & \cmark & \xmark & \cmark & \cmark & \xmark & \cmark & \xmark \\
Privacy and Security & \cmark & \xmark & \xmark & \xmark & \xmark & \xmark & \cmark \\
\midrule
Trajectory Analysis & \cmark & \xmark & \xmark & \xmark & \xmark & \xmark & \xmark \\
Hypothesis Generation & \cmark & \xmark & \cmark & \cmark & \cmark & \xmark & \xmark \\
Strategic Planning & \cmark & \xmark & \cmark & \xmark & \xmark & \xmark & \xmark \\
Causal Inference & \cmark & \xmark & \cmark & \xmark & \xmark & \xmark & \xmark \\
Adaptive Learning & \cmark & \cmark & \cmark & \xmark & \xmark & \xmark & \xmark \\
\midrule
Job Scheduling & \cmark & \xmark & \cmark & \xmark & \xmark & \xmark & \cmark \\
Checkpoint Management & \cmark & \xmark & \xmark & \xmark & \xmark & \xmark & \xmark \\
Experiment Tracking & \cmark & \xmark & \xmark & \xmark & \xmark & \xmark & \cmark(*) \\
Zero-Code Parallelization & \cmark & \xmark & \xmark & \xmark & \xmark & \xmark & \xmark \\
\midrule
Hypothesis Optimization& \cmark & \cmark(*) & \xmark & \xmark & \xmark & \xmark & \cmark \\
Self-Evaluation & \cmark & \xmark & \xmark & \xmark & \xmark & \xmark & \xmark \\
Benchmark Suite & \cmark & \xmark & \xmark & \xmark & \xmark & \cmark & \xmark \\
\bottomrule
\end{tabular}}
\begin{tablenotes}
\small
\item Note: The comparison is structured in four blocks: (1) System-level functions, (2) Cognitive functions primarily handled by the Cognitive Agent, (3) Execution functions managed by the Experiment Manager, and (4) Performance in optimization tasks. \cmark(*) denotes basic support.
\end{tablenotes}
\label{tab:functions_selfai}
\end{table*}

\subsubsection{User agent}
As illustrated in Fig.~\ref{fig:flowchart}b, User Agent serves as the user interface through which high-level research intent is translated into structured, machine-readable experimental specifications. Its primary function is to formalize natural-language objectives provided by users into a standardized configuration that defines the experimental task for the SelfAI system. For example, user requests such as “Design high-performance deep learning for image classification task on the ImageNet dataset” or “Identify the most influential method for image classification” are reformulated into precise experimental specifications. Specifically, User Agent maps user-provided descriptions of research goals, constraints, and preferences into a task-specific YAML configuration. This configuration explicitly specifies basic ideas, optimization objectives, the experimental search space, trial budget, and historical trial data (as shown in Supplementary Sect.~\ref{sec:supp_user_agent}). The resulting YAML file serves as the sole interface between user intent and downstream system components, including Cognitive Agent and Experiment Manager. Consequently, User Agent encodes the user’s research intent into a fixed configuration schema, which is implemented via a predefined prompt template. All interactions are restricted to the generation and refinement of the structured configuration, and users do not directly manipulate experimental parameters or outcomes once the configuration is finalized. 

\subsubsection{Cognitive agent}
\label{sec:cognitive}
Cognitive Agent performs iterative decision-making within the SelfAI framework by operating on structured experimental inputs and conducting trajectory-level analysis across multiple iterations. To support long-horizon optimization, Cognitive Agent operates in three functional stages: hypothesis generation, strategic planning, and stopping judgment. Hypothesis generation identifies promising regions of the search space based on accumulated experimental outcomes. Strategic planning translates these hypotheses into concrete experimental proposals that balance refinement of high-performing configurations with exploration of under-sampled or uncertain regions. Stopping judgment evaluates accumulated experimental evidence to determine whether continued experimentation is warranted under the current exploration strategy. At each iteration, Cognitive Agent receives a task-specific experimental configuration from User Agent, along with accumulated experimental histories and performance metrics from Experiment Manager. These inputs are provided in a structured format and define the current state of the search process. Using the accumulated evidence, Cognitive Agent evaluates experimental trajectories, assesses observed performance trends, and estimates coverage of the explored search space. Based on this analysis, it generates candidate experimental configurations for subsequent exploration. In addition to proposing new experiments, the Cognitive Agent produces a stopping decision based on trajectory-level evaluation of completed trials, including comparison with the initial configuration, assessment of diminishing returns in recent performance improvements, and estimation of the potential value of remaining unexplored regions. The outputs of Cognitive Agent consist of candidate experimental configurations for the next iteration and a stopping signal indicating whether the current exploration process should be terminated. Detailed implementation, including the prompt templates used for hypothesis generation, strategic planning, and stopping judgment, is provided in the Supplementary Materials~\ref{sec:supp_cognitive}.

\subsubsection{Experiment manager}
\label{sec:exp_management}
Experiment Manager is responsible for experiment orchestration and recovery, including resource management, task allocation, and progress tracking (Fig.~\ref{fig:flowchart}b). These capabilities enable efficient coordination of multi-instance parallel optimization, maximize resource utilization, and enhance training robustness.

1) {\bf Resource Management}. Experiment Manager monitors resource consumption and dynamically allocates available GPU, TPU, and memory resources. This granular allocation optimizes workload distribution across computing units and ensures stable execution of all trials.

2) {\bf Fault recovery and Checkpoint Reconnection}. In case of system interruptions or suboptimal model performance, Experiment Manager reports failures to Cognitive agent. Experiment Manager performs preliminary diagnostics, identifies potential issues, adjusts training parameters, and resumes training from the latest checkpoint.

3) {\bf Multi-Instance Parallel Optimization}. SelfAI instantiates each user program to run across diverse physical environments, independent of the target program framework. Experiment Manager coordinates multi-instance parallel training, synchronizes execution, and concurrently tests various configurations, thereby shortening overall training time and improving generalization across datasets and model parameters. For each parallel experiment, Experiment Manager identifies and supplies the necessary runtime parameters, ensuring experiments are conducted under the same environment optimized by Cognitive agent.

\subsection{
Evaluation for reasoning trajectories}
\label{sec:evaluation_reasoning}
While the reasoning process of LLMs in problem-solving often generates a diverse range of discrete insights and multiple potential chains of thought, this diversity, though valuable for exploration and exploitation, can pose a challenge to coherent reasoning evaluation across various experimentation and discovery phases. We propose a systematic evaluation metric that captures both the diversity of reasoning perspectives and the overall coherence of the reasoning process, ensuring a more robust and comprehensive assessment of reasoning capabilities.

\noindent
\textbf{Optimal Stopping Criteria}
In this work, we collected the best value point and the stop point from trials. Based on Optimal Stopping Criteria~\cite{hill2009knowing}, we can define the following measure formulas,
\begin{align}
    & \text{Gain} = \frac{1}{N} \sum^{N-1}_{i=0} \frac{v^*_i - v_{i,\min}}{v_{i,\max} - v_{i, \min}}
     \\
    & t_{\text{best}} = \frac{1}{N} \sum^{N-1}_{i=0}t^{\text{best}}_i = \frac{1}{N} \sum^{N-1}_{i=0} \frac{m_i}{M_i} \\
    & t_{\text{stop}} = \frac{1}{N} \sum^{N-1}_{i=0}t^{\text{stop}}_i = \frac{1}{N} \sum^{N-1}_{i=0} \frac{n_i}{M_i}
\end{align}
and
where $N$ is the number of tasks. For the $i$-th task, $M_i$ is the number of completed trials. $m_i$ is the best value point index. $t_i^{\text{best}}$ is the cost of the best value. In addition, we set the stop point index, $n_i$, and $t_i^{\text{stop}}$ is better when $n_i$ is closer to the best value point index. $S_i$ means the binarized value.

To obtain a comprehensive measure, we combine the last three measures, i.e., $\text{Rel}$, $t_{\text{best}}$, and $t_{\text{stop}}$, where we utilize  $\text{Rel}$ in underestimated penalty, then $t_{\text{best}}$ and $t_{\text{stop}}$ are the time penalty ($P_\text{best}$ and $P_\text{stop}$). Thus, the total penalty is
\begin{align}
    P_{\text{total}} &= \frac{t_i^{\text{stop}} + t_i^{\text{best}}}{2}
\end{align}

Finally, the score is denoted as
\begin{align}
    & \text{Score} = \frac{1}{N}\sum^{N-1}_{i=0} \text{Gain} \cdot (1-P_{total})
\end{align}
\noindent \textbf{Best Approximation/Candidate} In \cite{MLgym}, performance profiles and the AUP aim to measure available rates across $\texttt{m}$ tasks, where all performance metrics are threshold $\tau$, performance profiles ($ \rho_\texttt{m}(\tau)$) are computed as thresholds in all metrics (sorted by ascending) in the current task.
\begin{equation}
    \text{AUP}_\texttt{m} = \int^{\tau_{\max}}_1 \rho_\texttt{m}(\tau) d\tau
\end{equation}
It is noted that the above performance profile and $\text{AUP}_\texttt{m}$ score cannot measure the diversity of reasoning. Therefore, we rewrite the performance profile and AUP score:

First, the performance profile is defined in all completed trials $M_i$ in $i$-th task and the overall search space $\mathcal{H}$, as follows 
\begin{equation}
    r_i =
    \begin{cases}
         \frac{\max{\{v_k:\, k \in \mathcal{H}\}}}{v_i},\, \text{ascend}  
         \\
         \\
         \frac{v_i}{\min{\{v_k:\, k \in \mathcal{H}\}}},\, \text{descend}
    \end{cases}
\end{equation}
where ascend/descend denotes that the value is larger/smaller and the performance is better. For all trials, $\tau$ is the set of all obtained $r_i$ values. Then, we consider all completed trials $M_i$ in $i$-th task, 
\begin{align}
    & \rho_i(\tau) = \begin{cases}
         |\{k \in M_i: r_k >= \tau \}|, &\text{ascend} \\
        |\{k \in M_i: r_k <= \tau \}|, & \text{descend}
    \end{cases}
\end{align}
which captures how many evaluated configurations exceed a given performance threshold $\tau$. $\rho_i(\tau)$ is the cumulative distribution curve of the trajectory. 

The area term aggregates the overall concentration of strong configurations along the performance axis:
\begin{equation}
    A = \frac{1}{N} \sum^{N-1}_{i=0} \int_{\tau_{\min}}^{\tau_{\max}} \rho_{i}(\tau)d\tau.
\end{equation}
To capture the temporal asymmetry of discovery, we compute the centroid
\begin{equation}
    G = \frac{1}{A} \int_{\tau_{\min}}^{\tau_{\max}} x \cdot \rho_i(\tau) \, d\tau,
\end{equation}
and define the skewness
\begin{equation}
    S = \int_{\tau_{\min}}^{\tau_{\max}} \left( \frac{x}{G} \right)^3 \rho(x) \, d\tau.
\end{equation}
Since $S$ may be unbounded and may take both positive (left-skewed) and negative (right-skewed) values, we normalize it via a reference skewness value $S_\text{base}$ from the GS method and a smooth monotonic mapping:
\begin{align}
    & S’ = 1-\frac{S-S_\text{base}}{S_\text{base}}, \\
    & S' = \frac{\tanh(S) + 1}{2}, S' \in (0, 1).
\end{align}
Finally, the Area Under the Performance-Diversity curve ($\text{AUP}_D$) is defined as
\begin{equation}
    \text{AUP}_D = A / S',
\end{equation}
where trajectories that exhibit earlier concentration of high-performing configurations obtain larger and thus smaller $\text{AUP}_D$ values, whereas trajectories that concentrate improvements later yield smaller and therefore larger values.

\bibliography{0-nature}

@article{openai2023gpt,
  title={Gpt-4 technical report. arxiv 2303.08774},
  author={OpenAI, R},
  journal={View in Article},
  volume={2},
  number={5},
  year={2023}
}

@article{guo2025deepseek,
  title={DeepSeek-R1: Incentivizing Reasoning Capability in LLMs via Reinforcement Learning},
  author={Guo, Daya and Yang, Dejian and Zhang, Haowei and Song, Junxiao and Zhang, Ruoyu and Xu, Runxin and Zhu, Qihao and Ma, Shirong and Wang, Peiyi and Bi, Xiao and others},
  journal={arXiv preprint arXiv:2501.12948},
  year={2025}
}

@article{bai2023qwen,
  title={Qwen technical report},
  author={Bai, Jinze and Bai, Shuai and Chu, Yunfei and Cui, Zeyu and Dang, Kai and Deng, Xiaodong and Fan, Yang and Ge, Wenbin and Han, Yu and Huang, Fei and others},
  journal={arXiv preprint arXiv:2309.16609},
  year={2023}
}

@misc{boston,
    author = {Brad Huang},
    title = {[03/24] Boston Housing Dataset},
    year = {2020},
    howpublished = {\url{https://kaggle.com/competitions/2403-boston-housing-dataset}},
    note = {Kaggle}
}

@inproceedings{zhao2021graphsmote,
  title={Graphsmote: Imbalanced node classification on graphs with graph neural networks},
  author={Zhao, Tianxiang and Zhang, Xiang and Wang, Suhang},
  booktitle={Proceedings of the 14th ACM international conference on web search and data mining},
  pages={833--841},
  year={2021}
}

@article{yan2023unreal,
  title={Unreal: Unlabeled nodes retrieval and labeling for heavily-imbalanced node classification},
  author={Yan, Liang and Zhang, Shengzhong and Li, Bisheng and Zhou, Min and Huang, Zengfeng},
  journal={arXiv preprint arXiv:2303.10371},
  year={2023}
}

@article{sitzmann2020implicit,
  title={Implicit neural representations with periodic activation functions},
  author={Sitzmann, Vincent and Martel, Julien and Bergman, Alexander and Lindell, David and Wetzstein, Gordon},
  journal={Advances in neural information processing systems},
  volume={33},
  pages={7462--7473},
  year={2020}
}

@article{MLgym,
  title={{{MLGym}}: A New Framework and Benchmark for Advancing AI Research Agents},
  author={Nathani, Deepak and Madaan, Lovish and Roberts, Nicholas and Bashlykov, Nikolay and Menon, Ajay and Moens, Vincent and Budhiraja, Amar and Magka, Despoina and Vorotilov, Vladislav and Chaurasia, Gaurav and others},
  journal={arXiv preprint arXiv:2502.14499},
  year={2025}
}

@inproceedings{mae,
  title={Masked autoencoders are scalable vision learners},
  author={He, Kaiming and Chen, Xinlei and Xie, Saining and Li, Yanghao and Doll{\'a}r, Piotr and Girshick, Ross},
  booktitle={Proceedings of the IEEE/CVF conference on computer vision and pattern recognition},
  pages={16000--16009},
  year={2022}
}

@article{zimmer2021auto,
  title={Auto-pytorch: Multi-fidelity metalearning for efficient and robust autodl},
  author={Zimmer, Lucas and Lindauer, Marius and Hutter, Frank},
  journal={IEEE transactions on pattern analysis and machine intelligence},
  volume={43},
  number={9},
  pages={3079--3090},
  year={2021},
  publisher={IEEE}
}

@article{nnunet,
  title={{{nnU-Net}}: a self-configuring method for deep learning-based biomedical image segmentation},
  author={Isensee, Fabian and Jaeger, Paul F and Kohl, Simon AA and Petersen, Jens and Maier-Hein, Klaus H},
  journal={Nature methods},
  volume={18},
  number={2},
  pages={203--211},
  year={2021},
  publisher={Nature Publishing Group}
}

@inproceedings{nnunet_revisit,
  title={nnu-net revisited: A call for rigorous validation in 3d medical image segmentation},
  author={Isensee, Fabian and Wald, Tassilo and Ulrich, Constantin and Baumgartner, Michael and Roy, Saikat and Maier-Hein, Klaus and Jaeger, Paul F},
  booktitle={International Conference on Medical Image Computing and Computer-Assisted Intervention},
  pages={488--498},
  year={2024},
  organization={Springer}
}

@article{hackbusch2014numerical,
  title={Numerical tensor calculus},
  author={Hackbusch, Wolfgang},
  journal={Acta numerica},
  volume={23},
  pages={651--742},
  year={2014},
  publisher={Cambridge University Press}
}

@article{verstraete2008matrix,
  title={Matrix product states, projected entangled pair states, and variational renormalization group methods for quantum spin systems},
  author={Verstraete, Frank and Murg, Valentin and Cirac, J Ignacio},
  journal={Advances in Physics},
  volume={57},
  number={2},
  pages={143--224},
  year={2008},
  publisher={Taylor \& Francis}
}

@inproceedings{tw,
  title={Tensor Wheel Decomposition and Its Tensor Completion Application},
  author={Wu, Zhong-Cheng and Huang, Ting-Zhu and Deng, Liang-Jian and Dou, Hong-Xia and Meng, Deyu},
  booktitle={Advances in Neural Information Processing Systems},
year={2022}
}

@article{word2vec,
  title={Efficient estimation of word representations in vector space},
  author={Mikolov, Tomas and Chen, Kai and Corrado, Greg and Dean, Jeffrey},
  journal={arXiv preprint arXiv:1301.3781},
  year={2013}
}

@article{sentiment_analysis,
  title={Performance analysis of hyperparameters on a sentiment analysis model},
  author={Kandhro, Irfan Ali and Jumani, Sahar Zafar and Ali, Fayyaz and Shaikh, Zubair Uddin and Arain, Muhammad Arshad and Shaikh, Aftab Ahmed},
  journal={Engineering, Technology \& Applied Science Research},
  volume={10},
  number={4},
  pages={6016--6020},
  year={2020}
}

@article{biomni,
  title={Biomni: A General-Purpose Biomedical AI Agent},
  author={Huang, Kexin and Zhang, Serena and Wang, Hanchen and Qu, Yuanhao and Lu, Yingzhou and Roohani, Yusuf and Li, Ryan and Qiu, Lin and Zhang, Junze and Di, Yin and others},
  journal={bioRxiv},
  pages={2025--05},
  year={2025},
  publisher={Cold Spring Harbor Laboratory}
}

@article{mlagentbench,
  title={Mlagentbench: Evaluating language agents on machine learning experimentation},
  author={Huang, Qian and Vora, Jian and Liang, Percy and Leskovec, Jure},
  journal={arXiv preprint arXiv:2310.03302},
  year={2023}
}

@article{hill2009knowing,
  title={Knowing when to stop: How to gamble if you must-the mathematics of optimal stopping},
  author={Hill, Theodore P},
  journal={American Scientist},
  volume={97},
  number={2},
  pages={126--133},
  year={2009},
  publisher={Sigma Xi, The Scientific Research Society}
}

@inproceedings{resnet,
  title={Deep residual learning for image recognition},
  author={He, Kaiming and Zhang, Xiangyu and Ren, Shaoqing and Sun, Jian},
  booktitle={Proceedings of the IEEE/CVF conference on computer vision and pattern recognition},
  pages={770--778},
  year={2016}
}

@inproceedings{brahmavar2024generating,
  title={Generating novel leads for drug discovery using LLMs with logical feedback},
  author={Brahmavar, Shreyas Bhat and Srinivasan, Ashwin and Dash, Tirtharaj and Krishnan, Sowmya Ramaswamy and Vig, Lovekesh and Roy, Arijit and Aduri, Raviprasad},
  booktitle={Proceedings of the AAAI conference on artificial intelligence},
  volume={38},
  number={1},
  pages={21--29},
  year={2024}
}

@article{bongini2021molecular,
  title={Molecular generative graph neural networks for drug discovery},
  author={Bongini, Pietro and Bianchini, Monica and Scarselli, Franco},
  journal={Neurocomputing},
  volume={450},
  pages={242--252},
  year={2021},
  publisher={Elsevier}
}

@article{alphafold3,
  title={Accurate structure prediction of biomolecular interactions with AlphaFold 3},
  author={Abramson, Josh and Adler, Jonas and Dunger, Jack and Evans, Richard and Green, Tim and Pritzel, Alexander and Ronneberger, Olaf and Willmore, Lindsay and Ballard, Andrew J and Bambrick, Joshua and others},
  journal={Nature},
  volume={630},
  number={8016},
  pages={493--500},
  year={2024},
  publisher={Nature Publishing Group UK London}
}

@article{korotcov2017comparison,
  title={Comparison of deep learning with multiple machine learning methods and metrics using diverse drug discovery data sets},
  author={Korotcov, Alexandru and Tkachenko, Valery and Russo, Daniel P and Ekins, Sean},
  journal={Molecular pharmaceutics},
  volume={14},
  number={12},
  pages={4462--4475},
  year={2017},
  publisher={ACS Publications}
}

@article{Chagas,
  title={Machine learning models and pathway genome data base for Trypanosoma cruzi drug discovery},
  author={Ekins, Sean and Lage de Siqueira-Neto, Jair and McCall, Laura-Isobel and Sarker, Malabika and Yadav, Maneesh and Ponder, Elizabeth L and Kallel, E Adam and Kellar, Danielle and Chen, Steven and Arkin, Michelle and others},
  journal={PLoS neglected tropical diseases},
  volume={9},
  number={6},
  pages={e0003878},
  year={2015},
  publisher={Public Library of Science San Francisco, CA USA}
}

@article{llama,
  title={Llama: Open and efficient foundation language models},
  author={Touvron, Hugo and Lavril, Thibaut and Izacard, Gautier and Martinet, Xavier and Lachaux, Marie-Anne and Lacroix, Timoth{\'e}e and Rozi{\`e}re, Baptiste and Goyal, Naman and Hambro, Eric and Azhar, Faisal and others},
  journal={arXiv preprint arXiv:2302.13971},
  year={2023}
}

@article{ma2024u,
  title={U-mamba: Enhancing long-range dependency for biomedical image segmentation},
  author={Ma, Jun and Li, Feifei and Wang, Bo},
  journal={arXiv preprint arXiv:2401.04722},
  year={2024}
}

@inproceedings{ronneberger2015u,
  title={U-net: Convolutional networks for biomedical image segmentation},
  author={Ronneberger, Olaf and Fischer, Philipp and Brox, Thomas},
  booktitle={International Conference on Medical image computing and computer-assisted intervention},
  pages={234--241},
  year={2015},
  organization={Springer}
}

@article{vaswani2017attention,
  title={Attention is all you need},
  author={Vaswani, Ashish and Shazeer, Noam and Parmar, Niki and Uszkoreit, Jakob and Jones, Llion and Gomez, Aidan N and Kaiser, {\L}ukasz and Polosukhin, Illia},
  journal={Advances in neural information processing systems},
  volume={30},
  year={2017}
}

@article{gu2023mamba,
  title={Mamba: Linear-time sequence modeling with selective state spaces},
  author={Gu, Albert and Dao, Tri},
  journal={arXiv preprint arXiv:2312.00752},
  year={2023}
}

@inproceedings{tang2022self,
  title={Self-supervised pre-training of swin transformers for 3d medical image analysis},
  author={Tang, Yucheng and Yang, Dong and Li, Wenqi and Roth, Holger R and Landman, Bennett and Xu, Daguang and Nath, Vishwesh and Hatamizadeh, Ali},
  booktitle={Proceedings of the IEEE/CVF conference on computer vision and pattern recognition},
  pages={20730--20740},
  year={2022}
}

@inproceedings{roy2023mednext,
  title={Mednext: transformer-driven scaling of convnets for medical image segmentation},
  author={Roy, Saikat and Koehler, Gregor and Ulrich, Constantin and Baumgartner, Michael and Petersen, Jens and Isensee, Fabian and Jaeger, Paul F and Maier-Hein, Klaus H},
  booktitle={International Conference on Medical Image Computing and Computer-Assisted Intervention},
  pages={405--415},
  year={2023},
  organization={Springer}
}

@article{fliri2005biological,
  title={Biological spectra analysis: linking biological activity profiles to molecular structure},
  author={Fliri, Anton F and Loging, William T and Thadeio, Peter F and Volkmann, Robert A},
  journal={Proceedings of the National Academy of Sciences},
  volume={102},
  number={2},
  pages={261--266},
  year={2005},
  publisher={National Academy of Sciences}
}

@article{ferrag2025llm,
  title={From llm reasoning to autonomous ai agents: A comprehensive review},
  author={Ferrag, Mohamed Amine and Tihanyi, Norbert and Debbah, Merouane},
  journal={arXiv preprint arXiv:2504.19678},
  year={2025}
}

@article{huang2022towards,
  title={Towards reasoning in large language models: A survey},
  author={Huang, Jie and Chang, Kevin Chen-Chuan},
  journal={arXiv preprint arXiv:2212.10403},
  year={2022}
}

@inproceedings{optuna,
	title={Optuna: A Next-generation Hyperparameter Optimization Framework},
	author={Akiba, Takuya and Sano, Shotaro and Yanase, Toshihiko and Ohta, Takeru and Koyama, Masanori},
	booktitle={Proceedings of the 25th {ACM} {SIGKDD} International Conference on Knowledge Discovery and Data Mining},
	year={2019}
}

@inproceedings{btcv,
  title={Miccai multi-atlas labeling beyond the cranial vault--workshop and challenge},
  author={Landman, Bennett and Xu, Zhoubing and Igelsias, J and Styner, Martin and Langerak, T and Klein, Arno},
  booktitle={Proc. MICCAI Multi-Atlas Labeling Beyond Cranial Vault—Workshop Challenge},
  volume={5},
  pages={12},
  year={2015}
}

@misc{brats,
      title={The ASNR-MICCAI Brain Tumor Segmentation (BraTS) Challenge 2023: Intracranial Meningioma}, 
      author={Dominic LaBella and Maruf Adewole and Michelle Alonso-Basanta and Talissa Altes and Syed Muhammad Anwar and Ujjwal Baid and Timothy Bergquist and Radhika Bhalerao and Sully Chen and Verena Chung and Gian-Marco Conte and Farouk Dako and James Eddy and Ivan Ezhov and Devon Godfrey and Fathi Hilal and Ariana Familiar and Keyvan Farahani and Juan Eugenio Iglesias and Zhifan Jiang and Elaine Johanson and Anahita Fathi Kazerooni and Collin Kent and John Kirkpatrick and Florian Kofler and Koen Van Leemput and Hongwei Bran Li and Xinyang Liu and Aria Mahtabfar and Shan McBurney-Lin and Ryan McLean and Zeke Meier and Ahmed W Moawad and John Mongan and Pierre Nedelec and Maxence Pajot and Marie Piraud and Arif Rashid and Zachary Reitman and Russell Takeshi Shinohara and Yury Velichko and Chunhao Wang and Pranav Warman and Walter Wiggins and Mariam Aboian and Jake Albrecht and Udunna Anazodo and Spyridon Bakas and Adam Flanders and Anastasia Janas and Goldey Khanna and Marius George Linguraru and Bjoern Menze and Ayman Nada and Andreas M Rauschecker and Jeff Rudie and Nourel Hoda Tahon and Javier Villanueva-Meyer and Benedikt Wiestler and Evan Calabrese},
      year={2023},
      eprint={2305.07642},
      archivePrefix={arXiv},
      primaryClass={cs.CV}
}

@article{ai_co_scientist,
  title={Towards an AI co-scientist},
  author={Gottweis, Juraj and Weng, Wei-Hung and Daryin, Alexander and Tu, Tao and Palepu, Anil and Sirkovic, Petar and Myaskovsky, Artiom and Weissenberger, Felix and Rong, Keran and Tanno, Ryutaro and others},
  journal={arXiv preprint arXiv:2502.18864},
  year={2025}
}

@article{aira,
  title={AI Research Agents for Machine Learning: Search, Exploration, and Generalization in MLE-bench},
  author={Toledo, Edan and Hambardzumyan, Karen and Josifoski, Martin and Hazra, Rishi and Baldwin, Nicolas and Audran-Reiss, Alexis and Kuchnik, Michael and Magka, Despoina and Jiang, Minqi and Lupidi, Alisia Maria and others},
  journal={arXiv preprint arXiv:2507.02554},
  year={2025}
}

@article{kochnev2025optuna,
  title={Optuna vs Code Llama: Are LLMs a New Paradigm for Hyperparameter Tuning?},
  author={Kochnev, Roman and Goodarzi, Arash Torabi and Bentyn, Zofia Antonina and Ignatov, Dmitry and Timofte, Radu},
  journal={arXiv preprint arXiv:2504.06006},
  year={2025}
}

@article{berezutskii2025tensor,
  title={Tensor networks for quantum computing},
  author={Berezutskii, Aleksandr and Liu, Minzhao and Acharya, Atithi and Ellerbrock, Roman and Gray, Johnnie and Haghshenas, Reza and He, Zichang and Khan, Abid and Kuzmin, Viacheslav and Lyakh, Dmitry and others},
  journal={Nature Reviews Physics},
  pages={1--13},
  year={2025},
  publisher={Nature Publishing Group UK London}
}

@article{li2025lossless,
  title={Lossless data compression by large models},
  author={Li, Ziguang and Huang, Chao and Wang, Xuliang and Hu, Haibo and Wyeth, Cole and Bu, Dongbo and Yu, Quan and Gao, Wen and Liu, Xingwu and Li, Ming},
  journal={Nature Machine Intelligence},
  pages={1--6},
  year={2025},
  publisher={Nature Publishing Group UK London}
}

@article{steyaert2023multimodal,
  title={Multimodal data fusion for cancer biomarker discovery with deep learning},
  author={Steyaert, Sandra and Pizurica, Marija and Nagaraj, Divya and Khandelwal, Priya and Hernandez-Boussard, Tina and Gentles, Andrew J and Gevaert, Olivier},
  journal={Nature machine intelligence},
  volume={5},
  number={4},
  pages={351--362},
  year={2023},
  publisher={Nature Publishing Group UK London}
}

@article{dagdelen2024structured,
  title={Structured information extraction from scientific text with large language models},
  author={Dagdelen, John and Dunn, Alexander and Lee, Sanghoon and Walker, Nicholas and Rosen, Andrew S and Ceder, Gerbrand and Persson, Kristin A and Jain, Anubhav},
  journal={Nature communications},
  volume={15},
  number={1},
  pages={1418},
  year={2024},
  publisher={Nature Publishing Group UK London}
}

@article{gao2025chemical,
  title={A chemical autonomous robotic platform for end-to-end synthesis of nanoparticles},
  author={Gao, Fan and Li, Hongqiang and Chen, Zhilong and Yi, Yunai and Nie, Shihao and Cheng, Zihao and Liu, Zeming and Guo, Yuanfang and Liu, Shumin and Qin, Qizhen and others},
  journal={Nature Communications},
  volume={16},
  number={1},
  pages={7558},
  year={2025},
  publisher={Nature Publishing Group UK London}
}

@article{mandal2025evaluating,
  title={Evaluating large language model agents for automation of atomic force microscopy},
  author={Mandal, Indrajeet and Soni, Jitendra and Zaki, Mohd and Smedskjaer, Morten M and Wondraczek, Katrin and Wondraczek, Lothar and Gosvami, Nitya Nand and Krishnan, NM Anoop},
  journal={Nature Communications},
  volume={16},
  number={1},
  pages={9104},
  year={2025},
  publisher={Nature Publishing Group UK London}
}

@article{alampara2025probing,
  title={Probing the limitations of multimodal language models for chemistry and materials research},
  author={Alampara, Nawaf and Schilling-Wilhelmi, Mara and R{\'\i}os-Garc{\'\i}a, Marti{\~n}o and Mandal, Indrajeet and Khetarpal, Pranav and Grover, Hargun Singh and Krishnan, NM Anoop and Jablonka, Kevin Maik},
  journal={Nature computational science},
  pages={1--10},
  year={2025},
  publisher={Nature Publishing Group US New York}
}

@article{polak2024extracting,
  title={Extracting accurate materials data from research papers with conversational language models and prompt engineering},
  author={Polak, Maciej P and Morgan, Dane},
  journal={Nature Communications},
  volume={15},
  number={1},
  pages={1569},
  year={2024},
  publisher={Nature Publishing Group UK London}
}

@article{king2004functional,
  title={Functional genomic hypothesis generation and experimentation by a robot scientist},
  author={King, Ross D and Whelan, Kenneth E and Jones, Ffion M and Reiser, Philip GK and Bryant, Christopher H and Muggleton, Stephen H and Kell, Douglas B and Oliver, Stephen G},
  journal={Nature},
  volume={427},
  number={6971},
  pages={247--252},
  year={2004},
  publisher={Nature Publishing Group UK Lo,ndon}
}

@article{chan2024mle,
  title={Mle-bench: Evaluating machine learning agents on machine learning engineering},
  author={Chan, Jun Shern and Chowdhury, Neil and Jaffe, Oliver and Aung, James and Sherburn, Dane and Mays, Evan and Starace, Giulio and Liu, Kevin and Maksin, Leon and Patwardhan, Tejal and others},
  journal={arXiv preprint arXiv:2410.07095},
  year={2024}
}

@article{liao2025llm4eo,
  title={LLM4EO: Large Language Model for Evolutionary Optimization in Flexible Job Shop Scheduling},
  author={Liao, Rongjie and Qiu, Junhao and Chen, Xin and Li, Xiaoping},
  journal={arXiv preprint arXiv:2511.16485},
  year={2025}
}

@inproceedings{baek2025researchagent,
  title={Researchagent: Iterative research idea generation over scientific literature with large language models},
  author={Baek, Jinheon and Jauhar, Sujay Kumar and Cucerzan, Silviu and Hwang, Sung Ju},
  booktitle={Proceedings of the 2025 Conference of the Nations of the Americas Chapter of the Association for Computational Linguistics: Human Language Technologies (Volume 1: Long Papers)},
  pages={6709--6738},
  year={2025}
}

@article{jiang2025agenticsciml,
  title={AgenticSciML: Collaborative Multi-Agent Systems for Emergent Discovery in Scientific Machine Learning},
  author={Jiang, Qile and Karniadakis, George},
  journal={arXiv preprint arXiv:2511.07262},
  year={2025}
}

@article{weininger1988smiles,
  title={SMILES, a chemical language and information system. 1. Introduction to methodology and encoding rules},
  author={Weininger, David},
  journal={Journal of chemical information and computer sciences},
  volume={28},
  number={1},
  pages={31-36},
  year={1988}
}

@article{xin2025towards,
  title={Towards agentic science for advancing scientific discovery},
  author={Xin, Hongliang and Kitchin, John R and Kulik, Heather J},
  journal={Nature Machine Intelligence},
  pages={1--3},
  year={2025},
  publisher={Nature Publishing Group UK London}
}

@article{bradshaw1983studying,
  title={Studying scientific discovery by computer simulation},
  author={Bradshaw, Gary F and Langley, Patrick W and Simon, Herbert A},
  journal={Science},
  volume={222},
  number={4627},
  pages={971--975},
  year={1983},
  publisher={American Association for the Advancement of Science}
}

@article{lyu2024alphafold2,
  title={AlphaFold2 structures guide prospective ligand discovery},
  author={Lyu, Jiankun and Kapolka, Nicholas and Gumpper, Ryan and Alon, Assaf and Wang, Liang and Jain, Manish K and Barros-{\'A}lvarez, Ximena and Sakamoto, Kensuke and Kim, Yoojoong and DiBerto, Jeffrey and others},
  journal={Science},
  volume={384},
  number={6702},
  pages={eadn6354},
  year={2024},
  publisher={American Association for the Advancement of Science}
}

@article{lupoiu2025multi,
  title={A multi-agentic framework for real-time, autonomous freeform metasurface design},
  author={Lupoiu, Robert and Shao, Yixuan and Dai, Tianxiang and Mao, Chenkai and Ed{\'e}e, Kofi and Fan, Jonathan A},
  journal={Science Advances},
  volume={11},
  number={44},
  pages={eadx8006},
  year={2025},
  publisher={American Association for the Advancement of Science}
}

@article{angelopoulos2024transforming,
  title={Transforming science labs into automated factories of discovery},
  author={Angelopoulos, Angelos and Cahoon, James F and Alterovitz, Ron},
  journal={Science Robotics},
  volume={9},
  number={95},
  pages={eadm6991},
  year={2024},
  publisher={American Association for the Advancement of Science}
}

@article{jiang2022artificial,
  title={An artificial intelligence enabled chemical synthesis robot for exploration and optimization of nanomaterials},
  author={Jiang, Yibin and Salley, Daniel and Sharma, Abhishek and Keenan, Graham and Mullin, Margaret and Cronin, Leroy},
  journal={Science advances},
  volume={8},
  number={40},
  pages={eabo2626},
  year={2022},
  publisher={American Association for the Advancement of Science}
}

@article{grisoni2021combining,
  title={Combining generative artificial intelligence and on-chip synthesis for de novo drug design},
  author={Grisoni, Francesca and Huisman, Berend JH and Button, Alexander L and Moret, Michael and Atz, Kenneth and Merk, Daniel and Schneider, Gisbert},
  journal={Science Advances},
  volume={7},
  number={24},
  pages={eabg3338},
  year={2021},
  publisher={American Association for the Advancement of Science}
}

@article{yakavets2025machine,
  title={Machine learning-assisted exploration of multidrug-drug administration regimens for organoid arrays},
  author={Yakavets, Ilya and Kheiri, Sina and Cruickshank, Jennifer and Hickman, Riley J and Rakhshani, Faeze and Aldeghi, Matteo and Rajaonson, Ella M and Young, Edmond WK and Aspuru-Guzik, Al{\'a}n and Cescon, David W and others},
  journal={Science Advances},
  volume={11},
  number={31},
  pages={eadt1851},
  year={2025},
  publisher={American Association for the Advancement of Science}
}

@article{kaiser2025large,
  title={Large language models for human-machine collaborative particle accelerator tuning through natural language},
  author={Kaiser, Jan and Lauscher, Anne and Eichler, Annika},
  journal={Science advances},
  volume={11},
  number={1},
  pages={eadr4173},
  year={2025},
  publisher={American Association for the Advancement of Science}
}

@article{wang2025discovery,
  title={Discovery of diverse and high-quality mRNA capping enzymes through a language model--enabled platform},
  author={Wang, Tianze and Qin, Bowen R and Li, Sihong and Wang, Zimo and Li, Xuejian and Jiang, Yuanxu and Qin, Chenrui and Ouyang, Qi and Lou, Chunbo and Qian, Long},
  journal={Science advances},
  volume={11},
  number={15},
  pages={eadt0402},
  year={2025},
  publisher={American Association for the Advancement of Science}
}

@article{nguyen2024sequence,
  title={Sequence modeling and design from molecular to genome scale with Evo},
  author={Nguyen, Eric and Poli, Michael and Durrant, Matthew G and Kang, Brian and Katrekar, Dhruva and Li, David B and Bartie, Liam J and Thomas, Armin W and King, Samuel H and Brixi, Garyk and others},
  journal={Science},
  volume={386},
  number={6723},
  pages={eado9336},
  year={2024},
  publisher={American Association for the Advancement of Science}
}

@article{lin2023evolutionary,
  title={Evolutionary-scale prediction of atomic-level protein structure with a language model},
  author={Lin, Zeming and Akin, Halil and Rao, Roshan and Hie, Brian and Zhu, Zhongkai and Lu, Wenting and Smetanin, Nikita and Verkuil, Robert and Kabeli, Ori and Shmueli, Yaniv and others},
  journal={Science},
  volume={379},
  number={6637},
  pages={1123--1130},
  year={2023},
  publisher={American Association for the Advancement of Science}
}

@article{erps2021accelerated,
  title={Accelerated discovery of 3D printing materials using data-driven multiobjective optimization},
  author={Erps, Timothy and Foshey, Michael and Lukovi{\'c}, Mina Konakovi{\'c} and Shou, Wan and Goetzke, Hanns Hagen and Dietsch, Herve and Stoll, Klaus and von Vacano, Bernhard and Matusik, Wojciech},
  journal={Science advances},
  volume={7},
  number={42},
  pages={eabf7435},
  year={2021},
  publisher={American Association for the Advancement of Science}
}

\clearpage
\begin{appendices}

\section{LLM Agents in SelfAI System for Scientific Discovery}
\label{sec:supp_method}

The SelfAI framework supports the complete lifecycle of AI systems, covering design, training, evaluation, and deployment. It integrates three core agents: the User Agent, the Cognitive Agent, and the Experiment Manager, and provides an integrated toolkit for autonomous model training, experiment orchestration, and advanced reasoning, enabling a scalable and adaptive workflow.

\subsection{Prompt of User Agent: Idea Interaction and Experiment Configuration}
\label{sec:supp_user_agent}

\begin{tcolorbox}[breakable,colback=orange!5!white, colframe=black, title=User Agent Prompt for Configuration Interaction]

\textbf{System:} ...

\textbf{User:}

\pkd{SUMMARIZED\_CONTENT}

Please fill out the content following the YAML format:

\begin{bg_yaml}
\begin{minipage}{1.0\linewidth}
\begin{lstlisting}[style=yamlstyle, language=Yaml]
# Template for Configuration Interaction 
# to convert Idea Interaction into YAML format
- role: system
  content:
    model: $modelName
    description: You are a $role specializing in 
                studying $taskName. Please provide 
                professional and detailed answers.
    task: $taskName
    basic_idea: $basic_idea
    search_space:
      $hyper_name1: []
      $hyper_name2: []
      ...
    link: $paperLink
    instrustion: Complete instructions under limited trials.
- role: user
  content:
    max_trials: $maxTrials
    trials: []
\end{lstlisting}
\end{minipage}
\end{bg_yaml}
\label{selfai_yaml}


\end{tcolorbox}


\subsection{Reasoning Prompt for Scientific Discovery}
\label{sec:supp_cognitive}

\subsubsection{Trajectory Analysis and Hypothesis Generation}
Cognitive agent begins with \textbf{Task 1: Analyze the Current Task}, where it meticulously interprets the prompt to establish the task's core objectives, constraints, and hyperparameters. This initial analysis is used to perform a structured trajectory evaluation, creating a coherent chain of thought that guides all subsequent strategy and exploration.

Next, in \textbf{Task 2: Analysis of Completed Trials}, the agent systematically evaluates previous outcomes to generate new hypotheses. It specifically analyzes the performance trends of individual parameters and their combinations to identify promising directions in the broader hyperparameter space and propose novel configurations. This deep trend analysis allows the agent to anticipate promising regions of the search space and detect valuable reasoning trajectories that might otherwise be overlooked.

\begin{tcolorbox}[breakable, colback=orange!5!white, colframe=black, title={Cognitive Prompt for Trajectory Analysis and Hypothesis Generation}]
\textbf{System:} 

\pkd{System\_part\_of\_configuration}

\textbf{User:}
Completed trials:

\pkd{completed\_trials}

Task 1: Analyze the current task

Understand current tasks, basic ideas, objectives, and hyperparameters.

Task 2: Analysis of Completed Trials

Step 1: Summarize performance metrics for completed trials.

Step 2: Evaluate performance trends for hyperparameters.

Step 3: Highlight promising hyperparameter combinations.
\end{tcolorbox}

\subsubsection{Stopping Judgement}
\label{sec:supp_stop_criterion}
We introduce an optimal stopping criterion to guide prompt designs that balance exploration and exploitation. During the optimal stopping judgment, Cognitive agent first determines whether the current performance metrics surpass those of the initial configuration and conducts an optimal stopping judgment. Building on the prior analysis of completed trials, the agent systematically evaluates stopping conditions to avoid testing all configurations. 

By using prompt instructions, the optimal stopping judgment evaluates all completed trials by leveraging insights from trial analysis, observed performance trends, and identified key findings. In practice, we implement this judgment process via a structured prompt template comprising two main tasks and explicit stopping criteria. The language model then analyzes the relationship between completed trials and unexplored areas. The prompt is organized as follows:

\begin{tcolorbox}[breakable,colback=orange!5!white, colframe=black, title={Cognitive Prompt for Best Stopping Judgement}]

\textbf{User:}
Completed trials:
    
\pkd{completed\_trials}
    
The following **Search Space** contains **unexplored** trials.
\pkd{trials}

Instructions:
Task 1: Review Analysis of Completed Trials (trial analysis, performance trends, highlights, and other insights)

Task 2: Decide Whether to Stop Optimization

Based on the above analysis and **Completed Trials**, determine whether the optimization process should be stopped.

Carefully analyze each of the following stop rules and provide a short (1-2 sentences) justification for whether it is met:

1. Have all promising configurations identified based on performance trends been tested?

2. Is it unlikely that unexplored configurations will perform better based on the observed trends and the law of diminishing returns?

3. Has the best metric improved significantly?

Step 2: Decide whether **all** conditions are met.

If **all** criteria in Step 1 are met, Answer: Yes, with confidence score: \pkd{confidence\_socre}. Otherwise, Answer: No with confidence score: \pkd{confidence\_socre}.
\end{tcolorbox}

\subsubsection{Strategic Planning}
\label{sec:supp_planning}
This prompt implements the strategic planning stage of Cognitive Agent. Its objective is to generate candidate experimental configurations for the next iteration based on accumulated experimental evidence, while balancing refinement of high-performing configurations and exploration of under-sampled regions.

\begin{tcolorbox}[breakable,colback=orange!5!white, colframe=black, title={Cognitive Prompt for Strategic Planning}]

\textbf{User:}
Instructions:

Task 1: Review Analysis of All Completed Trials

Completed trials:

\pkd{completed\_trials}

The following **Search Space** contains **unexplored** trials:

\, \pkd{trials}

Instructions:

Task 2: Optimization Recommendation

Recommend exactly \pkd{n\_jobs} promising trials from the provided **Search Space** (include both number and params).

\textbf{Rules:}

1. ``params'' MUST include:

\,\,\pkd{HyperName}

2. All selected `params` must match exactly with the provided **Search Space**. Do NOT leave out any key.

3. Use the analysis in **Task 1** (trial analysis, performance trends, highlights, and other insights) to guide selection.

4. Based on the above analysis, explore under-explored regions only when there is clear evidence of potential performance gain.

5. Do not mix, modify, or create new values.

6. You MUST not output any JSON blocks in this part.

7. You MUST provide reasoning for each recommendation.
\end{tcolorbox}
\section{Details of SelfAI Benchmark}
\label{sec:supp_details_benchmark}
In this section, we summarize the averaged performance across all benchmark tasks. Classical baselines, including grid search and TPE-based Bayesian optimization, achieve competitive final objective values but obtain low Scores and high $\text{AUP}_D$, reflecting redundant late-stage exploration and delayed stopping (Table~\ref{tab:hit}). Naive LLM solvers improve early discovery in some tasks but exhibit substantial variability across domains. In contrast, SelfAI-driven solvers based on mid-sized models, such as Qwen2.5-7B and GPT4-o3-mini, consistently achieve the highest Scores with substantially lower $\text{AUP}_D$ and earlier stopping times, indicating more focused and efficient discovery trajectories. Larger models (e.g., Qwen2.5-72B and Llama3.3-70B) do not necessarily yield improved efficiency, highlighting that trajectory-aware reasoning within SelfAI play a more critical role than model scale. For completeness, we discuss representative atypical trajectory behaviors in the Results section, while corresponding examples are provided. These cases illustrate exploration behaviors and breakdown patterns observed in SelfAI Benchmark, including premature stopping, limited integration of earlier observations, and sensitivity to minor perturbations, and serve to contextualize the aggregate performance trends reported above.

\subsection{Machine Learning}
{\bf Boston house pricing prediction.} We evaluate SelfAI on the Boston housing price prediction task~\cite{boston} using a random forest regression model. A total of 162 trials were conducted across a search space defined by five hyperparameters: i.e., $\text{n-estimators}=[100, 200, 300]$, $\text{max-depth} = [\text{None}, 10, 20]$, $\text{min-samples-split}=[2, 5, 10]$, $\text{min-samples-leaf=[1,2,5]}$ and $\text{max-features}=[\text{``sqrt'', ``log2''}]$). 

As shown in Fig.~\ref{fig:bar_score_aup} and Supplementary Table~\ref{tab:Boston}, GPT-4o-mini achieves the highest overall Score (0.9811), ranking first among all evaluated solvers. This performance reflects its ability to rapidly identify high-performing configurations while terminating exploration shortly thereafter, yielding superior optimization efficiency. Within the DeepSeek-R1 family, the 32B and 7B variants rank second and third, respectively, exhibiting a favorable balance between early convergence (low Best-Time) and moderate exploration diversity among open-source models. In contrast, larger models such as DeepSeek-R1-70B and Llama3.3-70B demonstrate sustained and consistent exploration but exhibit less effective stopping behavior, leading to prolonged search trajectories and delayed convergence. Notably, nearly all solvers reach an identical and high Best Result (approximately 0.841), indicating that differences in final predictive accuracy are minimal. Instead, the primary distinction lies in how efficiently and reliably solutions are discovered, underscoring the importance of trajectory-aware reasoning and adaptive stopping in practical scientific optimization.

{\noindent\bf Sentiment analysis.} We perform experiments for sentiment analysis that focus on identifying opinions, emotions, and attitudes expressed in text. All experiments are found in~\cite{sentiment_analysis} with standard experimental settings, where pre-trained Word2Vec embeddings~\cite{word2vec} are used as input features. An LSTM network is then employed to model sentence sequences, converting them into a multi-class classification problem. This setup serves to evaluate the agent's reasoning and optimization capabilities in textual domains.
As reported in Supplementary Table~\ref{tab:LSTM}, GPT-4o-mini achieves the highest Score (0.8824), consistently identifying the optimal configuration at an early stage while maintaining a balanced level of exploration. In contrast, the DeepSeek-R1 and Qwen2.5 series exhibit substantially lower Scores, frequently showing delayed convergence or failing to reach the global optimum. Notably, the largest frontier model, GPT-4o, performs markedly worse (Score 0.2745, rank 11), despite its superior scale. This result indicates that model size alone does not guarantee effective iterative hyperparameter reasoning in recurrent neural architectures, and that disciplined trajectory reasoning combined with adaptive stopping plays a more decisive role in efficient optimization within textual learning tasks.

\subsubsection{Scientific Computing}
We selected the tensor decomposition method~\cite{tw}. The method is an important tool for high-dimensional data analysis and is crucial in applications such as data compression~\cite{li2025lossless}, computational acceleration~\cite{berezutskii2025tensor}, and multi-modal data fusion~\cite{steyaert2023multimodal}. All LLMs are provided with identical mathematical knowledge about TW decomposition generated by GPT-4o. As shown in Supplementary Table~\ref{tab:TW}, Qwen2.5 (7b and 14b) and GPT4-o3-mini perform well and rank highly. In contrast, Qwen2.5-72B exhibits reduced performance, suggesting that broader general reasoning alone may be insufficient to navigate the specific trade-offs inherent in tensor decomposition. In the DeepSeek series, although the DeepSeek-R1-7b model can't find the optimal solution, its search strategy was acceptable with a moderate Score, demonstrating its ability to perform mathematical reasoning. DeepSeek-R1-70b and GPT-4o, despite excellent search diversity and early stopping, received a mediocre score, reflecting that their exploration strategy was not well aligned with the optimization goal of tensor decomposition, possibly failing to find the optimal performance between data compression and computational efficiency.

\subsection{Computer Vision}
{\noindent\bf A. SIREN} We employed SIREN (Sinusoidal Representation Networks)~\citep{sitzmann2020implicit} to evaluate our framework on image segmentation and denoising tasks. Leveraging sine activations and coordinate-based inputs, SIREN excels at representing high-frequency signals through continuous implicit representations, making it widely used in scientific computing and physics-based problems. However, its performance is highly sensitive to hyperparameters (e.g., learning rate and regularization strength, etc.), requiring careful tuning per dataset. The unsupervised nature of SIREN further increases the risk of training instability or divergence with improper settings, making it a strong test case for SelfAI's hyperparameter optimization capability. 

\begin{figure*}[t]
    \centering
    \begin{tabular}{cccc}

        \multicolumn{4}{c}{\textbf{BS}} \\
        \includegraphics[width=0.22\textwidth]{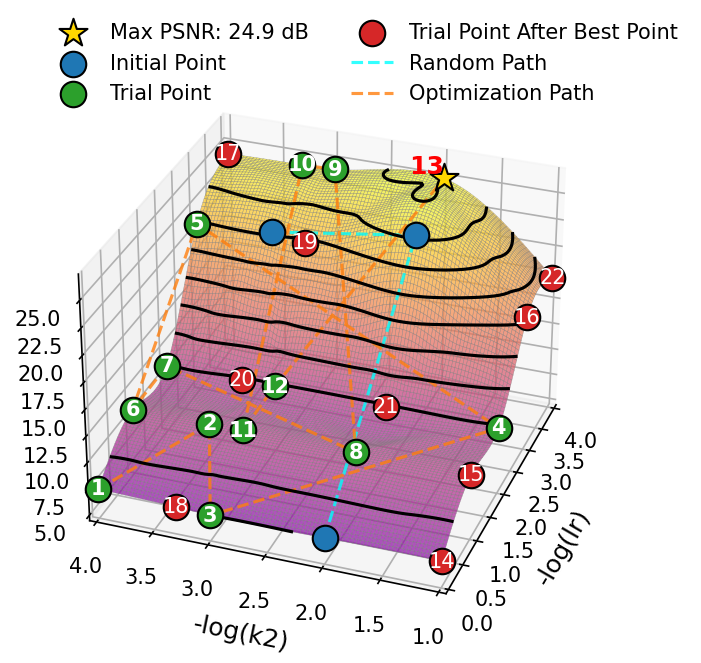}
        & \includegraphics[width=0.22\textwidth]{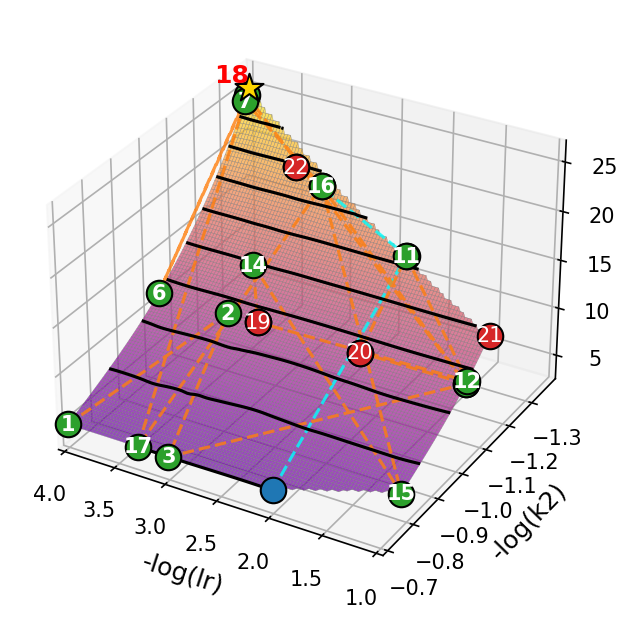}
        & \includegraphics[width=0.22\textwidth]{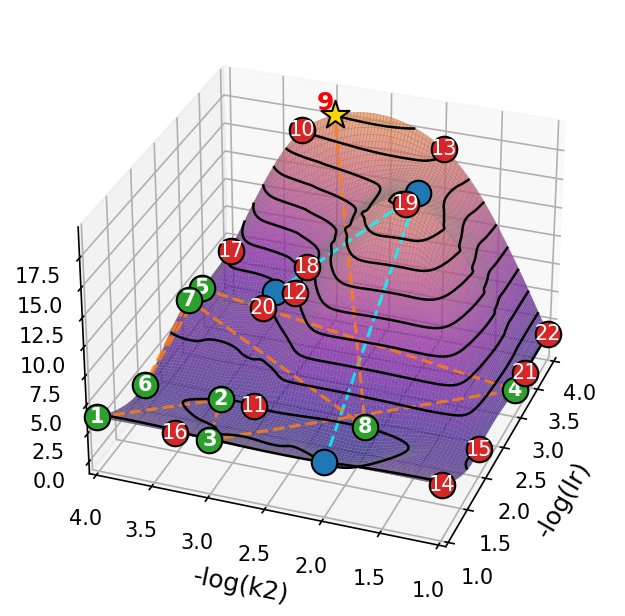}
        & \includegraphics[width=0.22\textwidth]{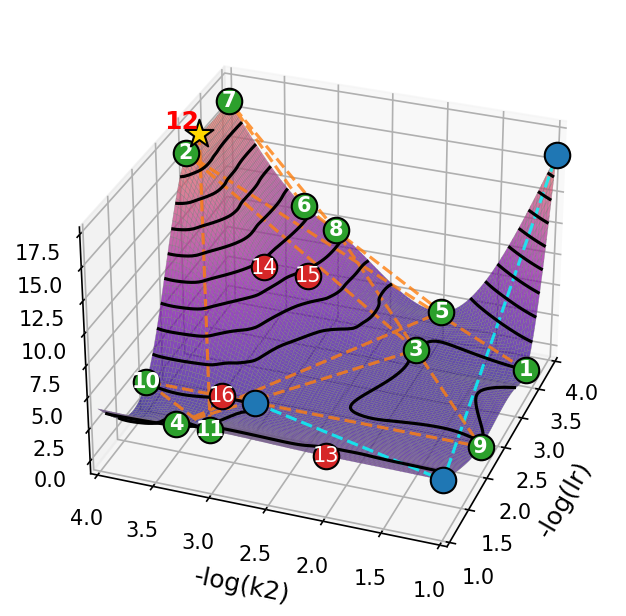}\\

        \multicolumn{4}{c}{\textbf{GPT4-o3-mini}} \\
        \includegraphics[width=0.22\textwidth]{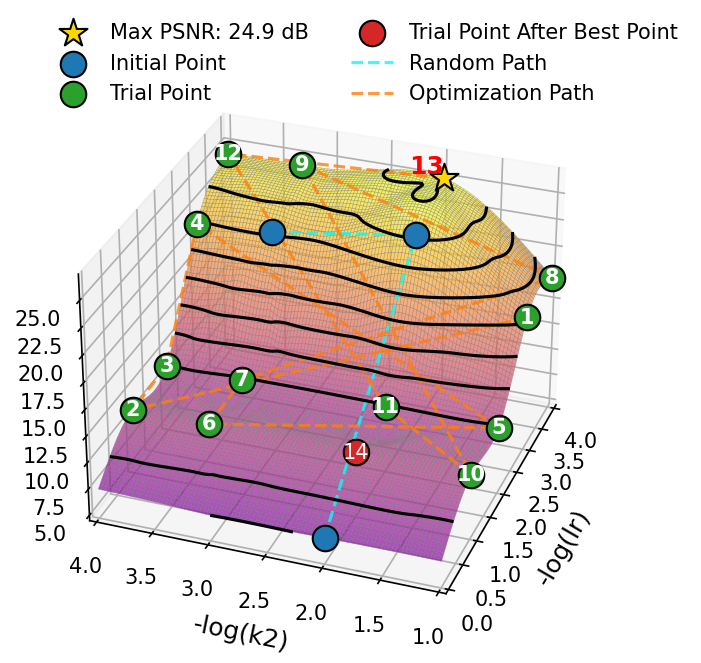}
        & \includegraphics[width=0.22\textwidth]{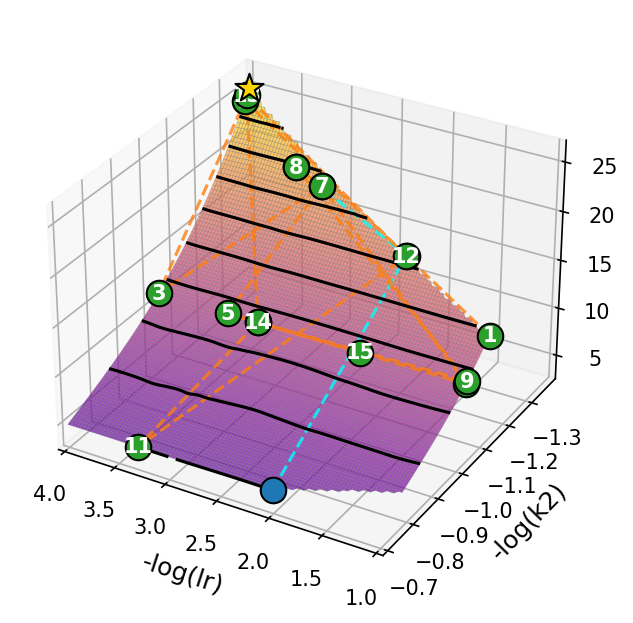}
        & \includegraphics[width=0.22\textwidth]{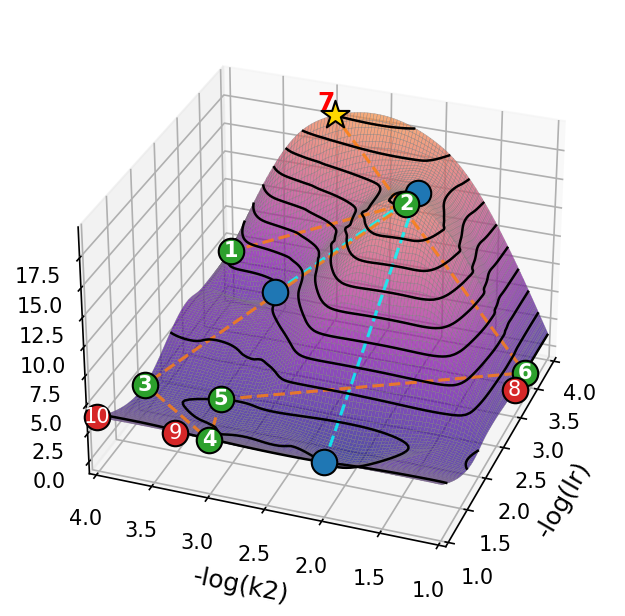}
        & \includegraphics[width=0.22\textwidth]{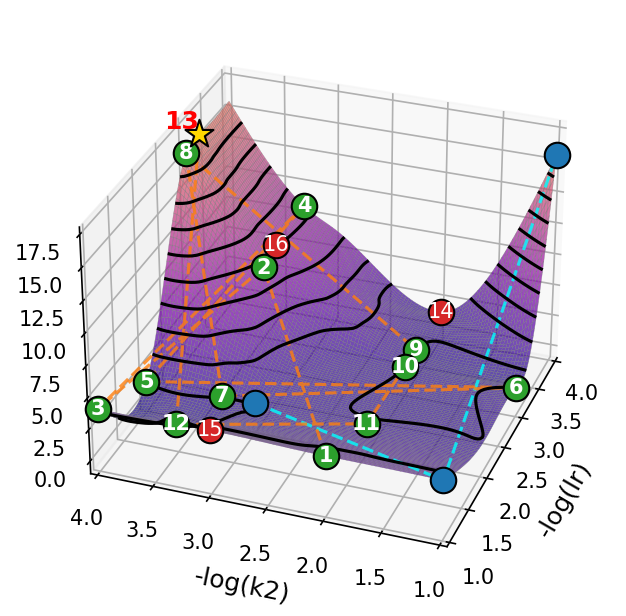} \\

        \multicolumn{4}{c}{\textbf{GPT4-o3}} \\
        \includegraphics[width=0.22\textwidth]{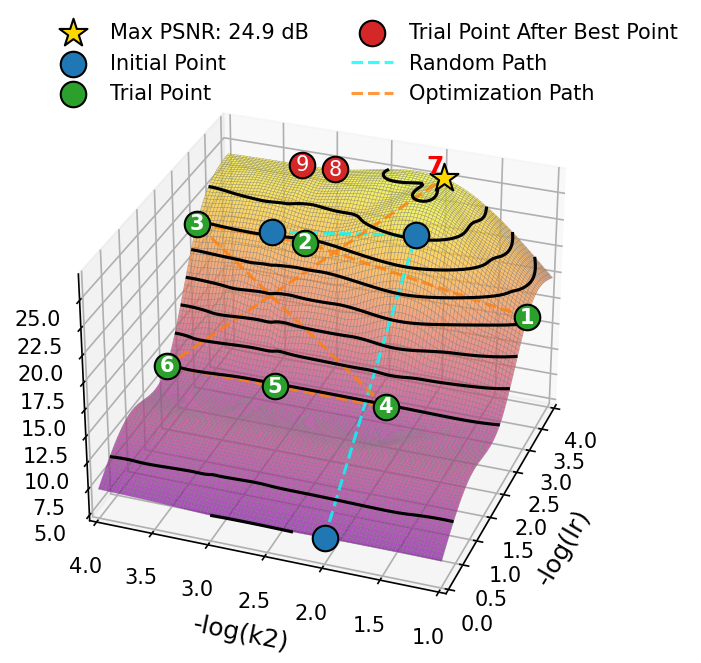}
        & \includegraphics[width=0.22\textwidth]{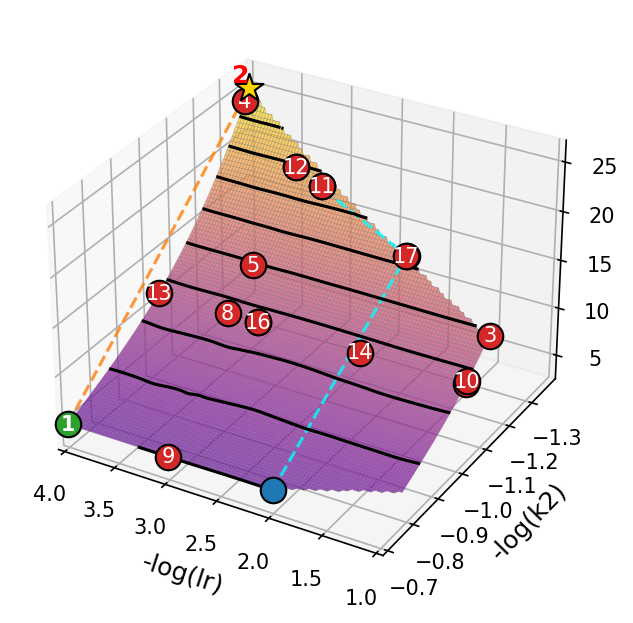}
        & \includegraphics[width=0.22\textwidth]{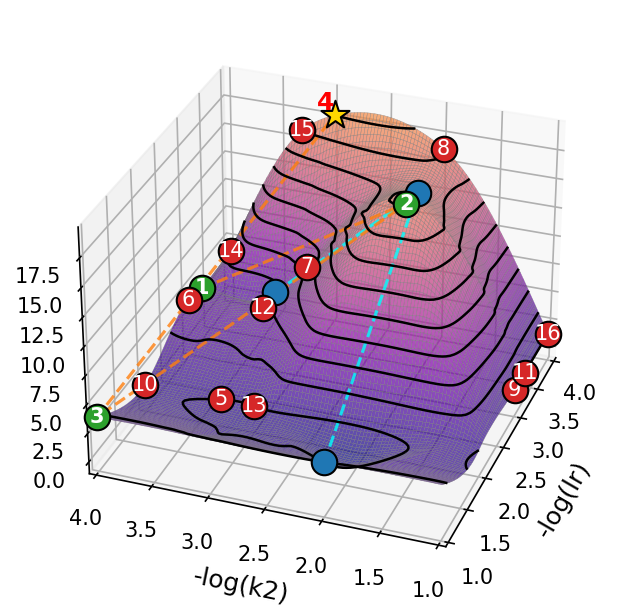}
        & \includegraphics[width=0.22\textwidth]{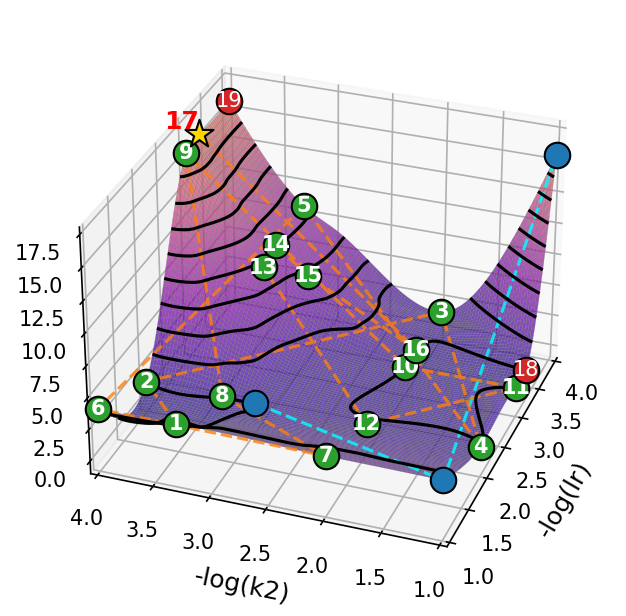} \\

        \multicolumn{4}{c}{\textbf{Llama3.3-70B}} \\
        \includegraphics[width=0.22\textwidth]{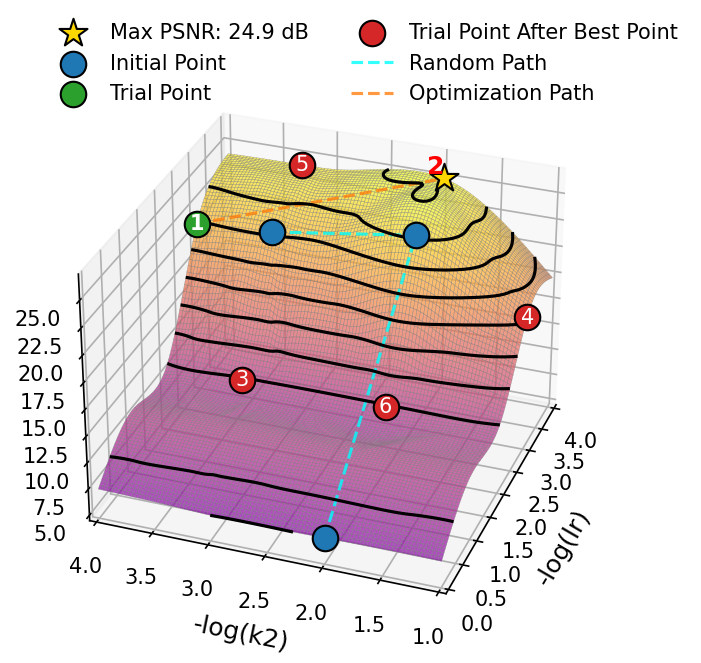}
        & \includegraphics[width=0.22\textwidth]{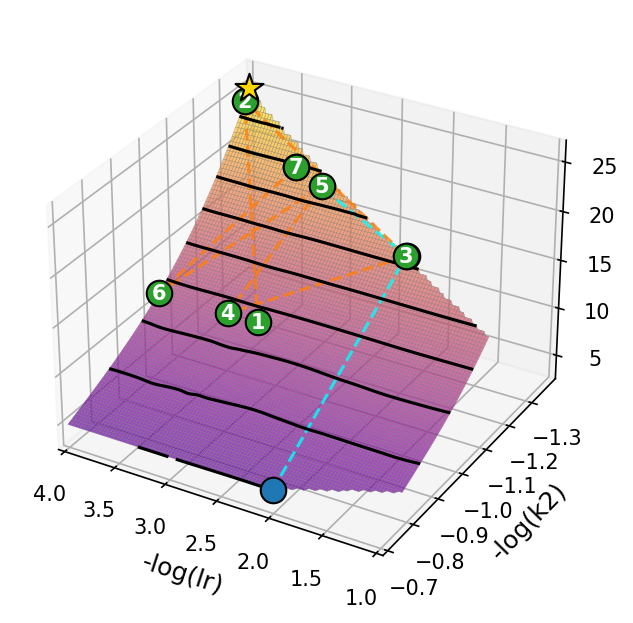}
        & \includegraphics[width=0.22\textwidth]{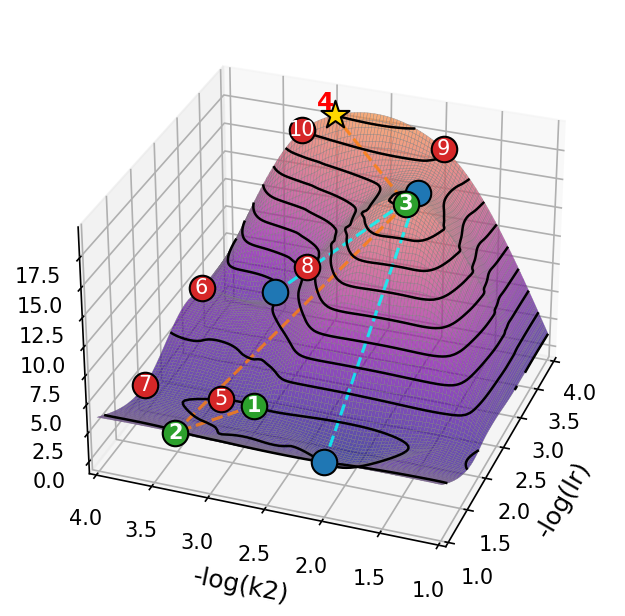}
        & \includegraphics[width=0.22\textwidth]{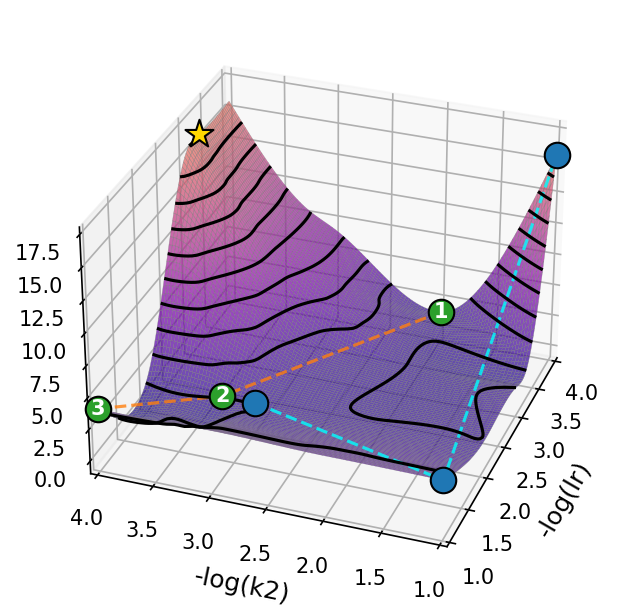} \\
    \end{tabular}
    \caption{Illustration of the optimized trajectory using the SIREN method for additional cases in image denoising (first two columns) and segmentation (last two columns). Blue points are initial points. Green points represent suggested points before reaching the optimum. Red points indicate redundant suggestions generated after the optimal point has been reached. The $\star$ marks the optimal point. The numbered labels indicate the sequence of recommendations provided by LLMs.}
    \label{fig:siren_surface1}
\end{figure*}

\begin{figure*}[t]
    \centering
    \begin{tabular}{cccc}
        \multicolumn{4}{c}{\textbf{Qwen2.5-7B}} \\
        \includegraphics[width=0.22\textwidth]{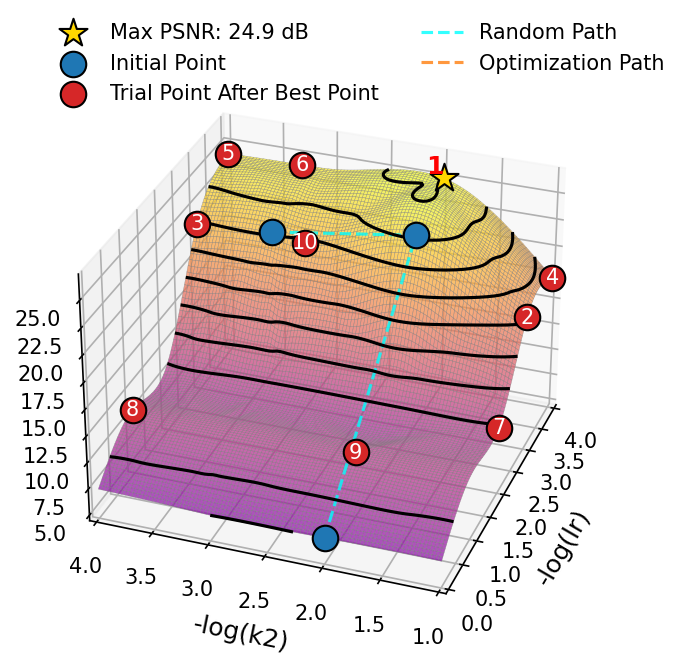}
        & \includegraphics[width=0.22\textwidth]{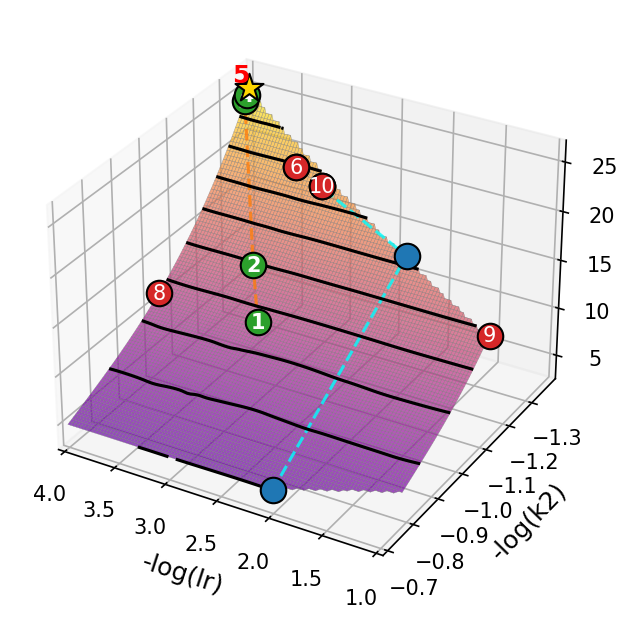}
        & \includegraphics[width=0.22\textwidth]{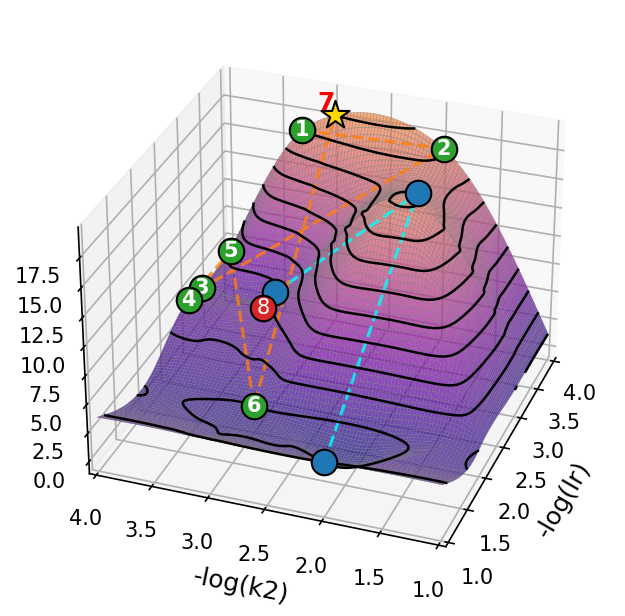}
        & \includegraphics[width=0.22\textwidth]{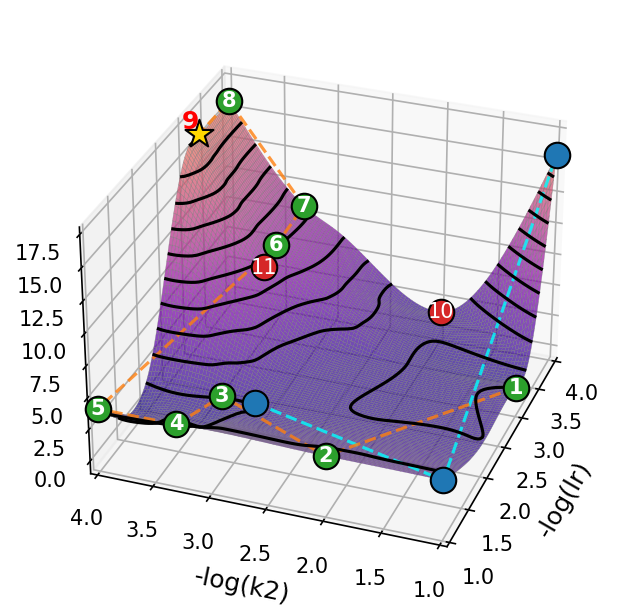} \\

        \multicolumn{4}{c}{\textbf{Qwen2.5-14B}} \\
        \includegraphics[width=0.22\textwidth]{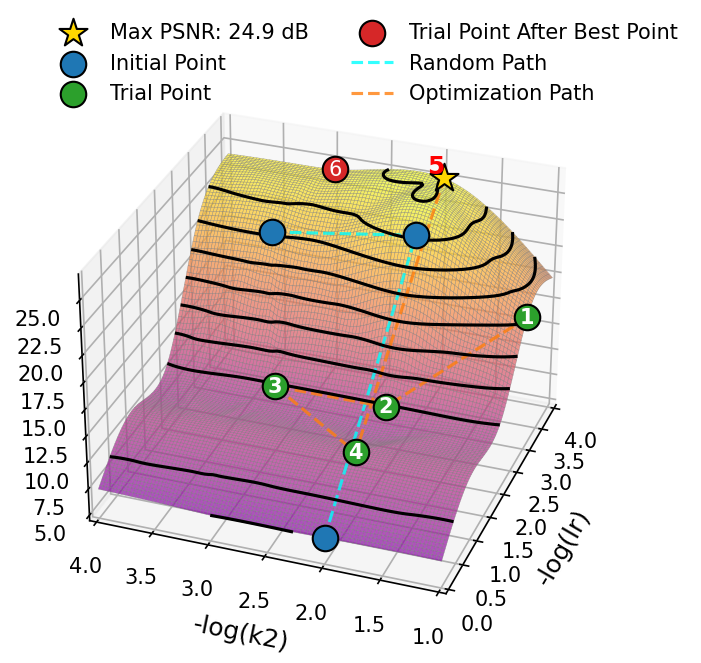}
        & \includegraphics[width=0.22\textwidth]{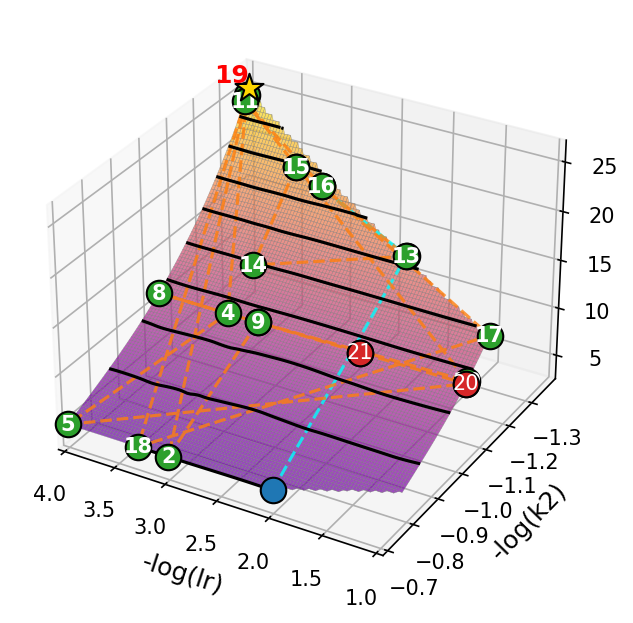}
        & \includegraphics[width=0.22\textwidth]{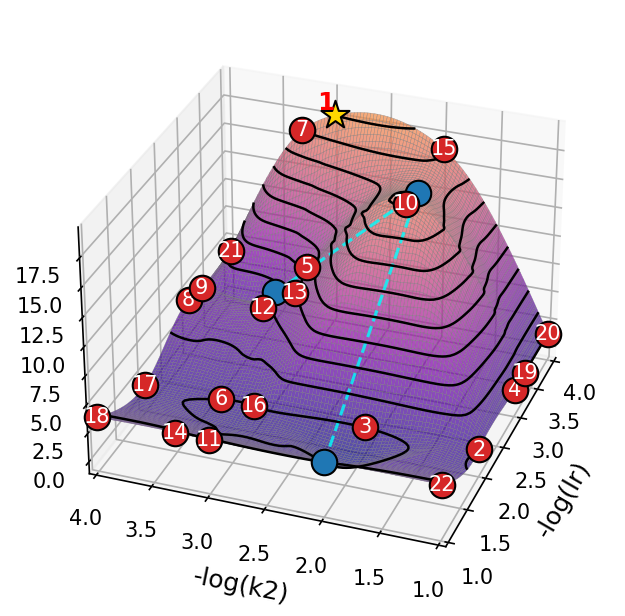}
        & \includegraphics[width=0.22\textwidth]{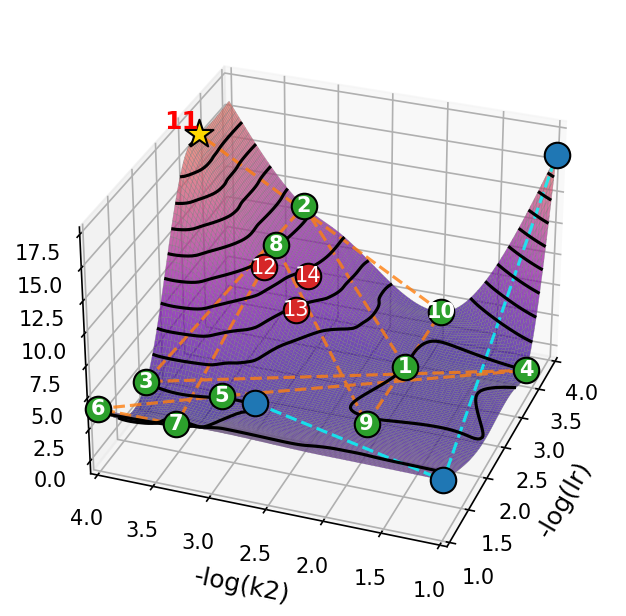} \\

        \multicolumn{4}{c}{\textbf{Qwen2.5-72B}} \\
        \includegraphics[width=0.22\textwidth]{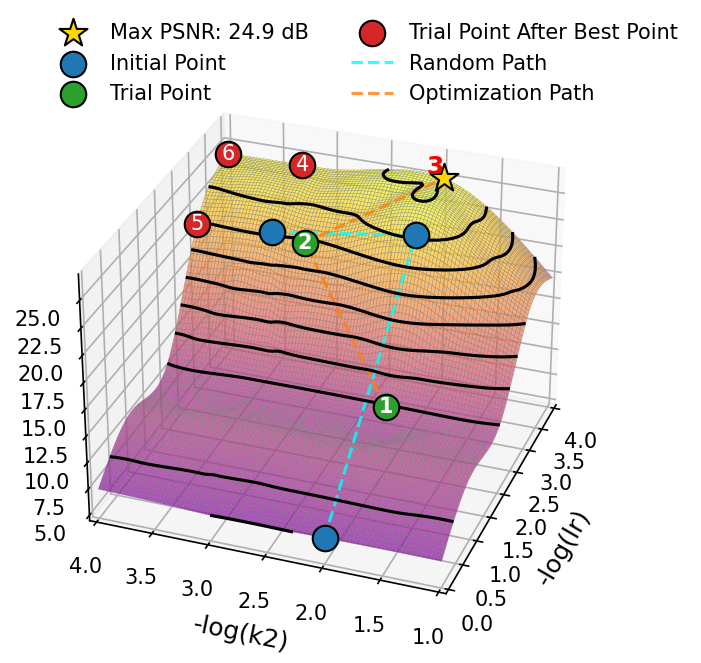}
        & \includegraphics[width=0.22\textwidth]{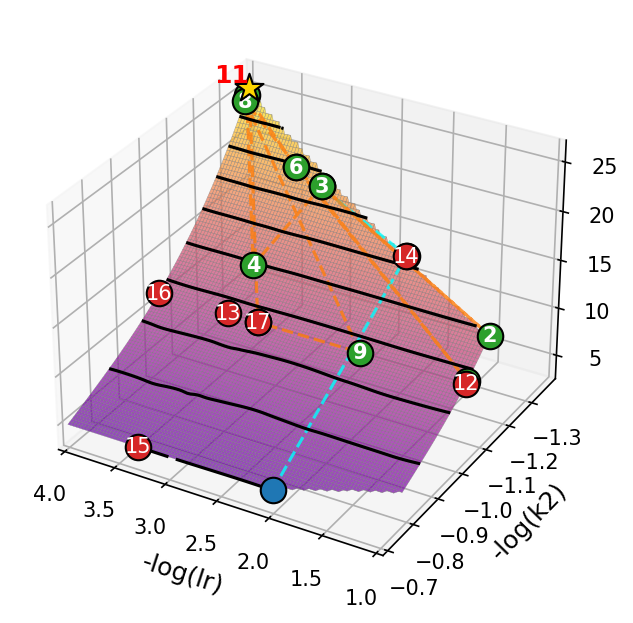}
        & \includegraphics[width=0.22\textwidth]{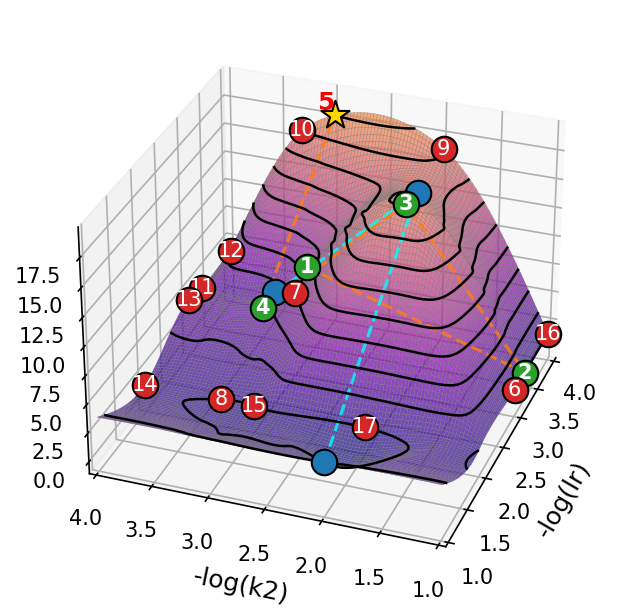}
        & \includegraphics[width=0.22\textwidth]{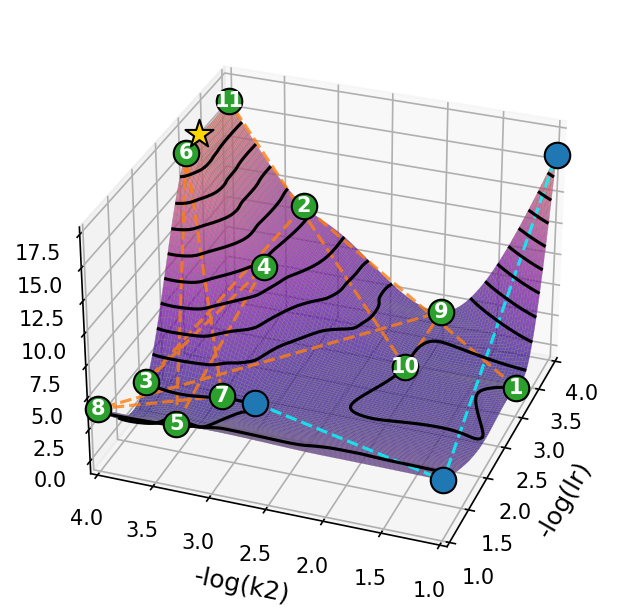} \\
    
        \multicolumn{4}{c}{\textbf{Qwen2.5-72B}} \\
        \includegraphics[width=0.22\textwidth]{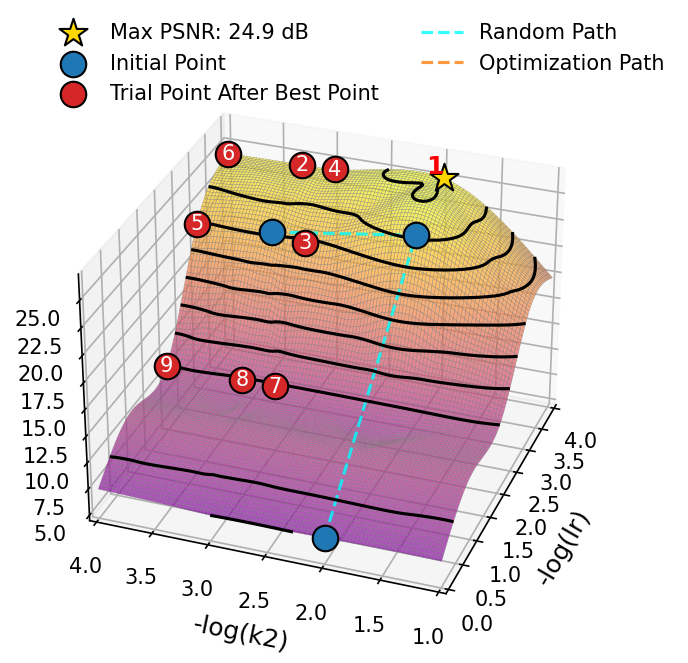}
        & \includegraphics[width=0.22\textwidth]{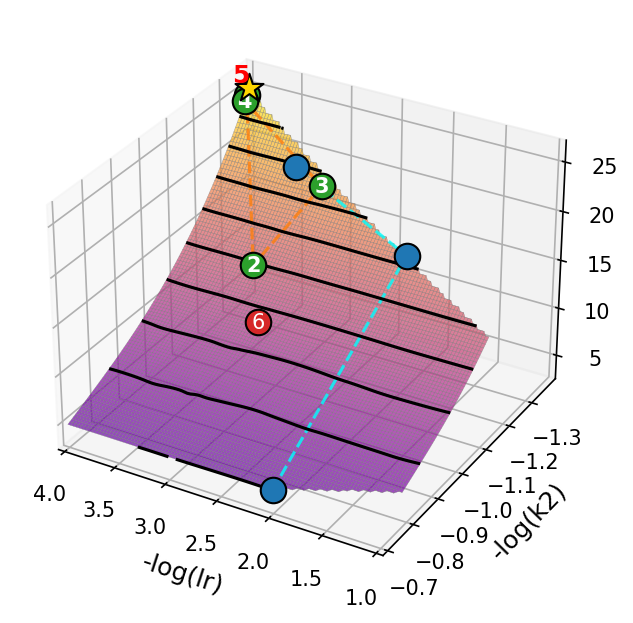}
        & \includegraphics[width=0.22\textwidth]{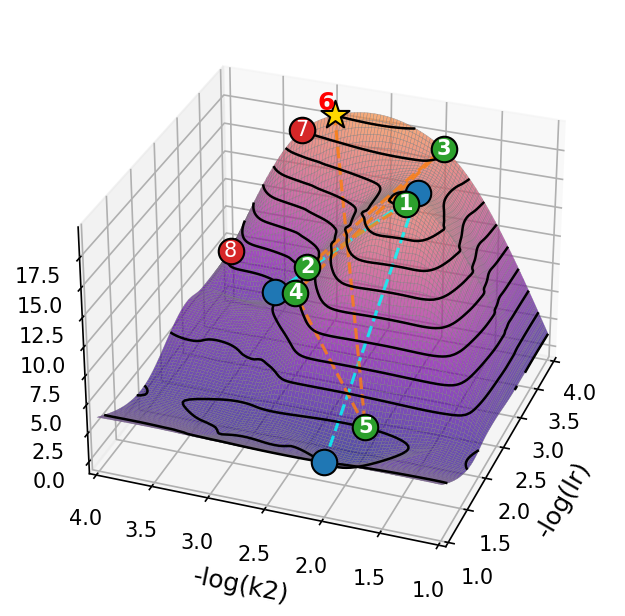}
        & \includegraphics[width=0.22\textwidth]{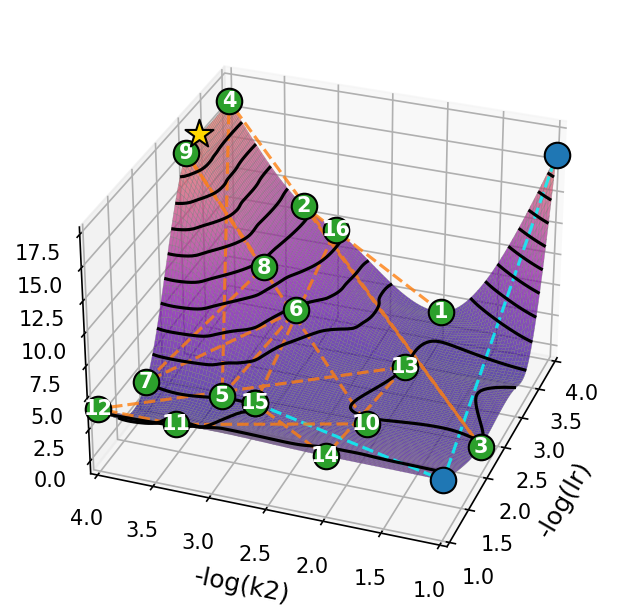} \\
    \end{tabular}
    \caption{More optimized trajectories of Fig.~\ref{fig:siren_surface1}. }
    \label{fig:siren_surface2}
\end{figure*}

\begin{figure*}[t]
    \centering
    \begin{tabular}{cccc}
        \multicolumn{4}{c}{\textbf{DeepSeek-R1-7B}} \\
        \includegraphics[width=0.22\textwidth]{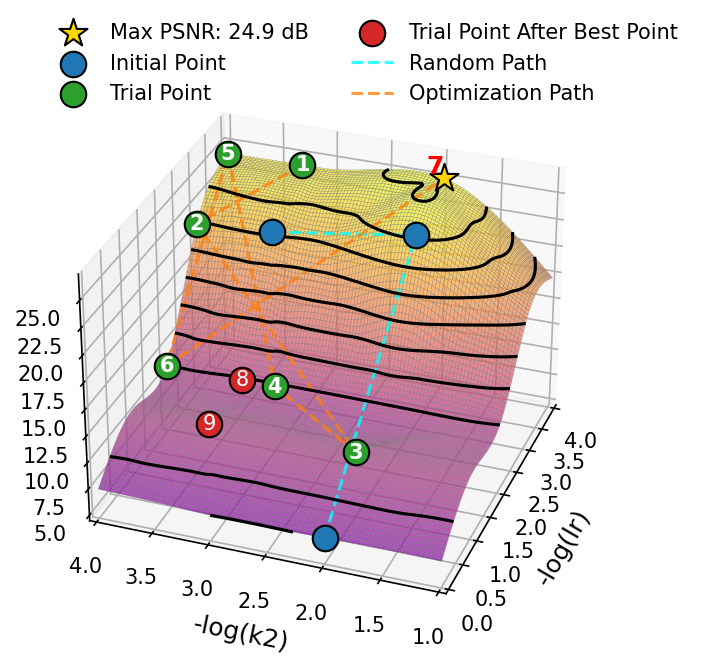}
        & \includegraphics[width=0.22\textwidth]{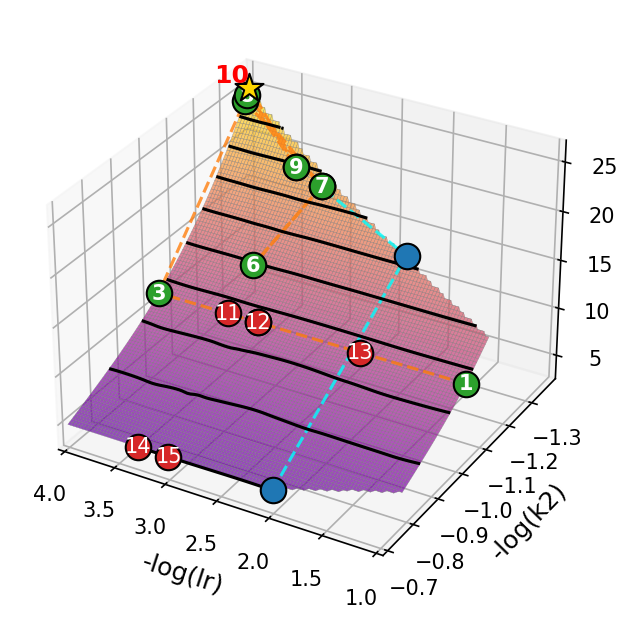}
        & \includegraphics[width=0.22\textwidth]{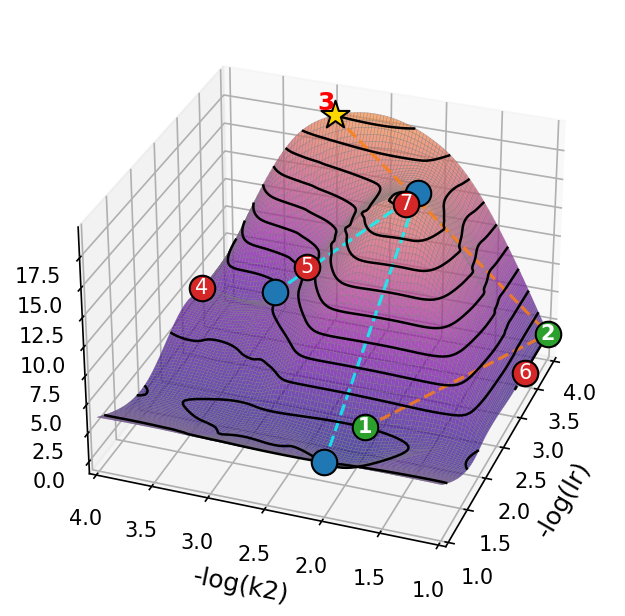}
        & \includegraphics[width=0.22\textwidth]{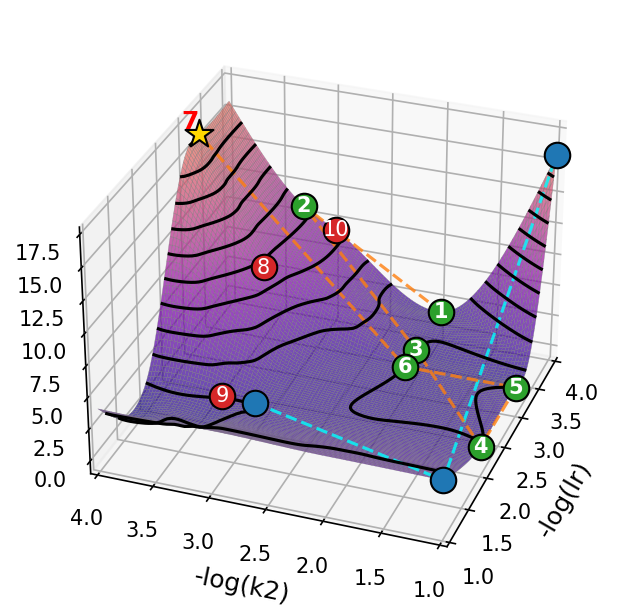} \\

        \multicolumn{4}{c}{\textbf{DeepSeek-R1-14B}} \\
        \includegraphics[width=0.22\textwidth]{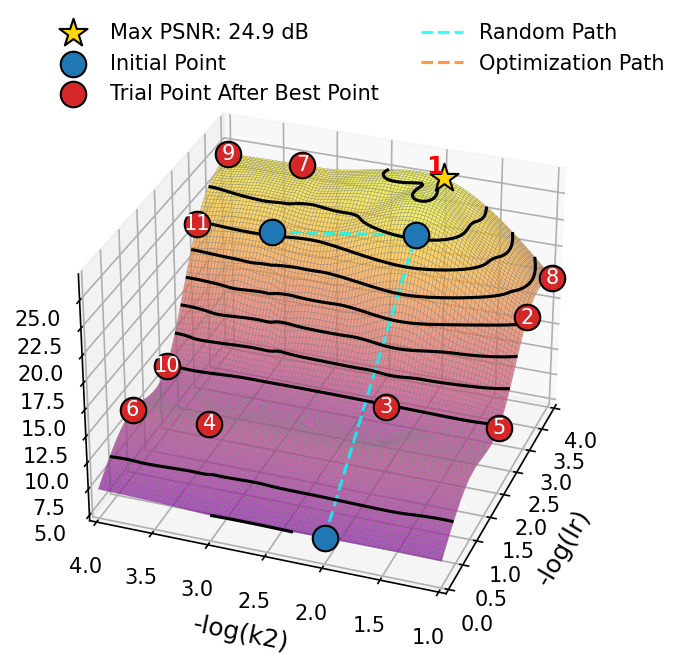}
        & \includegraphics[width=0.22\textwidth]{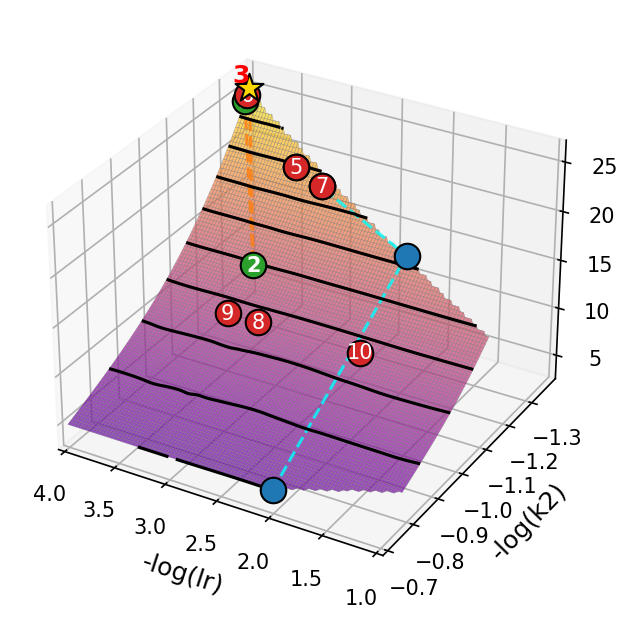}
        & \includegraphics[width=0.22\textwidth]{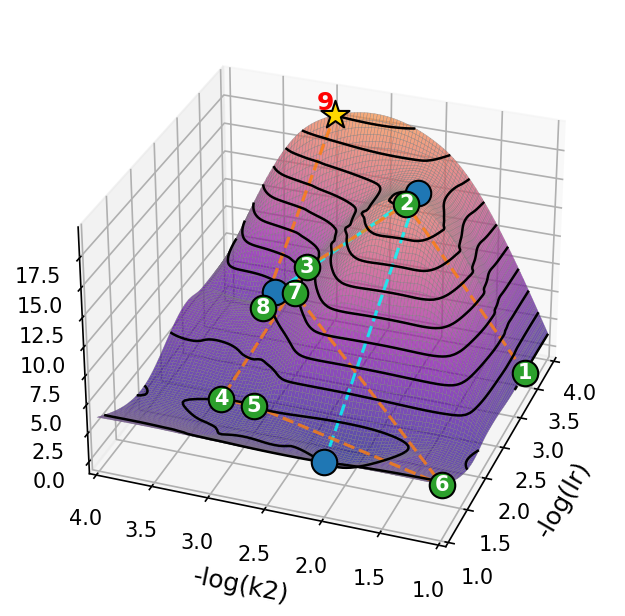}
        & \includegraphics[width=0.22\textwidth]{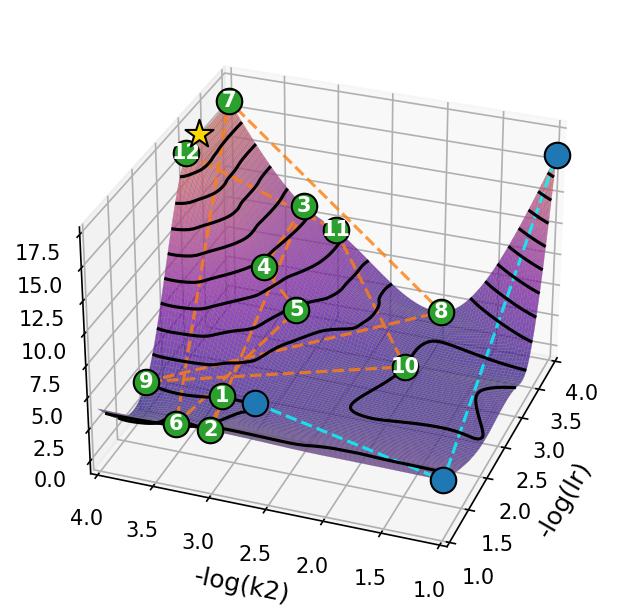} \\

        \multicolumn{4}{c}{\textbf{DeepSeek-R1-32B}} \\
        \includegraphics[width=0.22\textwidth]{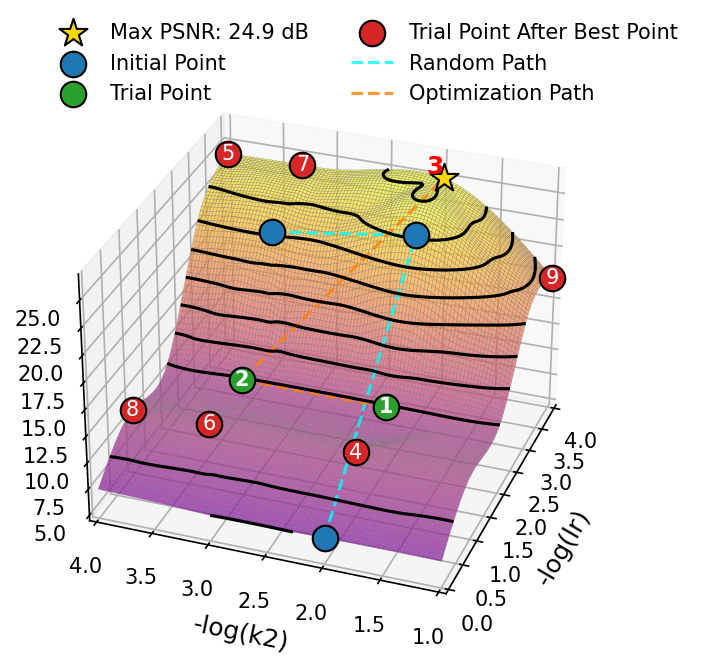}
        & \includegraphics[width=0.22\textwidth]{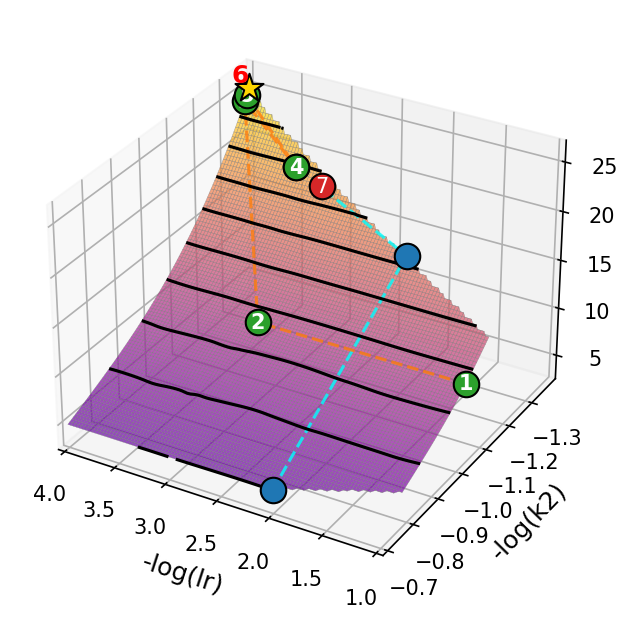}
        & \includegraphics[width=0.22\textwidth]{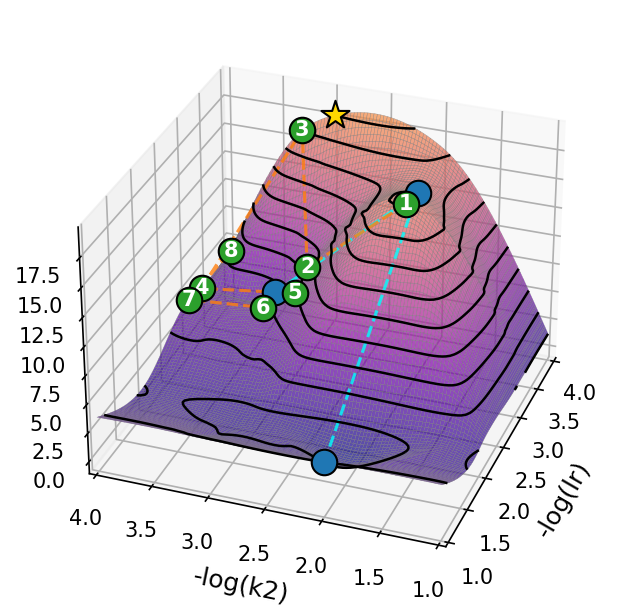}
        & \includegraphics[width=0.22\textwidth]{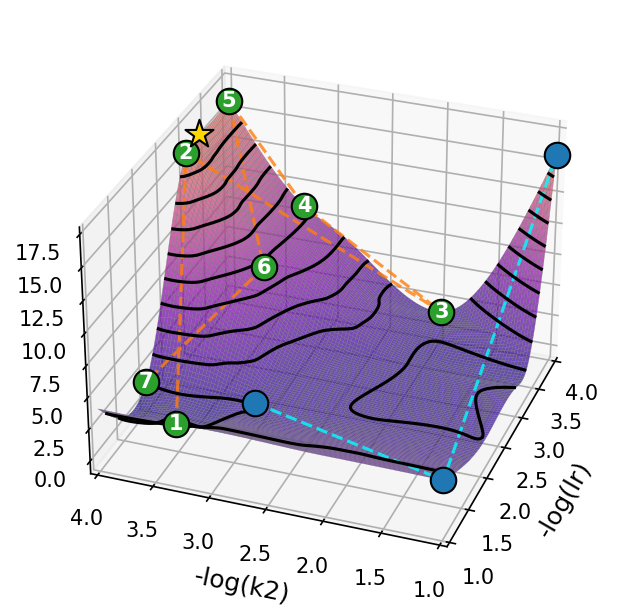} \\
    
        \multicolumn{4}{c}{\textbf{DeepSeek-R1-70B}} \\
        \includegraphics[width=0.22\textwidth]{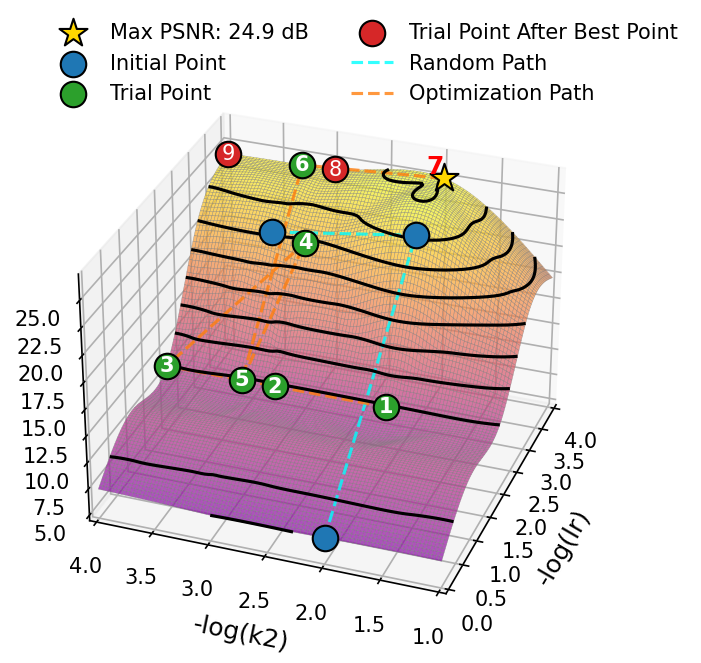}
        & \includegraphics[width=0.22\textwidth]{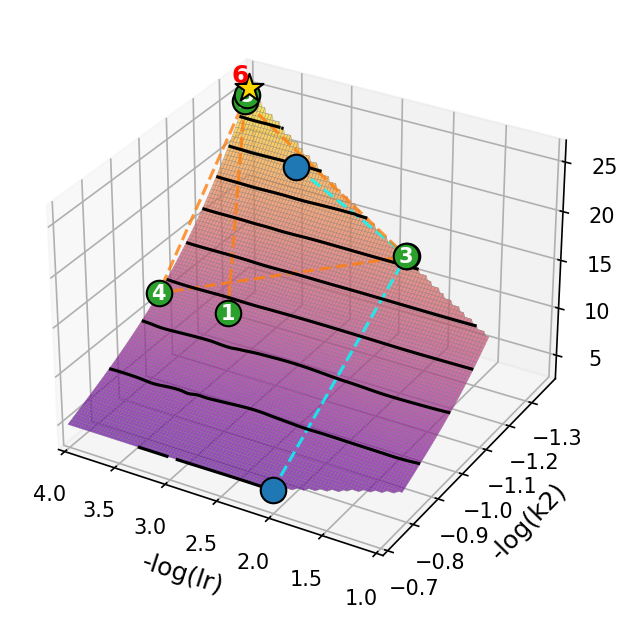}
        & \includegraphics[width=0.22\textwidth]{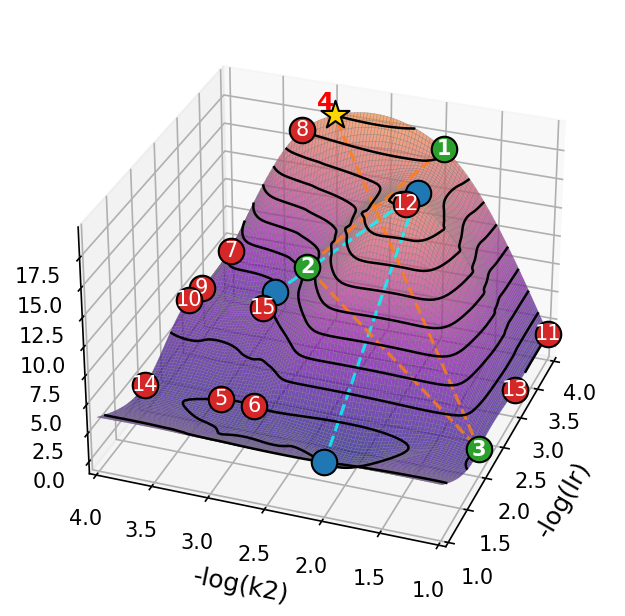}
        & \includegraphics[width=0.22\textwidth]{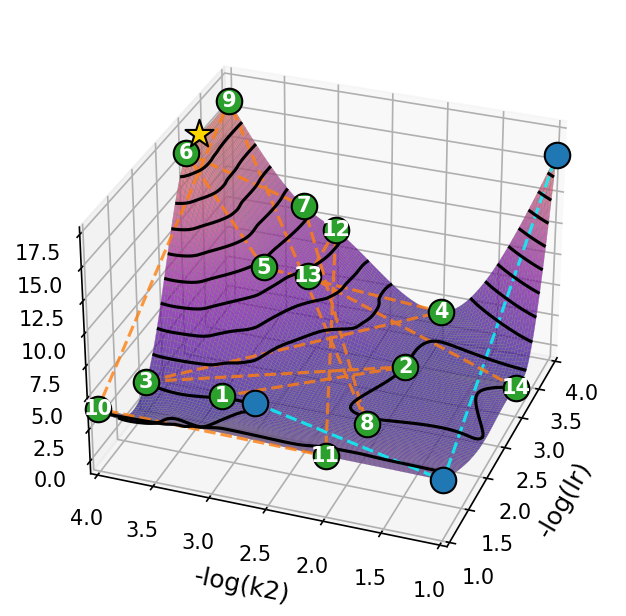} \\
    \end{tabular}
    \caption{More optimized trajectories of Fig.~\ref{fig:siren_surface1}. These LLMs have different suggestions for non-convex hyperparameter optimization. }
    \label{fig:siren_surface3}
\end{figure*}

Fig.~\ref{fig:siren_surface} illustrates hyperparameter search trajectories for image segmentation using SIREN (see also Supplementary Figs.~\ref{fig:siren_surface1}, \ref{fig:siren_surface2}, and \ref{fig:siren_surface3}), where the surface, smoothed from the original steep, multi-peak data, reveals distinct search behaviors. Given the same three initial points, the tree-structured Bayesian optimizer (BS) follows a spiral trajectory that underutilizes promising starting regions and fails to explore broadly, often converging to local minima. In contrast, LLM-based optimizers infer trends from evaluated points, incorporate causal reasoning, and explore more broadly to locate the global optimum efficiently. They also monitor progress and halt when improvements plateau, a capability absent in traditional methods.

Quantitative results (Supplementary Tables~\ref{tab:siren_seg}-\ref{tab:siren_denoising}) reveal distinct task-dependent performance profiles. In segmentation, DeepSeek-R1-7B achieves the highest Score (0.693) and ranks first, followed by Qwen2.5-7B and GPT-4-o3-mini. Larger models such as Qwen2.5-72B and Llama-3.3-70B rank only mid-range, indicating that scale alone does not ensure effective optimization. In denoising, the landscape shifts: Qwen2.5-72B delivers the best performance (Score 0.761), while DeepSeek-R1-14B and DeepSeek-R1-32B jointly occupy second place. Notably, models that perform well in segmentation tasks do not necessarily perform well in denoising tasks, suggesting that the effectiveness of the solver largely depends on the degree of matching between its inference strategy and the task.

{\noindent\bf B. Image Classification} We also benchmark two commonly used supervised learning methods in computer vision: Masked Autoencoder (MAE)~\cite{mae} and ResNet~\cite{resnet}. These represent fundamentally different learning paradigms where hyperparameter optimization plays a crucial role in achieving the latest performance.

{\noindent\bf Mask Autoencoder (MAE)} MAE introduces two key hyperparameters: training strategies (linear detection vs. fine-tuning) and masking rate, which are explored over a range: $ [0.10, 0.90] $ with an interval of 0.1. The masking rate directly affects the difficulty of the reconstruction task and the quality of the learning representation, while the training strategy determines how to adapt to the downstream task. In our experiments (see Supplementary Table~\ref{tab:MAE} for more details), GPT-4o-mini achieved explicit performance with efficiency, demonstrating the extraordinary ability to identify the best mask configuration and training strategies. Qwen2.5-7B achieves early convergence and disciplined stopping. The DeepSeek-R1 series shows consistent scaling behavior: performance improves from 7B to 14B to 32B, but slightly drops at 70B. In traditional solvers, GS and BS lag significantly behind LLM-based solvers, highlighting the advantages of LLM-enabled discovery. Apart from Qwen2.5-14b and DeepSeek-R1-7b, most solvers can find the optimal solution, but significant differences exist among the various methods in terms of search efficiency and convergence speed.

{\noindent\bf ResNet Family} On the ImageNet ResNet hyperparameter benchmark, LLMs in the 7B-32B range demonstrate superior optimization efficacy. The top-performing solvers, such as Qwen2.5-14B (1st), DeepSeek-R1-32B (2nd), and Qwen2.5-7B with DeepSeek-R1-14B (tied 3rd), consistently identify the optimal ResNet configuration within one or two trials, halting immediately with near-perfect sample efficiency. In contrast, larger variants from the same model families drop to middle or lower rankings, consuming over half the search budget with notably higher exploration overhead. While GPT-series models, as large-scale counterparts, still achieve competitive results, the 7B-32B class exhibits a clear advantage in balancing accuracy and efficiency. For the architectural search on ImageNet, we explored key dimensions such as depth and bottleneck design. Qwen2.5-14B attains theoretical peak performance, with its 7B and 32B versions completing further exploration efficiently. These results reflect an ability to identify ideal network configurations with minimal wasted exploration.

{\noindent\bf Large-scale benchmark} We evaluate SelfAI on the standard AutoML benchmark~\cite{zimmer2021auto} using Bayesian optimization over 2000 rounds of hyperparameter search, covering critical parameters including learning rate, batch size, network depth, and dropout rate (Supplementary Fig.~\ref{fig:LCBench_analysis}). These trials exceed the maximum token limit of GPT-family models, thereby SelfAI does not employ GPT4-o3-mini or GPT4-o3. Supplementary Fig.~\ref{fig:LCBench_analysis}a-e presents that accuracy depends on interacting hyperparameters such as weight decay, hidden-layer width, and depth, and the high-performance region is sharply localized. The optimization process using the Tree Parzen Estimator (TPE) optimizer is shown with performance fluctuations that cannot be stopped (Supplementary Fig.~\ref{fig:LCBench_analysis}f), illustrating highly sparse promising regions. Supplementary Fig.~\ref{fig:LCBench_analysis}g presents sharply local high-performance regions are presented arising from interacting hyperparameters. We explore these regions in Supplementary Table~\ref{tab:LCBenchmark}, where leading models (DeepSeek-R1-70B, Qwen2.5-14B, and Llama3.3-70B) efficiently identify near-optimal performance by evaluating only a minimal set of candidate solutions, achieving an order-of-magnitude improvement in stop-time compared to classical methods, as promising regions are identified arising from interacting hyperparameters.


\subsection{Medical Image Analysis}
In medical image analysis, nnU-Net~\cite{nnunet} is a landmark framework known for its strong generalization and automated design. Despite its adaptive network structure, preprocessing, and training strategies for various segmentation tasks, the exploration of novel architectures persists. To address this, nnU-Net-Revisited~\cite{nnunet_revisit} establishes a comprehensive benchmark including 19 mainstream models (CNN-based~\cite{ronneberger2015u,nnunet,roy2023mednext}, Transformer-based~\cite{vaswani2017attention,tang2022self}, and Mamba-based~\cite{gu2023mamba,ma2024u}), offering a solid basis for fair evaluation. Results are shown in Figs.~\ref{fig:bar_score_aup} and Supplementary Tables~\ref{tab:MIABench}-~\ref{tab:nnUnet}. We evaluate SelfAI on this benchmark across multiple datasets, with results summarized in Fig.~\ref{fig:bar_score_aup} and Supplementary Tables~\ref{tab:MIABench}-\ref{tab:nnUnet}. On the BTCV dataset~\cite{btcv}, SelfAI instantiated with GPT4-o3-mini achieves the highest Score (0.6875) and ranks first, indicating rapid identification of high-quality configurations combined with efficient stopping behavior. Notably, Qwen2.5-7B secures second place, outperforming all larger variants within the same model family, suggesting that effective optimization in this setting does not scale monotonically with model size. A similar pattern emerges in nnU-Net hyperparameter tuning on the BraTS dataset~\cite{brats}. Qwen2.5-14B attains the top ranking (Score 0.4333), followed closely by GPT4-o3-mini, while Qwen2.5-32B and DeepSeek-R1-70B tie for third. Across both datasets, smaller (7B) and very large (72B) models outperform medium-sized counterparts (14B-32B), reinforcing the observation that increased model capacity alone does not guarantee superior optimization performance. These results highlight the importance of trajectory-aware decision-making under noisy and high-variance evaluation regimes, which are characteristic of medical image analysis. By committing early to promising configurations while avoiding prolonged exploration with diminishing returns, SelfAI enables efficient discovery in settings where excessive experimentation is both computationally costly and practically constrained.

\subsection{Imbalanced Node Classification}
Graph neural networks (GNNs) are a central modeling paradigm for relational data and have become indispensable across a range of scientific domains. In biomedicine, GNNs underpin modern protein structure prediction systems, such as AlphaFold3, by enabling explicit modeling of spatial interactions among amino acid residues, thereby advancing structural biology at unprecedented resolution~\cite{alphafold3}. In drug discovery, GNNs are widely applied to molecular property prediction, compound-target interaction modeling, and de novo molecule generation~\cite{bongini2021molecular}. Despite their success, optimizing GNNs on imbalanced datasets remains particularly challenging, due to sparse and asymmetric reward signals, complex relational dependencies, and high-dimensional hyperparameter spaces. These characteristics make such tasks especially sensitive to exploration strategy and stopping decisions. To investigate whether trajectory-aware reasoning can address these challenges, we evaluate GraphSAGE on the imbalanced Cora benchmark~\cite{zhao2021graphsmote} under the setup~\cite{yan2023unreal}.

As shown in Fig.~\ref{fig:bar_score_aup}, SelfAI instantiated with the GPT4-o3 series consistently achieves near-optimal performance, characterized by rapid identification of optimal configurations, efficient stopping decisions, and sustained exploration diversity. This behavior indicates that SelfAI effectively balances exploitation and exploration in structurally complex optimization landscapes. In contrast, Qwen2.5 models (with the exception of the 72B variant) exhibit stable but middling performance across scales, while larger models tend to converge later and incur substantial redundant exploration. This pattern suggests potential limitations in capturing the structural optimization requirements of relational domains such as molecular graphs and protein interaction networks (see Supplementary Table~\ref{tab:GraphSAGE}). Across different language model backbones, SelfAI consistently demonstrates fast discovery of high-quality configurations while avoiding premature convergence. These results provide evidence that trajectory-level reasoning and adaptive stopping generalize beyond continuous parameter optimization, extending effectively to structured, relational learning problems that are common in real-world scientific applications.

\subsection{Drug Discovery}
Drug discovery increasingly benefits from advances in machine learning and data-driven modeling, which serve as versatile frameworks for tasks such as bioactivity prediction, virtual screening, and compound prioritization. These approaches have demonstrated the potential to accelerate early-stage discovery, reduce experimental burden, and shorten development timelines~\cite{brahmavar2024generating,grisoni2021combining}. Following the evaluation practices and modeling standards summarized by Korotcov et al.~\cite{korotcov2017comparison}, this study employs the Chagas EP20 dataset~\cite{Chagas}, which measures the activity of compounds against Trypanosoma cruzi, a neglected but medically significant target in tropical disease research. The core challenge lies in learning an effective mapping from molecular representations, such as SMILES sequences~\cite{weininger1988smiles}, to experimentally measured bioactivity~\cite{fliri2005biological}. We adopt SelfAI to adaptively refine model performance, monitor discovery trajectories, and determine principled stopping criteria for iterative optimization. Experimental results (see Supplementary Table~\ref{tab:chagas}) reveal distinct performance patterns among solvers in hyperparameter optimization. SelfAI and select LLM-based solvers achieve superior Scores by rapidly identifying promising hyperparameter regions and terminating exploration once gains plateau. Notably, these results demonstrate efficient optimization under realistic biochemical uncertainty, highlighting trajectory-aware optimization as a practical computational enabler for scalable and resource-efficient bioactivity modeling in drug discovery pipelines.


\begin{figure}
    \centering
    \includegraphics[width=1.0\linewidth]{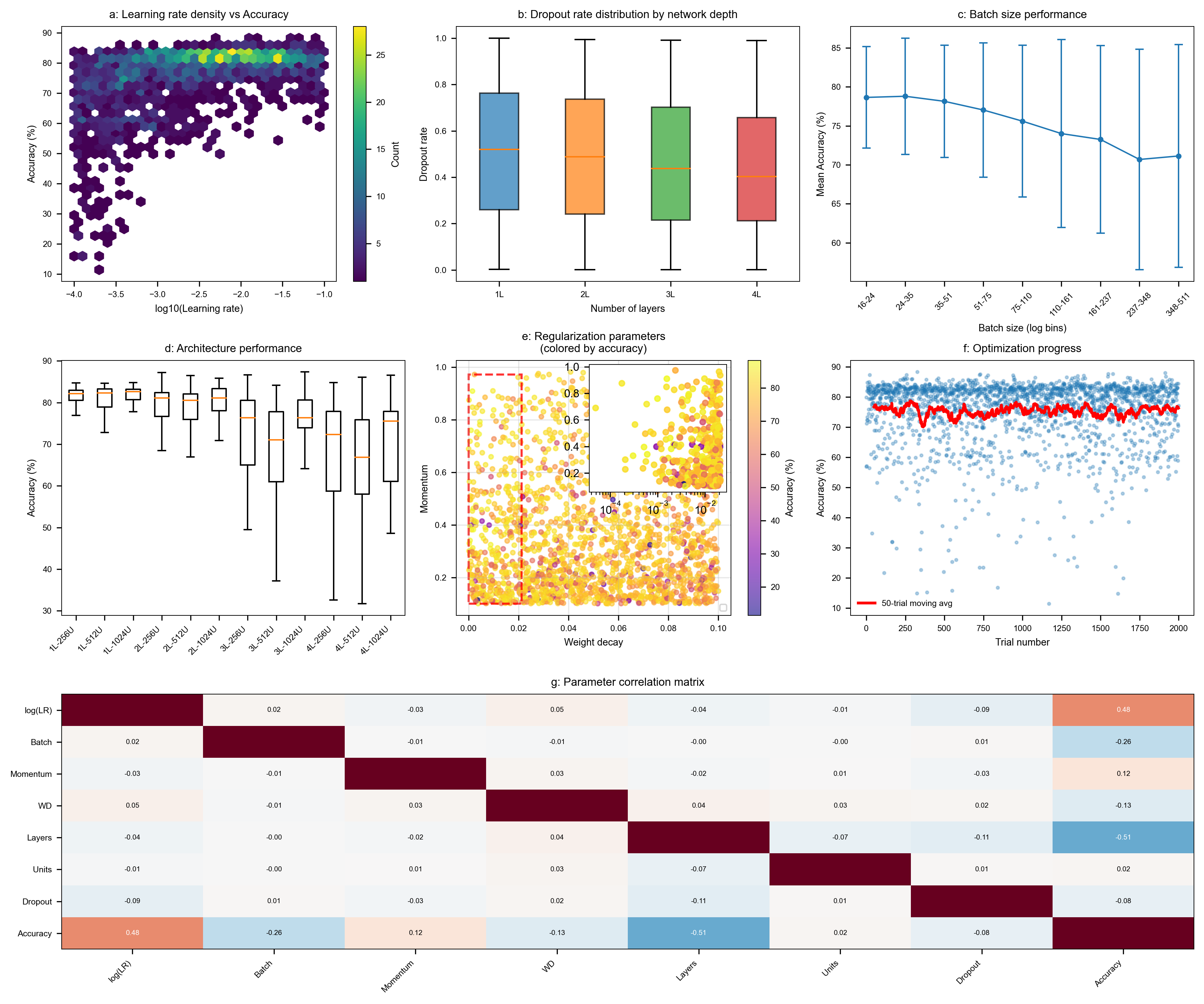}
    \caption{The relationship between different parameter settings and accuracy. \textbf{a}, A hexbin plot showing the joint density of learning rate (log-scaled) and accuracy. \textbf{b}, Box plots illustrating the distribution of dropout rates used for models of different depths. \textbf{c}, Mean accuracy and standard error across logarithmically binned batch sizes. \textbf{d}, Performance comparison of different architectural configurations (layers and units). \textbf{e}, A scatter plot of weight decay against momentum, colored by accuracy. \textbf{f}, The optimization trajectory, showing accuracy improvement over successive trials, with a moving average trend line. \textbf{g} A correlation matrix quantifying linear relationships between all parameters and the accuracy.}
    \label{fig:LCBench_analysis}
\end{figure}



\begin{table}[h]
    \renewcommand\arraystretch{1.2}
    \setlength{\tabcolsep}{2pt}
    \centering
    \scriptsize
    \caption{Averaged performance comparison of SelfAI across different tasks.}
    \label{tab:hit}
    \begin{tabular}{llccccccc}
    \toprule
    Solver & Score$\uparrow$ & $\text{AUP}_D\downarrow$ & Best-Time$\downarrow$ & Stop-Time$\downarrow$ & Best Result$\uparrow$ & Hit-Rate$\uparrow$ & Rank \\
    \midrule
    GS & 0.2453 & 1.0000 & 0.5094 & 1.0000 & 1.0000 & 1.0000 & 14 \\
    BS & 0.1927 & 0.8106 & 0.6265 & 0.9881 & 1.0000 & 1.0000 & 15 \\
     LLM & 0.3526 & 0.7638 & 0.2949 & 1.0000 & 1.0000 & 0.9286 & 13 \\
    LLM-ES & 0.5294 & 0.2349 & 0.4691 & 0.4582 & 0.9981 & 0.6429 & 3 \\
    \midrule
    Qwen2.5-7b      & 0.5562 & 0.2154 & 0.4805 & 0.3945 & 0.9957 & 0.7857 & 2 \\
    Qwen2.5-14b     & 0.5015 & 0.3310 & 0.4926 & 0.4997 & 0.9969 & 0.7857 & 5 \\
    Qwen2.5-32b     & 0.4287 & 0.4684 & 0.5252 & 0.6133 & 0.9972 & 0.7857 & 10 \\
    Qwen2.5-72b     & 0.4189 & 0.5358 & 0.4531 & 0.7087 & 0.9995 & 0.8571 & 11 \\
    DeepSeek-r1-7b  & 0.4769 & 0.1802 & 0.7020 & 0.3093 & 0.9927 & 0.5000 & 8 \\
    DeepSeek-r1-14b & 0.4793 & 0.3956 & 0.4953 & 0.5100 & 0.9948 & 0.7143 & 7 \\
    DeepSeek-r1-32b & 0.4989 & 0.3476 & 0.4535 & 0.5433 & 0.9933 & 0.7857 & 6 \\
    DeepSeek-r1-70b & 0.4556 & 0.3513 & 0.5392 & 0.5299 & 0.9962 & 0.7143 & 9 \\
    Llama3.3-70b    & 0.3625 & 0.5483 & 0.5099 & 0.7271 & 0.9683 & 0.7143 & 12 \\
    GPT4-o3-mini    & 0.6433 & 0.2259 & 0.3168 & 0.3961 & 0.9989 & 0.8571 & 1 \\
    GPT4-o3         & 0.5140 & 0.2284 & 0.5477 & 0.3961 & 0.9966 & 0.6429 & 4 \\
    \bottomrule
    \end{tabular}
\end{table}

\begin{figure}
    \centering
    \includegraphics[width=1.0\linewidth]{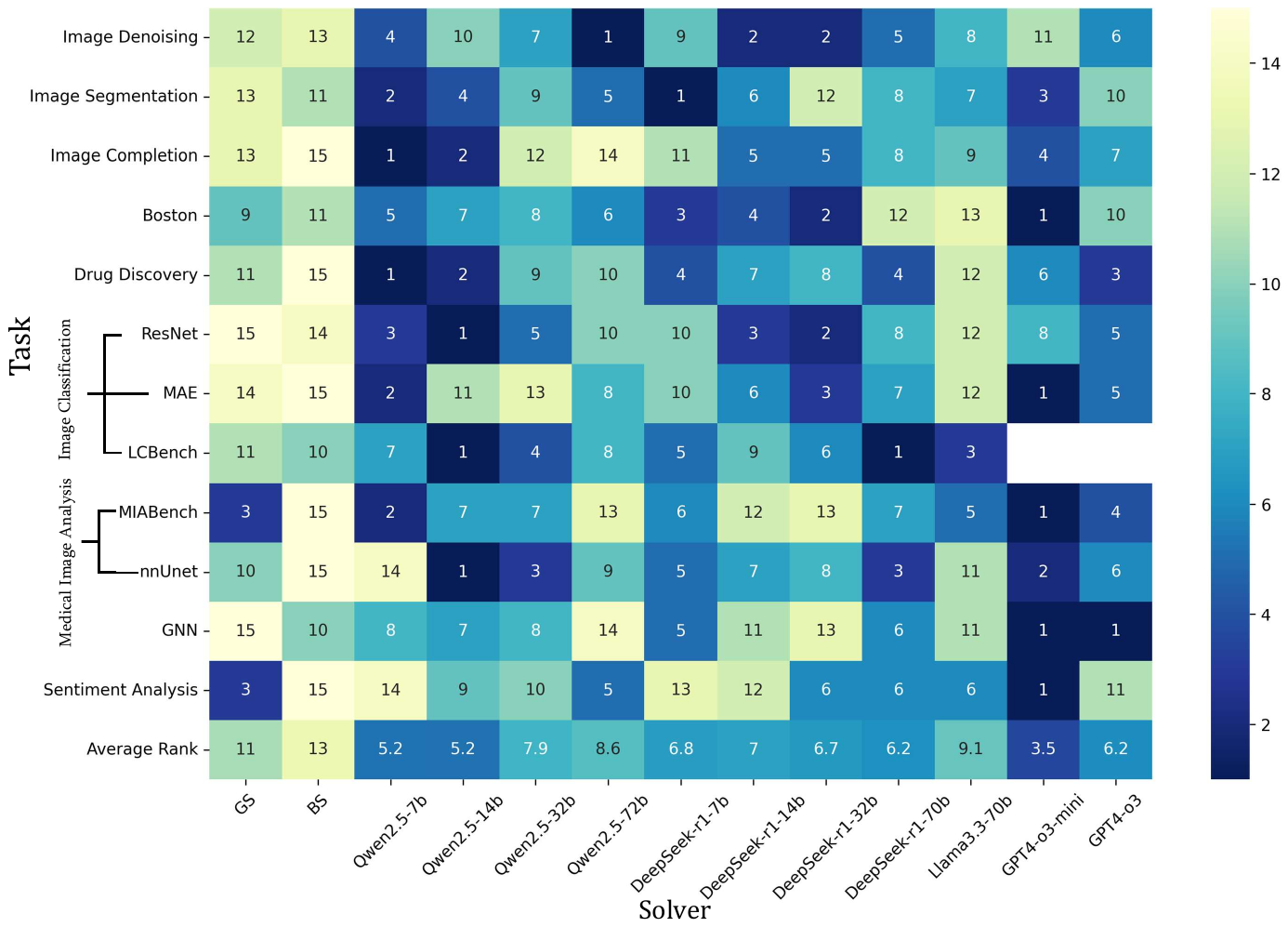}
    \caption{Performance ranking of different methods across multiple tasks. The heatmap displays the rank of each method (rows) for every task (columns), where lower numbers (darker colors) indicate better performance (e.g., 1st place). The final average rank is summarized on the right.}
    \label{fig:rank_heatmap}
\end{figure}

\subsection{Details of limitations}
\label{sec:supp_limitations}
\noindent\textbf{1) Context limitations.}
LCBench contains approximately 2000 trials, constituting a large and highly sparse search space. Supplementary Fig.~\ref{fig:LCBench_analysis} provides a structural analysis of this landscape. Performance depends on interacting hyperparameters such as weight decay, hidden-layer width, and depth, and high-performing regions are sharply localized (Supplementary Fig.~\ref{fig:LCBench_analysis}a-e). In practice, such large and sparse search spaces are common in scientific discovery tasks and can exceed the maximum context length supported by current GPT-family models.

\begin{figure}
    \centering
    \includegraphics[width=1.0\linewidth]{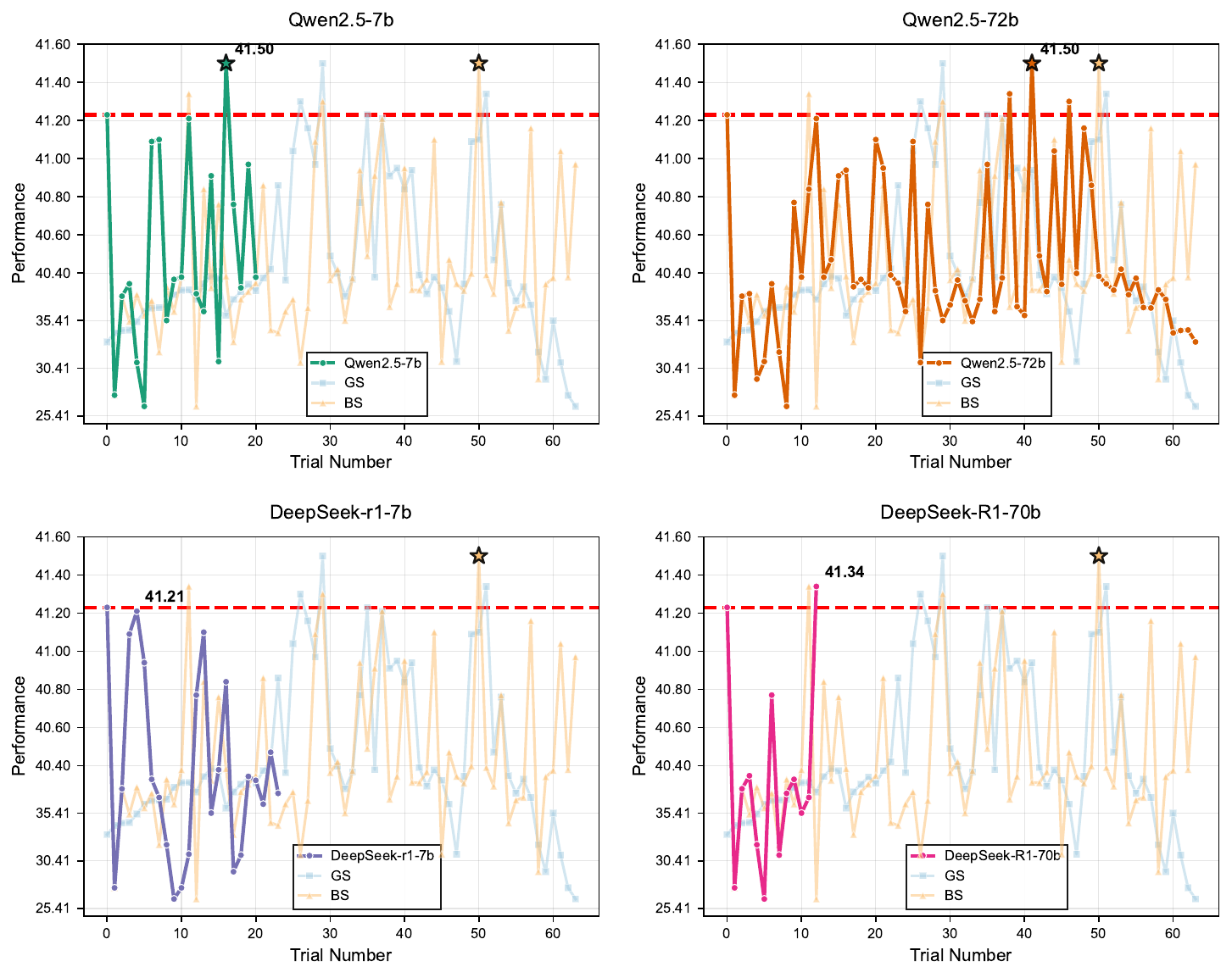}
    \caption{Failure Cases for the DeepSeek-R1 Family in the scientific computing field. Both DeepSeek-R1-7B and DeepSeek-R1-70B fail to reach the optimal results, marked with a star symbol ($\star$). Notably, DeepSeek-R1-7B fails to improve beyond the initial results, as indicated by the red dotted line, suggesting limited exploration capability in the search space.
    }
    \label{fig:failure_case}
\end{figure}

2) {\noindent\bf Computational fragility.} Supplementary Table~\ref{tab:hit} compares the best result hit rates across solvers. The performance of LLM-based solvers is highly sensitive to the reasoning strategy, and early stopping often leads to a decline in hit rate. In the supplementary tables, we report the best result \( t_{\text{best}} \) for each solver. \( t_{\text{best}} = 1 \) indicates that the optimal result was not found. Moreover, when \( t_{\text{best}} = 1 \), the score metric improves significantly, highlighting the solver's weakness in optimal stopping. As illustrated in Fig.~\ref{fig:failure_case}, computational fragility also hinders reliable performance improvement through parameter tuning, as seen in the non-monotonic performance of the DeepSeek-R1 model series.

3) {\noindent\bf Insufficient trajectory integration.} Supplementary Fig.~\ref{fig:failure_case} provides representative trajectory visualizations that highlight differences in evidence integration and stopping behavior. The Qwen2.5 models typically identify high-performing configurations early, with Qwen2.5-7B reaching near-optimal regions within the first 10-15 trials, followed by stable refinement with limited oscillation. In contrast, DeepSeek-R1 trajectories are more frequently non-monotonic. DeepSeek-R1-7B often oscillates between high- and low-performing regions, indicating instability in how accumulated evidence is leveraged, while DeepSeek-R1-70B can terminate prematurely after early improvements, leaving promising regions under-refined. Classical baselines (GS and BS) may also reach high-performing regions, but their improvements tend to be slower and less punctuated by rapid breakthroughs. Overall, these examples complement the aggregate metrics in the main text by illustrating how limitations can manifest as reduced trajectory stability, delayed breakthroughs, or suboptimal stopping decisions.

\clearpage
\section{Supplementary Tables}
\label{sec:supp_tables}
\begin{table}[h]
    \centering
    \setlength{\tabcolsep}{3pt}
    \renewcommand\arraystretch{1.5}
    \caption{Performance comparison of different solvers for the Boston housing prices based on the Random Forest Regressor.}
    \begin{tabular}{l|cccccc}
    \toprule
    Solver           & Score$\uparrow$ & $\text{AUP}_D\downarrow$ & Best-Time$\downarrow$ & Stop-Time$\downarrow$ & Best Result$\uparrow$ & Rank \\
    \midrule
    GS & 0.4969 & 1.0000 & 0.0062 & 1.0000 & 0.841 & 9 \\
    BS & 0.4654 & 0.9745 & 0.0755 & 0.9937 & 0.841 & 11 \\
    Llama3.3-70b & 0.3831 & 0.1330 & 1.0000 & 0.1509 & 0.837 & 13 \\
    Qwen2.5-7b & 0.7893 & 0.3463 & 0.0943 & 0.3270 & 0.841 & 5 \\
    Qwen2.5-14b & 0.7013 & 0.6108 & 0.0063 & 0.5912 & 0.841 & 7 \\
    Qwen2.5-32b & 0.6981 & 0.4081 & 0.2579 & 0.3459 & 0.841 & 8 \\
    Qwen2.5-72b & 0.7862 & 0.5250 & 0.0126 & 0.4151 & 0.841 & 6 \\
    DeepSeek-R1-7b & 0.9057 & 0.0910 & 0.0881 & 0.1006 & 0.841 & 3 \\
    DeepSeek-R1-14b & 0.8522 & 0.3401 & 0.0063 & 0.2893 & 0.841 & 4 \\
    DeepSeek-R1-32b & 0.9403 & 0.1618 & 0.0063 & 0.1132 & 0.841 & 2 \\
    DeepSeek-R1-70b & 0.3948 & 0.0215 & 1.0000 & 0.0189 & 0.833 & 12 \\
    GPT4-o3-mini & 0.9811 & 0.0062 & 0.0189 & 0.0189 & 0.841 & 1 \\
    GPT4-o3 & 0.4694 & 0.0560 & 1.0000 & 0.0377 & 0.840 & 10 \\
    \bottomrule
    \end{tabular}
    \label{tab:Boston}
\end{table}

\begin{table}[h]
    \centering
    \setlength{\tabcolsep}{3pt}
    \renewcommand\arraystretch{1.5}
    \caption{Performance comparison of different solvers for the sentiment analysis task based on the LSTM model.}
    \begin{tabular}{l|cccccc}
    \toprule
    Solver           & Score$\uparrow$ & $\text{AUP}_D\downarrow$ & Best-Time$\downarrow$ & Stop-Time$\downarrow$ & Best Result$\uparrow$ & Rank \\
    \midrule
    GS & 0.4750 & 1.0000 & 0.0500 & 1.0000 & 0.96 & 2 \\
    BS & 0.0294 & 0.8817 & 0.9412 & 1.0000 & 0.96 & 13 \\
    Llama3.3-70b & 0.3824 & 0.8817 & 0.2353 & 1.0000 & 0.96 & 4 \\
    Qwen2.5-7b & 0.1830 & 0.2857 & 1.0000 & 0.5294 & 0.94 & 12 \\
    Qwen2.5-14b & 0.3529 & 0.8318 & 0.4118 & 0.8824 & 0.96 & 7 \\
    Qwen2.5-32b & 0.3529 & 0.6383 & 0.5882 & 0.7059 & 0.96 & 8 \\
    Qwen2.5-72b & 0.4412 & 0.8220 & 0.2353 & 0.8824 & 0.96 & 3 \\
    DeepSeek-R1-7b & 0.2157 & 0.1419 & 1.0000 & 0.3529 & 0.93 & 11 \\
    DeepSeek-R1-14b & 0.2353 & 0.8817 & 0.5294 & 1.0000 & 0.96 & 10 \\
    DeepSeek-R1-32b & 0.3824 & 0.8817 & 0.2353 & 1.0000 & 0.96 & 4 \\
    DeepSeek-R1-70b & 0.3824 & 0.8817 & 0.2353 & 1.0000 & 0.96 & 4 \\
    GPT4-o3-mini & 0.8824 & 0.1494 & 0.0588 & 0.1765 & 0.96 & 1 \\
    GPT4-o3 & 0.2745 & 0.0469 & 1.0000 & 0.1765 & 0.93 & 9 \\
    \bottomrule
    \end{tabular}
    \label{tab:LSTM}
\end{table}

\begin{table}[h]
    \centering
    \setlength{\tabcolsep}{3pt}
    \renewcommand\arraystretch{1.5}
    \caption{Performance comparison of different solvers for scientific computing fields, where we evaluate the tensor wheel decomposition method on the multi-spectral image completion dataset.}
    \begin{tabular}{l|cccccc}
    \toprule
    Solver           & Score$\uparrow$ & $\text{AUP}_D\downarrow$ & Best-Time$\downarrow$ & Stop-Time$\downarrow$ & Best Result$\uparrow$ & Rank \\
    \midrule
    GS & 0.2656 & 1.0000 & 0.4688 & 1.0000 & 41.50 & 11 \\
    BS & 0.1066 & 0.9638 & 0.7869 & 1.0000 & 41.50 & 13 \\
    Llama3.3-70b & 0.4016 & 0.8979 & 0.2623 & 0.9344 & 41.50 & 8 \\
    Qwen2.5-7b & 0.7377 & 0.2904 & 0.2295 & 0.2951 & 41.50 & 1 \\
    Qwen2.5-14b & 0.5410 & 0.2779 & 0.4262 & 0.4918 & 41.50 & 2 \\
    Qwen2.5-32b & 0.2869 & 0.9638 & 0.4262 & 1.0000 & 41.50 & 10 \\
    Qwen2.5-72b & 0.1803 & 0.9638 & 0.6393 & 1.0000 & 41.50 & 12 \\
    DeepSeek-R1-7b & 0.3216 & 0.3009 & 1.0000 & 0.3443 & 41.21 & 9 \\
    DeepSeek-R1-14b & 0.4426 & 0.8687 & 0.2295 & 0.8852 & 41.50 & 4 \\
    DeepSeek-R1-32b & 0.4426 & 0.6645 & 0.4426 & 0.6721 & 41.50 & 4 \\
    DeepSeek-R1-70b & 0.4136 & 0.1477 & 1.0000 & 0.1639 & 41.34 & 7 \\
    GPT4-o3-mini & 0.5410 & 0.2623 & 0.4590 & 0.4590 & 41.50 & 3 \\
    GPT4-o3 & 0.4217 & 0.0539 & 1.0000 & 0.1475 & 41.34 & 6 \\
    \bottomrule
    \end{tabular}
    \label{tab:TW}
\end{table}

\begin{table}[h]
    \centering
    \setlength{\tabcolsep}{3pt}
    \renewcommand\arraystretch{1.5} 
    \caption{Performance comparison of different solvers for the image segmentation task based on the SIREN model.}
    \begin{tabular}{l|cccccc}
    \toprule
    Solver           & Score$\uparrow$ & $\text{AUP}_D\downarrow$ & Best-Time$\downarrow$ & Stop-Time$\downarrow$ & Best Result$\uparrow$ & Rank \\
    \midrule
    GS & 0.1100 & 1.0000 & 0.7800 & 1.0000 & 16.63 & 13 \\
    BS & 0.3295 & 0.7612 & 0.4773 & 0.8636 & 16.63 & 11 \\
    Qwen2.5-7b & 0.6023 & 0.4876 & 0.3636 & 0.4318 & 16.63 & 2 \\
    Qwen2.5-14b & 0.4545 & 0.7085 & 0.2727 & 0.8182 & 16.63 & 4 \\
    Qwen2.5-32b & 0.3743 & 0.5382 & 0.6136 & 0.6364 & 16.60 & 9 \\
    Qwen2.5-72b & 0.4088 & 0.5362 & 0.6364 & 0.5455 & 16.60 & 5 \\
    DeepSeek-r1-7b & 0.6932 & 0.3187 & 0.2273 & 0.3864 & 16.63 & 1 \\
    DeepSeek-r1-14b & 0.4085 & 0.4718 & 0.7045 & 0.4773 & 16.60 & 6 \\
    DeepSeek-r1-32b & 0.3125 & 0.1984 & 1.0000 & 0.3409 & 15.77 & 12 \\
    DeepSeek-r1-70b & 0.3746 & 0.5763 & 0.5909 & 0.6591 & 16.60 & 8 \\
    Llama3.3-70b & 0.3803 & 0.2859 & 0.5909 & 0.2955 & 10.79 & 7 \\
    GPT4-o3-mini & 0.4773 & 0.5177 & 0.4545 & 0.5909 & 16.63 & 3 \\
    GPT4-o3 & 0.3636 & 0.6714 & 0.4773 & 0.7955 & 16.63 & 10 \\
    \bottomrule
    \end{tabular}
    \label{tab:siren_seg}
\end{table}

\begin{table}[h]
    \centering
    \setlength{\tabcolsep}{3pt}
    \renewcommand\arraystretch{1.5}
    \caption{Performance comparison of different solvers for the image denoising task based on the SIREN model.}
    \begin{tabular}{l|cccccc}
    \toprule
    Solver & Score$\uparrow$ & $\text{AUP}_D\downarrow$ & Best-Time$\downarrow$ & Stop-Time$\downarrow$ & Best Result$\uparrow$ & Rank \\
    \midrule
    GS & 0.1500 & 1.0000 & 0.7000 & 1.0000 & 24.78 & 12 \\
    BS & 0.1477 & 0.8819 & 0.7045 & 1.0000 & 24.78 & 13 \\
    Qwen2.5-7b & 0.7045 & 0.1802 & 0.1364 & 0.4545 & 24.78 & 4 \\
    Llama3.3-70b & 0.5706 & 0.1105 & 0.5455 & 0.2955 & 24.27 & 8 \\
    Qwen2.5-14b & 0.4205 & 0.5205 & 0.5455 & 0.6136 & 24.78 & 10 \\
    Qwen2.5-32b & 0.5795 & 0.4762 & 0.3182 & 0.5227 & 24.78 & 7 \\
    Qwen2.5-72b & 0.7614 & 0.1543 & 0.1364 & 0.3409 & 24.78 & 1 \\
    DeepSeek-R1-7b & 0.5341 & 0.4193 & 0.3864 & 0.5455 & 24.78 & 9 \\
    DeepSeek-R1-14b & 0.7159 & 0.1639 & 0.0909 & 0.4773 & 24.78 & 2 \\
    DeepSeek-R1-32b & 0.7159 & 0.1284 & 0.2045 & 0.3636 & 24.78 & 2 \\
    DeepSeek-R1-70b & 0.6818 & 0.1359 & 0.2955 & 0.3409 & 24.78 & 5 \\
    GPT4-o3-mini & 0.2703 & 0.4389 & 0.7955 & 0.6591 & 24.48 & 11 \\
    GPT4-o3 & 0.6023 & 0.4786 & 0.2045 & 0.5909 & 24.78 & 6 \\
    \bottomrule
    \end{tabular}
    \label{tab:siren_denoising}
\end{table}

\begin{table}[h]
    \centering
    \setlength{\tabcolsep}{3pt}
    \renewcommand\arraystretch{1.5}
    \caption{Performance comparison of different solvers for Mask AutoEncoders (MAE)~\cite{mae}.}
    \begin{tabular}{l|cccccc}
    \toprule
    Solver           & Score$\uparrow$ & $\text{AUP}_D\downarrow$ & Best-Time$\downarrow$ & Stop-Time$\downarrow$ & Best Result$\uparrow$ & Rank \\
    \midrule
    GS & 0.1000 & 1.0000 & 0.8000 & 1.0000 & 85.0 & 12 \\
    BS & 0.0294 & 0.8093 & 0.9412 & 1.0000 & 85.0 & 13 \\
    Llama3.3-70b & 0.2941 & 0.8093 & 0.4118 & 1.0000 & 85.0 & 10 \\
    Qwen2.5-7b & 0.7647 & 0.0112 & 0.2353 & 0.2353 & 85.0 & 2 \\
    Qwen2.5-14b & 0.3471 & 0.1513 & 1.0000 & 0.2941 & 84.5 & 9 \\
    Qwen2.5-32b & 0.2647 & 0.8093 & 0.4706 & 1.0000 & 85.0 & 11 \\
    Qwen2.5-72b & 0.4412 & 0.8093 & 0.1176 & 1.0000 & 85.0 & 7 \\
    DeepSeek-R1-7b & 0.4104 & 0.0080 & 1.0000 & 0.1765 & 84.9 & 8 \\
    DeepSeek-R1-14b & 0.6176 & 0.3853 & 0.2941 & 0.4706 & 85.0 & 5 \\
    DeepSeek-R1-32b & 0.6765 & 0.1108 & 0.2941 & 0.3529 & 85.0 & 3 \\
    DeepSeek-R1-70b & 0.5588 & 0.2235 & 0.1765 & 0.7059 & 85.0 & 6 \\
    GPT4-o3-mini & 0.8235 & 0.0088 & 0.1176 & 0.2353 & 85.0 & 1 \\
    GPT4-o3 & 0.6471 & 0.1141 & 0.3529 & 0.3529 & 85.0 & 4 \\
    \bottomrule
    \end{tabular}
    \label{tab:MAE}
\end{table}

\begin{table}[h]
    \centering
    \setlength{\tabcolsep}{3pt}
    \renewcommand\arraystretch{1.5}
    \caption{Performance comparison of different solvers for the image classification task on the ImageNet dataset reported from the ResNet benchmark.}
    \begin{tabular}{l|cccccc}
    \toprule
    Solver           & Score$\uparrow$ & $\text{AUP}_D\downarrow$ & Best-Time$\downarrow$ & Stop-Time$\downarrow$ & Best Result$\uparrow$ & Rank \\
    \midrule
    GS & 0.0000 & 1.0000 & 1.0000 & 1.0000 & 21.43 \\
    BS & 0.0833 & 0.6394 & 0.8333 & 1.0000 & 21.43 \\
    Llama3.3-70b & 0.3333 & 0.6394 & 0.3333 & 1.0000 & 21.43 \\
    Qwen2.5-7b & 0.6667 & 0.0865 & 0.3333 & 0.3333 & 21.43 \\
    Qwen2.5-14b & 0.8333 & 0.0000 & 0.1667 & 0.1667 & 21.43 \\
    Qwen2.5-32b & 0.5833 & 0.2192 & 0.3333 & 0.5000 & 21.43 \\
    Qwen2.5-72b & 0.4167 & 0.2988 & 0.5000 & 0.6667 & 21.43 \\
    DeepSeek-R1-7b & 0.4167 & 0.3322 & 0.5000 & 0.6667 & 21.43 \\
    DeepSeek-R1-14b & 0.6667 & 0.0865 & 0.3333 & 0.3333 & 21.43 \\
    DeepSeek-R1-32b & 0.7500 & 0.0865 & 0.1667 & 0.3333 & 21.43 \\
    DeepSeek-R1-70b & 0.5000 & 0.2443 & 0.5000 & 0.5000 & 21.43 \\
    GPT4-o3-mini & 0.5000 & 0.2192 & 0.5000 & 0.5000 & 21.43 \\
    GPT4-o3 & 0.5833 & 0.2192 & 0.3333 & 0.5000 & 21.43 \\
    \bottomrule
    \end{tabular}
    \label{tab:ResNet}
\end{table}


\begin{table}[h]
    \centering
    \setlength{\tabcolsep}{3pt}
    \renewcommand\arraystretch{1.5}
    \caption{Performance comparison of different solvers for the image classification task on the LCBench dataset reported from~\cite{zimmer2021auto}.}
        \begin{tabular}{l|cccccc}
        \toprule
        Solver           & Score$\uparrow$ & $\text{AUP}_D\downarrow$ & Best-Time$\downarrow$ & Stop-Time$\downarrow$ & Best Result$\uparrow$ & Rank \\
        \midrule
        GS & 0.1750 & 1.0000 & 0.6500 & 1.0000 & 88.29 \\
        BS & 0.4189 & 0.9985 & 0.1622 & 1.0000 & 88.29 \\
        Llama3.3-70b & 0.4738 & 0.0052 & 1.0000 & 0.0486 & 87.98 \\
        Qwen2.5-7b & 0.4619 & 0.0047 & 1.0000 & 0.0451 & 85.78 \\
        Qwen2.5-14b & 0.4739 & 0.0017 & 1.0000 & 0.0180 & 85.62 \\
        Qwen2.5-32b & 0.4718 & 0.0022 & 1.0000 & 0.0225 & 85.62 \\
        Qwen2.5-72b & 0.4599 & 0.0081 & 1.0000 & 0.0766 & 87.98 \\
        DeepSeek-R1-7b & 0.4691 & 0.0027 & 1.0000 & 0.0280 & 85.62 \\
        DeepSeek-R1-14b & 0.4579 & 0.0086 & 1.0000 & 0.0806 & 87.98 \\
        DeepSeek-R1-32b & 0.4677 & 0.0035 & 1.0000 & 0.0331 & 85.78 \\
        DeepSeek-R1-70b & 0.4739 & 0.0014 & 1.0000 & 0.0180 & 85.62 \\
        GPT4-o3-mini & - & - & - & - & - & - \\
        GPT4-o3 & - & - & - & - & - & - \\
        \bottomrule
    \end{tabular}
    \label{tab:LCBenchmark}
\end{table}

\begin{table}[h]
    \centering
    \setlength{\tabcolsep}{3pt}
    \renewcommand\arraystretch{1.5}
    \caption{Performance comparison of different solvers for the medical image segmentation task on the BTCV dataset reported from the nnUnet benchmark~\cite{nnunet_revisit}.}
    \begin{tabular}{l|cccccc}
    \toprule
    Solver           & Score$\uparrow$ & $\text{AUP}_D\downarrow$ & Best-Time$\downarrow$ & Stop-Time$\downarrow$ & Best Result$\uparrow$ & Rank \\
    \midrule
    GS & 0.4211 & 1.0000 & 0.1579 & 1.0000 & 85.04 & 3 \\
    BS & 0.0625 & 0.6453 & 0.8750 & 1.0000 & 85.04 & 13 \\
    Llama3.3-70b & 0.3750 & 0.6453 & 0.2500 & 1.0000 & 85.04 & 5 \\
    Qwen2.5-7b & 0.4688 & 0.3441 & 0.5000 & 0.5625 & 85.04 & 2 \\
    Qwen2.5-14b & 0.2813 & 0.6453 & 0.4375 & 1.0000 & 85.04 & 7 \\
    Qwen2.5-32b & 0.2813 & 0.4742 & 0.6875 & 0.7500 & 85.04 & 7 \\
    Qwen2.5-72b & 0.1875 & 0.6453 & 0.6250 & 1.0000 & 85.04 & 11 \\
    DeepSeek-R1-7b & 0.3680 & 0.0813 & 1.0000 & 0.1250 & 83.69 & 6 \\
    DeepSeek-R1-14b & 0.1983 & 0.0505 & 1.0000 & 0.1875 & 80.69 & 10 \\
    DeepSeek-R1-32b & 0.1875 & 0.6453 & 0.6250 & 1.0000 & 85.04 & 11 \\
    DeepSeek-R1-70b & 0.2813 & 0.6453 & 0.4375 & 1.0000 & 85.04 & 7 \\
    GPT4-o3-mini & 0.6875 & 0.1466 & 0.2500 & 0.3750 & 85.04 & 1 \\
    GPT4-o3 & 0.4063 & 0.5081 & 0.4375 & 0.7500 & 85.04 & 4 \\
    \bottomrule
    \end{tabular}
    \label{tab:MIABench}
\end{table}

\begin{table}[h]
    \centering
    \setlength{\tabcolsep}{3pt}
    \renewcommand\arraystretch{1.5}
    \caption{Performance comparison of different solvers for the nnUnet model on the BraTS dataset~\cite{brats}.}
    \begin{tabular}{l|cccccc}
    \toprule
    Solver           & Score$\uparrow$ & $\text{AUP}_D\downarrow$ & Best-Time$\downarrow$ & Stop-Time$\downarrow$ & Best Result$\uparrow$ & Rank \\
    \midrule
    GS & 0.1667 & 1.0000 & 0.6667 & 1.0000 & 82.45 & 10 \\
    BS & 0.0333 & 0.3786 & 0.9333 & 1.0000 & 82.45 & 13 \\
    Llama3.3-70b & 0.1333 & 0.3786 & 0.7333 & 1.0000 & 82.45 & 11 \\
    Qwen2.5-7b & 0.1000 & 0.3786 & 0.8000 & 1.0000 & 82.45 & 12 \\
    Qwen2.5-14b & 0.4333 & 0.0698 & 0.5333 & 0.6000 & 82.45 & 1 \\
    Qwen2.5-32b & 0.3333 & 0.3118 & 0.5333 & 0.8000 & 82.45 & 3 \\
    Qwen2.5-72b & 0.2333 & 0.0883 & 0.7333 & 0.8000 & 82.45 & 9 \\
    DeepSeek-R1-7b & 0.2995 & 0.0481 & 1.0000 & 0.4000 & 82.41 & 5 \\
    DeepSeek-R1-14b & 0.2940 & 0.0406 & 1.0000 & 0.4000 & 82.00 & 7 \\
    DeepSeek-R1-32b & 0.2667 & 0.0839 & 0.6667 & 0.8000 & 82.45 & 8 \\
    DeepSeek-R1-70b & 0.3333 & 0.3126 & 0.5333 & 0.8000 & 82.45 & 3 \\
    GPT4-o3-mini & 0.4000 & 0.2347 & 0.6000 & 0.6000 & 82.45 & 2 \\
    GPT4-o3 & 0.2984 & 0.0484 & 1.0000 & 0.4000 & 82.33 & 6 \\
    \bottomrule
    \end{tabular}
    \label{tab:nnUnet}
\end{table}

\begin{table}[h]
    \centering
    \setlength{\tabcolsep}{3pt}
    \renewcommand\arraystretch{1.5}
    \caption{Performance comparison of different solvers for the GraphSAGE model on the imbalanced node classification task.}
    \begin{tabular}{l|cccccc}
    \toprule
    Solver           & Score$\uparrow$ & $\text{AUP}_D\downarrow$ & Best-Time$\downarrow$ & Stop-Time$\downarrow$ & Best Result$\uparrow$ & Rank \\
    \midrule
    GS & 0.1000 & 1.0000 & 0.8000 & 1.0000 & 89.28 & 13 \\
    BS & 0.2727 & 0.8982 & 0.4545 & 1.0000 & 89.28 & 8 \\
    Llama3.3-70b & 0.1591 & 0.8982 & 0.6818 & 1.0000 & 89.28 & 9 \\
    Qwen2.5-7b & 0.2880 & 0.0928 & 1.0000 & 0.4091 & 89.12 & 6 \\
    Qwen2.5-14b & 0.2894 & 0.0890 & 1.0000 & 0.4091 & 89.15 & 5 \\
    Qwen2.5-32b & 0.2880 & 0.3564 & 1.0000 & 0.4091 & 89.12 & 6 \\
    Qwen2.5-72b & 0.1364 & 0.8982 & 0.7273 & 1.0000 & 89.28 & 12 \\
    DeepSeek-R1-7b & 0.3116 & 0.3083 & 1.0000 & 0.3636 & 89.15 & 3 \\
    DeepSeek-R1-14b & 0.1591 & 0.8982 & 0.6818 & 1.0000 & 89.28 & 9 \\
    DeepSeek-R1-32b & 0.1591 & 0.8657 & 0.7273 & 0.9545 & 89.28 & 11 \\
    DeepSeek-R1-70b & 0.2955 & 0.7096 & 0.5909 & 0.8182 & 89.28 & 4 \\
    GPT4-o3-mini & 0.7727 & 0.1463 & 0.0455 & 0.4091 & 89.282 & 1 \\
    GPT4-o3 & 0.7727 & 0.2540 & 0.1818 & 0.2727 & 89.282 & 1 \\
    \bottomrule
    \end{tabular}
    \label{tab:GraphSAGE}
\end{table}


\begin{table}[h]
    \centering
    \setlength{\tabcolsep}{3pt}
    \renewcommand\arraystretch{1.5}
    \caption{Performance comparison of different solvers for bioactivity prediction on Chagas EP20 dataset~\cite{Chagas}}
    \begin{tabular}{l|cccccc}
    \toprule
    Solver           & Score$\uparrow$ & $\text{AUP}_D\downarrow$ & Best-Time$\downarrow$ & Stop-Time$\downarrow$ & Best Result$\uparrow$ & Rank \\
    \midrule
    GS & 0.4833 & 1.0000 & 0.0333 & 1.0000 & 0.754 & 11 \\
    BS & 0.3333 & 0.8941 & 0.3333 & 1.0000 & 0.754 & 13 \\
    Llama3.3-70b & 0.4630 & 0.8941 & 0.0741 & 1.0000 & 0.754 & 12 \\
    Qwen2.5-7b & 0.9074 & 0.0772 & 0.0741 & 0.1111 & 0.754 & 1 \\
    Qwen2.5-14b & 0.8889 & 0.0649 & 0.1111 & 0.1111 & 0.754 & 2 \\
    Qwen2.5-32b & 0.6296 & 0.4231 & 0.0741 & 0.6667 & 0.754 & 9 \\
    Qwen2.5-72b & 0.5741 & 0.6799 & 0.0741 & 0.7778 & 0.754 & 10 \\
    DeepSeek-R1-7b & 0.7778 & 0.1104 & 0.2222 & 0.2222 & 0.754 & 4 \\
    DeepSeek-R1-14b & 0.7037 & 0.5510 & 0.0741 & 0.5185 & 0.754 & 7 \\
    DeepSeek-R1-32b & 0.6852 & 0.3405 & 0.0741 & 0.5556 & 0.754 & 8 \\
    DeepSeek-R1-70b & 0.7778 & 0.3160 & 0.1111 & 0.3333 & 0.754 & 4 \\
    GPT4-o3-mini & 0.7407 & 0.3546 & 0.1852 & 0.3333 & 0.754 & 6 \\
    GPT4-o3 & 0.8148 & 0.0622 & 0.0370 & 0.3333 & 0.754 & 3 \\
    \bottomrule
    \end{tabular}
    \label{tab:chagas}
\end{table}

\section{Prompts of Compared Methods}
\label{sec:supp_prompt}

\begin{tcolorbox}[breakable,colback=orange!5!white, colframe=black, title={Search Prompt of LLM solver}]
Instructions:

**Search Space** (Numbering starts from 0, excluding Completed Trials):
\pkd{trials}
    
Task 1: Optimization Recommendation
Recommend exactly \pkd{n\_jobs} promising trials from the provided **Search Space** (include both number and params).

\textbf{Rules:}

1. ``params'' MUST include:

\,\,\pkd{HyperName}

2. All selected `params` must match exactly with the provided **Search Space**. Do NOT leave out any key.

3. Use the analysis in **Task 1** (trial analysis, performance trends, highlights, and other insights) to guide selection.

4. Based on the above analysis, explore under-explored regions only when there is clear evidence of potential performance gain.

5. Do not mix, modify, or create new values.

6. You MUST not output any JSON blocks in this part.

7. You MUST provide reasoning for each recommendation.
\end{tcolorbox}

\begin{tcolorbox}[breakable,colback=orange!5!white, colframe=black, title={Early-Stopping Prompt of LLM-ES solver}]
Completed trials:
\pkd{completed\_trials}

The following **Search Space** contains **unexplored** trials.
\pkd{trials}

If the optimization process should be stopped, Answer: Yes with confidence score: \pkd{confidence\_socre}. Otherwise, Answer: No with confidence score: \pkd{confidence\_socre}.

Finally, you MUST output 'Answer: No/Yes' with confidence score: \pkd{confidence\_socre}.
\end{tcolorbox}

\end{appendices}

\end{document}